\documentclass[]{ant_tech_report}

\usepackage{latexsym}
\usepackage[utf8]{inputenc}
\usepackage{inconsolata}
\ifdefined\pdfimageresolution
\fi
\usepackage{siunitx}
\usepackage{url}
\usepackage{algorithm}
\usepackage{multicol}
\usepackage{colortbl}
\usepackage{wrapfig}
\usepackage{makecell}
\usepackage{adjustbox}
\usepackage[noend]{algpseudocode}
\usepackage{fancyvrb}
\usepackage{fvextra}
\usepackage{soul}
\usepackage{float}
\usepackage{tikz}
\usepackage{fontawesome5}
\titleformat*{\paragraph}{\bfseries\itshape}

\usepackage{hyperref}
\usepackage{url}

\usepackage{graphicx}
\usepackage{booktabs}
\usepackage{multicol}
\usepackage{multirow}
\usepackage{colortbl}
\usepackage{pifont}
\usepackage{enumitem}
\usepackage[table]{xcolor}
\usepackage{wrapfig}
\usepackage{caption}
\usepackage{subcaption}
\usepackage{array}

\usepackage{amsthm}

\definecolor{oursrow}{RGB}{233,241,252}
\definecolor{gaingreen}{RGB}{27,120,55}
\definecolor{dropred}{RGB}{165,15,21}

\usepackage{arydshln}
\newcommand{\dashmidrule}{%
  \noalign{\vskip\aboverulesep}%
  \cdashline{1-1}%
  \cdashline{2-8}%
  \cdashline{9-10}%
  \noalign{\vskip\belowrulesep}%
}

\newcommand{\modelscale}[1]{%
  \noalign{\vskip 1.5pt}%
  \multicolumn{10}{@{}l@{}}{\textbf{\textit{#1}}} \\
  \noalign{\vskip 1.5pt}%
}

\newcommand{\improve}[1]{\textcolor{gaingreen}{\textbf{#1}}}
\newcommand{\gain}[1]{{\scriptsize\textcolor{gaingreen}{$\,\uparrow$\textbf{#1}}}}
\newcommand{\drop}[1]{{\scriptsize\textcolor{dropred}{$\,\downarrow$\textbf{#1}}}}
\newcommand{\std}[1]{{\fontsize{6pt}{6pt}\selectfont$\pm$#1}}

\definecolor{memguiblue}{HTML}{E8F1FA}
\definecolor{cvprblue}{rgb}{0.21,0.49,0.74}
\definecolor{memorycolor}{RGB}{216, 228, 244}
\definecolor{foldcolor}{RGB}{229, 190, 217}
\definecolor{stepcolor}{RGB}{232, 204, 204}
\definecolor{memguigreen}{RGB}{71, 172, 20}
\definecolor{memguired}{RGB}{238, 68, 51}
\definecolor{toolblue}{HTML}{263D4A}
\definecolor{histcolor}{HTML}{B79AD0}
\definecolor{recentcolor}{HTML}{E46F6F}
\definecolor{memstatecolor}{HTML}{F2A65A}
\definecolor{codegray}{gray}{0.95}

\newenvironment{paperresources}
  {\par\noindent\begin{minipage}{0.92\textwidth}
    \centering\footnotesize\sffamily\color{black!82}
    \setlength{\parindent}{0pt}\setlength{\parskip}{2pt}}
  {\end{minipage}\par\vspace{1.7mm}}

\newcommand{\resourcegithubicon}{\textcolor{antblue}{\faGithub}}

\newcommand{\paperresourceicon}[4]{%
  #1\ \textbf{#2:}\ \href{#3}{{\ttfamily\fontsize{8}{9}\selectfont #4}}%
}

\tcbset{
  promptstyle/.style={
    colback=codegray,
    colframe=black,
    fontupper=\ttfamily\scriptsize,
    fonttitle=\ttfamily\scriptsize\raggedright,
    coltitle=white,
    colbacktitle=black,
    sharp corners,
    boxrule=0.5pt,
    enhanced,
    breakable,
    left=4pt,
    right=4pt,
    top=4pt,
    bottom=4pt,
  }
}

\newcounter{insight}

\definecolor{hlbestcolor}{HTML}{CFE2FF}
\definecolor{hlsecondcolor}{HTML}{E8F1FF}
\definecolor{oursrowcolor}{HTML}{F4EEF8}
\definecolor{oursrowcolorsecond}{HTML}{F9F2FB}

\definecolor{risecolor}{HTML}{159A9C}

\newcommand{\includefig}[2][width=\linewidth]{\IfFileExists{#2}{\includegraphics[#1]{#2}}{\centering\fbox{\parbox{\dimexpr\linewidth-2\fboxsep-2\fboxrule}{\centering\small [Placeholder: \texttt{#2}]}}}}

\title{Learning from All-Failed Groups via First-Visit Coverage in Long-Horizon Agentic Tasks}

\author{%
\parbox{\textwidth}{\centering
Kaibing Yang$^{1,2}$,
Guangfeng Cai$^{1}$,
Shengtian Yang$^{1,2}$,
Shuo He$^{3}$,
Yu Li$^{1}$,
Mengyi Liu$^{2}$\\[1mm]
Pengwei Chen$^{2}$,
Jun Xu$^{2,}$\textsuperscript{*},
Lei Feng$^{1,\S}$
}}

\affiliation{%
\parbox{\textwidth}{\centering\small
$^1$Southeast University \quad $^2$Kuaishou Technology
\quad $^3$Meituan
}}

\renewcommand{\contributionlist}{\vspace{-2mm}}
\newcommand{\authorfootnotes}{%
{\scriptsize
\begin{tabular}{@{}l@{}}
\rule{34mm}{0.4pt}\\[-0.2ex]
\textsuperscript{*}Work done during an internship at Kuaishou Technology, supervised by Jun Xu. \\
$^\S$Corresponding author.
\end{tabular}}}
\fancypagestyle{firststyle}{
    \fancyhf{}
    \fancyhead[R]{
        \vskip 3mm
    \includegraphics[height=13mm]{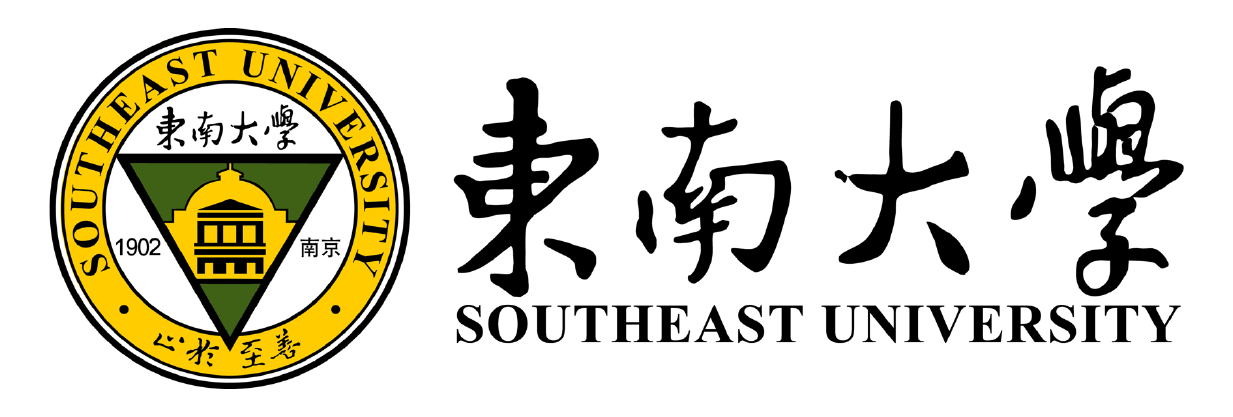}
    }
    \fancyhead[L]{
        \vskip 3mm
    \includegraphics[height=10mm]{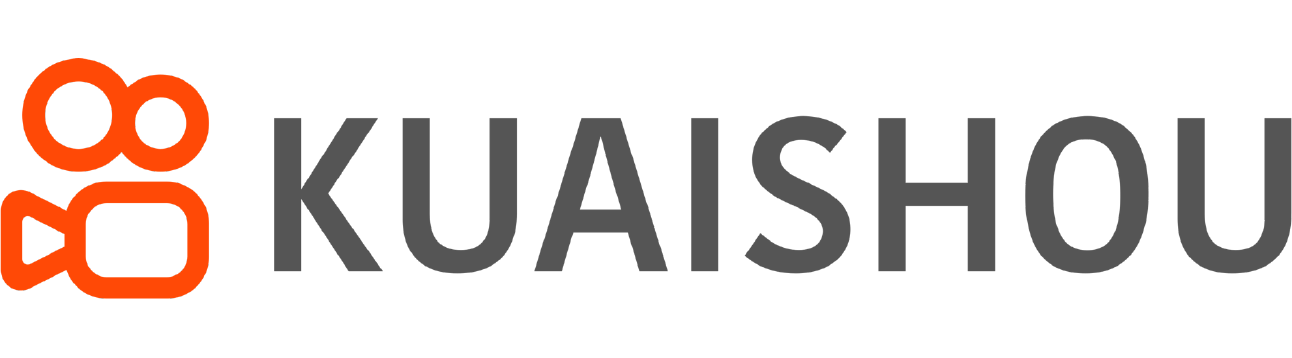}
    }
    \fancyfoot[L]{\raisebox{1.15\baselineskip}[0pt][0pt]{\authorfootnotes}}
    \fancyfoot[C]{\thepage}
}
\renewcommand{\checkdatalist}{%
\begin{paperresources}
\paperresourceicon{\resourcegithubicon}{Code}{https://github.com/KeibingYang/Coverage-conditioned-Group-Policy-Optimization-for-Long-Horizon-Agentic-Task}{github.com/KeibingYang/CovGPO}

\end{paperresources}}

\abstract{\begin{abstract}

Group-based policy optimization has been increasingly used to train large language model (LLM) agents from sparse outcome rewards by comparing trajectories or steps within a group.
However, on difficult long-horizon agentic tasks, we found this comparison suffers from a sampling imbalance: \textit{repeated or low-effect actions dominate the high-probability region of the policy}, leaving \textit{useful state-changing actions under-sampled}.
This imbalance produces many \textbf{all-failed rollout groups}, where the relative advantages within such a group provide no direction for updating the policy.
Consequently, the sampling imbalance and the all-failed groups collapse into a self-perpetuating \emph{\textbf{credit trap}}: \textit{failure-dominated sampling yields no learning signal, allowing repeated low-effect actions to persist.}
To break this trap, we propose \underline{Cov}erage-conditioned \underline{G}roup \underline{P}olicy \underline{O}ptimization (CovGPO), which uses first-visit observation coverage only when all samples in a group receive zero outcome reward.
Specifically, within such groups, CovGPO assigns higher relative advantages to state-exploring actions without corrupting the original reward signal, as reaching new observations is a prerequisite for task success.
Experiments on two challenging agentic benchmarks, ALFWorld and WebShop, show that CovGPO consistently improves over group-based baselines, with particularly large gains on hard tasks. 

\end{abstract}

}

\begin{document}
\maketitle
\section{Introduction}
\label{sec:intro}
LLM agents~\citep{achiam2023gpt, team2023gemini, yang2025qwen3, team2025kimi}
are increasingly moving beyond single-turn response generation toward interactive problem solving in open-ended environments such as household tasks~\citep{shridhar2020alfworld, li2024embodied}, web navigation~\citep{gou2025navigating, furuta2024multimodal}, and tool use~\citep{schick2023toolformer, qian2026toolrl}. 
These tasks require agents to plan, act, observe, and revise over many steps~\citep{feng2025towards, wang2023voyager, wang2024mobile}, while success is often revealed only through sparse outcome feedback. 
Such sparse supervision leaves most intermediate decisions without direct task feedback, making credit assignment over long trajectories difficult.

To train agents under such sparse outcome feedback, recent work has adopted group-based policy optimization~\citep{shao2024deepseekmath, yu2026dapo, kool2019buy}, which samples multiple trajectories for the same task and updates the policy using their relative outcomes instead of a learned value function. 
At the trajectory level, GRPO compares the complete trajectories within each rollout group and assigns a single advantage to their constituent steps.
To provide finer-grained credit, GiGPO~\citep{feng2026group} groups steps that share an anchor observation and produces step-varying advantages within the same trajectory, while HGPO~\citep{he2026hierarchy} further groups steps by their shared interaction history.
However, they all rely on the same assumption: \textit{outcome variation among compared trajectories or steps}. When this variation vanishes, the comparison yields no signal.

\begin{figure}
    \vspace{-3mm}
    \centering
    \includegraphics[width=1\linewidth]{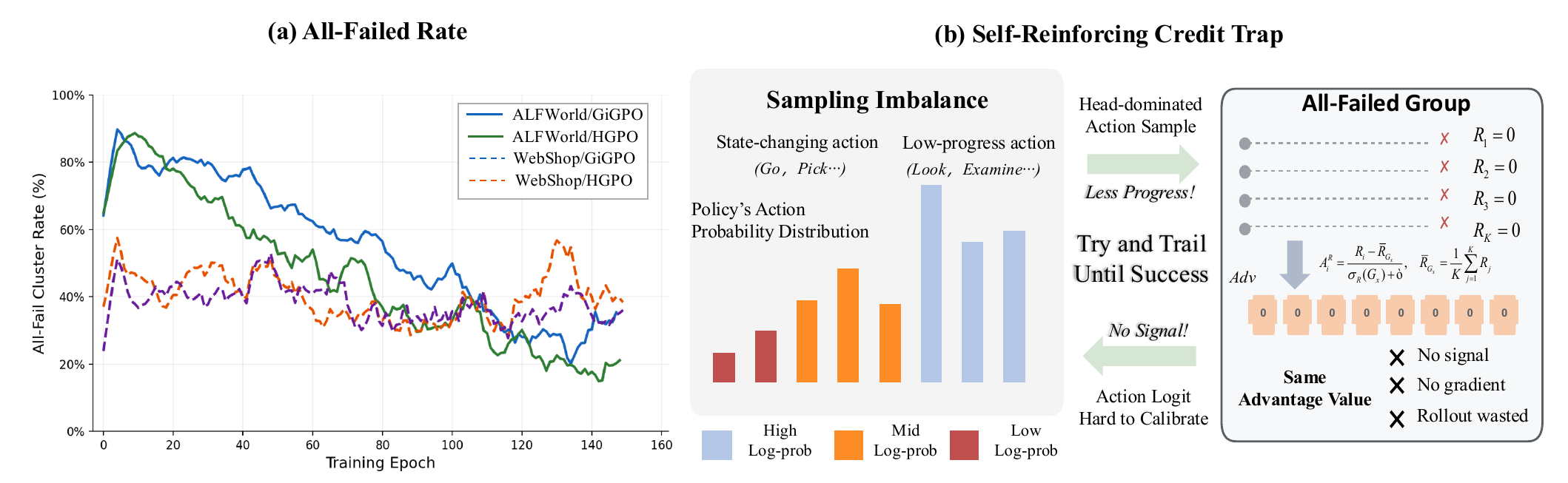}
    \vspace{-3mm}
    \caption{All-failed rate across training and credit trap. (a) Most rollout groups are all-failed during early training, yielding zero reward-based advantages. (b) The base policy concentrates probability mass on low-effect actions, leaving state-changing actions under-sampled. These two effects reinforce each other: imbalanced sampling produces more all-failed groups, while the resulting zero gradients prevent the policy from correcting the imbalance.}
    \vspace{-4mm}
    \label{fig:teaser}
\end{figure}

This assumption is often reasonable in short-horizon or moderately difficult tasks~\citep{liu2025understanding, yu2026dapo, wei2026swe}, where some sampled trajectories succeed while others fail.
However, in hard long-horizon environments~\citep{shridhar2020alfworld, yao2022webshop}, many \emph{all-failed rollout groups} occur: \textbf{every trajectory receives the same failure outcome}.
In such groups, the policy exhibits a sampling imbalance: \textit{high-probability actions tend to be low-effect while effective actions remain under-sampled} (Fig.~\ref{fig:teaser}).
This leads to a \emph{self-reinforcing credit trap}: failure-dominated sampling produces all-failed groups, and because every trajectory shares the same outcome, the reward-based comparison provides no direction for correcting the imbalanced sampling.

To address this issue, recent research has shifted to learned progress models~\citep{wang2025spa, xi2026agentprm, chai2026progra}, world-model guidance~\citep{yu2026reinforcement}, and turn-level credit designs~\citep{ma2026tspo}. However, these designs primarily depend on auxiliary supervision or learned models, instead of mining the intrinsic behavioral signals within the rollouts themselves.
Therefore, an insightful question naturally arises: \textbf{\textit{Can we recover an actionable signal directly from existing all-failed rollouts, without relying on auxiliary supervision or learned models?}}

\begin{figure}
    \centering
    \includegraphics[width=1\linewidth]{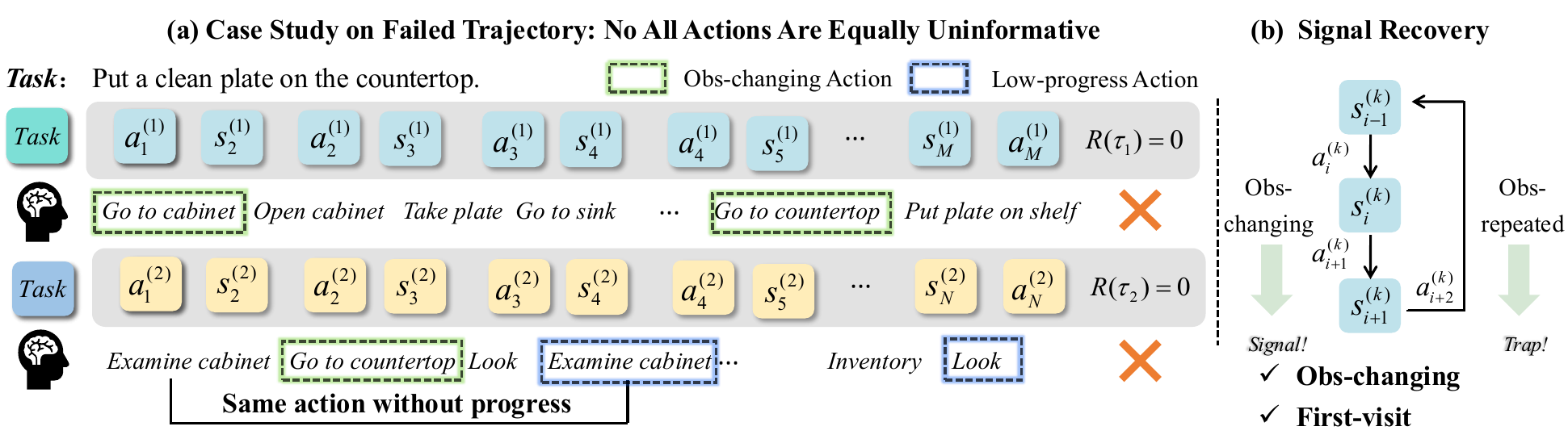}
    \vspace{-4mm}
    \caption{Not all failures are equal and progress signal. (a) Within the same all-failed group, some trajectories visit novel states while others stagnate on repeated no-change actions, revealing behavioral contrasts that reward-only algorithm discard.
    (b) We recover this signal via first-visit observation coverage, where each transition to a previously unseen observation contributes to coverage.
    }  
    \label{fig:teaser_case}
    \vspace{-5mm}
\end{figure}

In this paper, we argue that the credit trap does not imply that failed rollouts are uninformative: equal failure outcomes do not imply equal behavior.
Within the same all-failed group, one trajectory may repeatedly revisit the same observations, while another reaches more distinct observations before ultimately failing (Fig.~\ref{fig:teaser_case} (a)). Although both receive the same terminal outcome, the latter exhibits more progress than the former.
Existing methods fail to capture this contrast: GRPO assigns both trajectories the same zero advantage, while GiGPO and HGPO compare steps whose inherited returns are also identical, causing their step-level advantages to vanish as well.

Motivated by this insight, we propose \underline{Cov}erage-conditioned \underline{G}roup \underline{P}olicy \underline{O}ptimization (CovGPO), a fallback algorithm that derives an exploration signal from all-failed groups via \textbf{first-visit observation coverage}. 
Specifically, CovGPO identifies an interaction step as coverage-expanding if and only if it transitions to an observation not previously visited within the same trajectory. 
CovGPO credits transitions that expand observation coverage while assigning zero credit to actions that merely revisit or preserve existing states, as reaching novel observations is a necessary prerequisite for task success.
Crucially, CovGPO applies this coverage credit \textit{exclusively} as a fallback mechanism within all-failed groups, reverting strictly to the original reward-based advantage whenever outcome variance exists. This avoids rewarding aimless exploration over concise task completion.
In this way, CovGPO breaks the self-reinforcing credit trap: within all-failed groups, trajectories reaching new observations receive positive advantages, promoting the coverage-expanding actions the base policy was under-sampling. 
Across two challenging agentic benchmarks, ALFWorld and WebShop, using Qwen2.5-1.5B/7B-Instruct models, CovGPO consistently boosts strong existing baselines. Our contributions are summarized as follows:

\begin{itemize}[leftmargin=*, itemsep=0pt, topsep=0pt, parsep=1pt]
    \item We identify the \emph{all-failed credit trap}: imbalanced sampling produces all-failed rollout groups whose identical outcomes prevent reward-based algorithm from correcting the imbalance, despite the behavioral contrast within these groups.
    \item We propose CovGPO, a conditional fallback algorithm that recovers credit in all-failed groups via first-visit observation coverage, a necessary precondition for task success, and reverts exactly to the base advantage when outcome variation returns.
    \item On ALFWorld and WebShop with Qwen2.5-1.5B/7B-Instruct, CovGPO consistently boosts strong group-based baselines under the same computational constraints.
\end{itemize}

\section{Related Work}

\noindent\textbf{Group-Based Reinforcement Learning for Long-Horizon LLM Agents.} Critic-free methods such as GRPO~\cite{shao2024deepseekmath} and RLOO~\citep{ahmadian2024back} estimate policy advantages from relative outcomes among multiple samples, reducing the memory and computation required by PPO-style actor--critic training~\citep{schulman2017proximal, zhou2024archer, hou2026single}. Their agentic extensions increasingly refine the comparison unit: GiGPO~\citep{feng2026group} combines trajectory-level and anchor-state groups, ProxMO~\citep{fang2026proxMO} introduces difficulty-aware modulation and semantic soft aggregation, and HGPO~\citep{he2026hierarchy} forms hierarchical groups to control historical-context inconsistency. 
SALT~\citep{li2026salt} and GraphGPO~\citep{cheng2026graphPO} exploit trajectory or state-transition graphs to propagate structural credit across related states. 
These methods improve the granularity or accuracy of credit assignment once a usable comparison signal exists. CovGPO addresses a complementary regime in which all sampled trajectories fail and the outcome statistics provide no signal, using first-visit observation coverage as a conditional fallback instead of introducing another grouping or value-estimation structure.

\noindent\textbf{Learning from all-failed and outcome-degenerate groups.}
Group-based policy optimization suffers from vanishing gradients when all sampled rollouts yield identical outcomes. 
In single-turn reasoning, existing remedies mitigate this degenerate-group issue via prompt resampling~\citep{yu2026dapo}, external scaffolding~\citep{bamba2025xrpo,zhang2025scaf}, virtual baselines~\citep{nan2025ngrpo}, or entropy-guided surrogate advantages~\citep{le2025no}. Moving to sequential decision-making\citep{cai2026beyond}, multi-turn search frameworks such as TSPO~\citep{ma2026tspo} and IGPO~\citep{wang2025IGPO} tackle credit assignment by designing heuristic step-level rewards or intermediate verifiers; however, they rely heavily on answer-dependent progress supervision or domain-specific scoring. While classic exploration methods (e.g., count-based bonuses or state-novelty measures like RIDE) encourage broader state visitation, they typically introduce non-stationary intrinsic rewards that risk distorting agent policies. 
However, these paradigms primarily cater to single-turn tasks or rely on task-specific heuristics, leaving the fundamental challenge of all-failure trajectories unresolved: in multi-step agent environments, extracting robust learning signals from purely negative trajectories is intrinsically difficult due to compound credit assignment. 
To our knowledge, CovGPO is the first framework tailored to resolve the all-failed regime in multi-turn agentic RL. 
Crucially, rather than enforcing answer-dependent intermediate supervision or altering the environment distribution, CovGPO quantifies relative credit via task-agnostic \textit{observation coverage} directly recycled from existing on-policy interactions, serving as a plug-and-play auxiliary signal that seamlessly integrates into diverse comparison group structures (e.g., GRPO, GiGPO, and HGPO).

\noindent\textbf{Auxiliary progress and exploration signals for agentic credit assignment.}
Sparse terminal rewards have motivated dense supervision from learned progress models, interaction structure, and intrinsic exploration.
SPA-RL~\citep{wang2025sparl} decomposes terminal outcomes into stepwise progress rewards, while iStar~\citep{liu2025istar} and AgentPRM~\citep{xi2026agentprm} learn process reward models for intermediate agent decisions.
IGPO~\citep{wang2025IGPO} measures progress through information gain.
Classical approaches such as RIDE~\citep{raileanu2020ride} instead reward actions that induce substantial state changes.
Unlike these persistent or model-based auxiliary signals, CovGPO uses model-free first-visit coverage only when outcome rewards provide no within-group ordering.

\section{Preliminaries and Problem Formulation}

\noindent\textbf{Multi-Turn Agentic Tasks.}
Let $x \sim p(X)$ be a task instruction and $\pi_\theta$ an LLM-based policy. At each step $t$, the policy produces $a_t \sim \pi_\theta(\cdot \mid h_t)$ conditioned on the interaction history $h_t=(x,o_1,a_1,\ldots,o_t)$, and the environment returns the next observation $o_{t+1}$. The resulting trajectory is $\tau=(o_1,a_1,\ldots,o_T,a_T)$, where $T$ is bounded by the interaction budget. 
Environments provide only a delayed terminal reward 
$R(\tau)\in[0,1]$, evaluated via binary completion in ALFWorld and fractional matching in WebShop, leaving intermediate transitions entirely unrewarded.

\noindent\textbf{Group-Based Reinforcement Learning.}
\label{sec:group_rl}
Group-based RL methods such as GRPO~\citep{shao2024deepseekmath} estimate advantages critic-free from intra-group statistics. For task $x$, a group $G_x=\{\tau_i\}_{i=1}^{K}$ of $K$ trajectories is sampled from $\pi_\theta$, each receiving an outcome reward $R_i=R(\tau_i)$. The trajectory-level group-relative advantage is
\begin{equation}
A_i^{R}
=
\frac{R_i-\bar{R}_{G_x}}
{\sigma_R(G_x)+\epsilon},
\quad
\bar{R}_{G_x}=\frac{1}{K}\sum_{j=1}^{K}R_j,
\label{eq:grpo_advantage}
\end{equation}
and every step in $\tau_i$ shares $A_i^{R}$. For finer-grained credit, recent variants form step-level groups: GiGPO~\citep{feng2026group} collects steps across $G_x$ that share the same environment state $s$ into $G^{S}(s)$ and normalizes their discounted returns $R^{S}_{i,t}=\gamma^{T_i-t}R_i$ within it, while HGPO~\citep{he2026hierarchy} groups steps by shared interaction history prefixes instead of a single anchor state.
\begin{equation}
A^{S}_{i,t}
=
\frac{R^{S}_{i,t}-\bar{R}_{G^{S}(s)}}
{\sigma_{R}(G^{S}(s))+\epsilon},
\label{eq:step_advantage}
\end{equation}
The policy is updated via a clipped surrogate objective with KL regularization against a reference $\pi_{\mathrm{ref}}$. 
Regardless of granularity, these algorithms work only when \emph{within-group reward variation} exists.

\noindent\textbf{All-Failed Groups and the Credit Trap.}
\label{sec:all_fail}
In hard long-horizon tasks, success is rare early in training, and a rollout group may contain no successful trajectory. We call such $G_x$ an \emph{all-failed rollout group}:
\begin{equation}
G_x \text{ is all-failed} \Longleftrightarrow R_i = 0,\forall \tau_i \in G_x \Longleftrightarrow \bar{R}_{G_x} = 0,
\label{eq:all_fail}
\end{equation}
where the second equivalence uses $R_i\ge0$. The trajectory-level algorithm in Eq.~\ref{eq:grpo_advantage} then degenerates:
\begin{equation}
\sigma_R(G_x)=0 \Longrightarrow A_i^{R}=0, \forall i\in{1,\ldots,K}.
\label{eq:degenerate}
\end{equation}
When a group is all-failed, within-group reward variance vanishes ($\sigma_R(G_x) = 0$), and every discounted step return collapses to zero ($R^{S}_{i,t} = \gamma^{T_i-t} \cdot 0 = 0$). Across all grouping granularities, such groups yield no relative gradient signal, as identified by the unified condition in Eq.~\ref{eq:all_fail}.
Unlike all-successful groups with positive rewards or fractional-reward settings with non-zero partial returns, all-failed groups offer strictly zero feedback across all members.

\section{Training Agents with CovGPO for Long-Horizon Agentic Tasks}
\label{sec:CovGPO}

\begin{figure}
    \vspace{-3mm}
    \centering
    \includegraphics[width=1\linewidth]{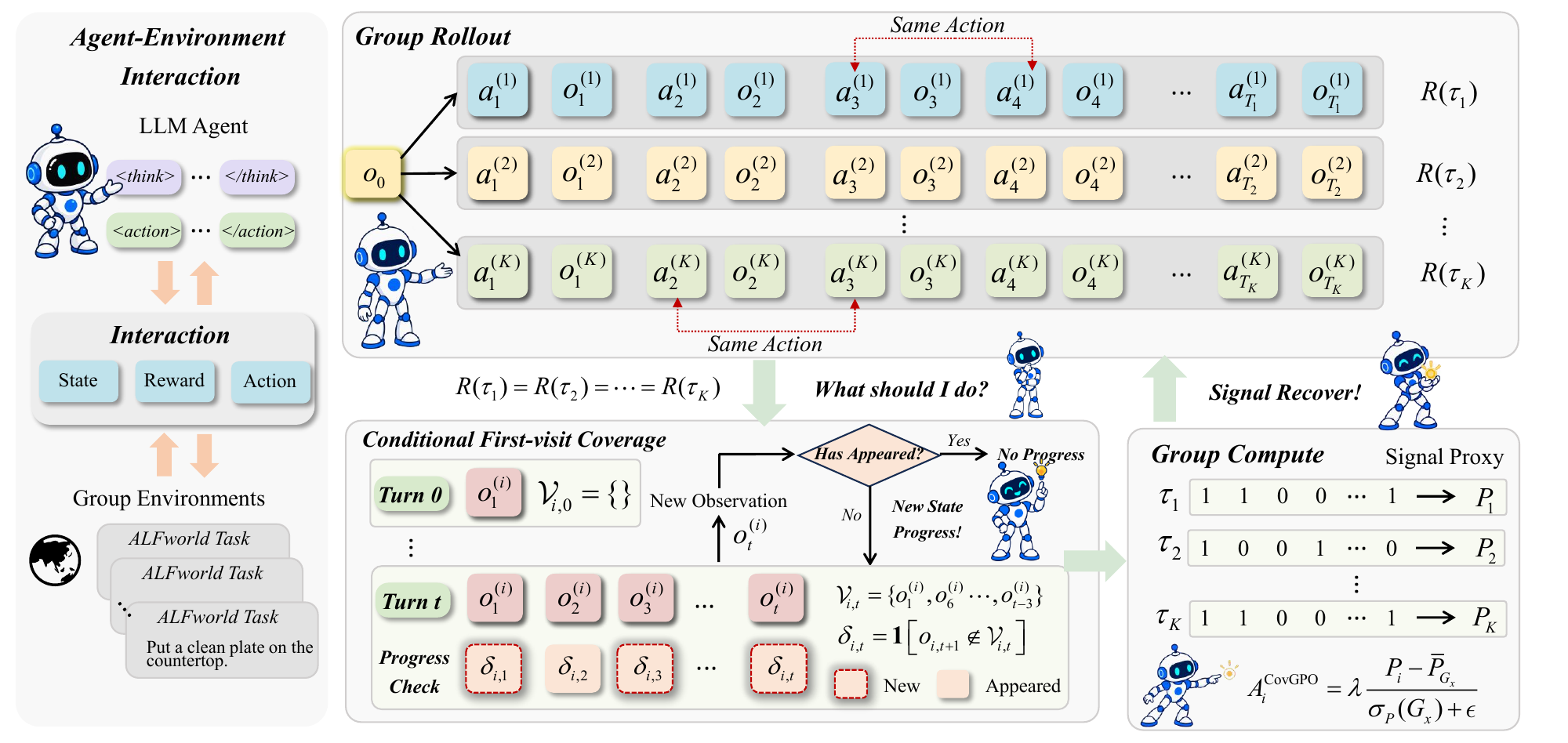}
    \vspace{-8mm}
    \caption{Overview of CovGPO. For each task, the agent samples a group of $K$ rollouts from the same initial state and prompt. 
    When the group is all-failed, CovGPO scores each trajectory by its first-visit observation coverage, and normalizes the scores within the group into a fallback advantage. 
    }
    \label{fig:pipeline}
    \vspace{-4mm}
\end{figure}

A simple insight behind our method is that \textbf{\textit{the behavioral contrast among failed trajectories is observable and can serve as a fallback learning signal when the reward degenerates}}. 
We first analyze the impact of this degeneracy (Section~\ref{sec:escape}), then operationalize the repair through two components, illustrated in Figure~\ref{fig:pipeline}: a \emph{first-visit coverage score} (Section~\ref{sec:coverage}) that quantifies the contrast, and a \emph{conditional switch} (Section~\ref{sec:conditional}) that uses this score only when the reward-based algorithm has degenerated. 
All proofs are deferred to Appendix~\ref{app:proofs}.

\subsection{Why the Trap Is Exact}
\label{sec:escape}

Equation~\ref{eq:degenerate} shows that all-failed groups yield identically zero advantages, where within-group credit signals do not merely weaken under sparse rewards but vanish entirely.

\begin{proposition}[Exact degeneracy and fixed-policy sampling]
\label{prop:escape}
Let $\mathcal{C}_H$ be the depth-$H$ chain: states $s_0,\ldots,s_H$, actions $a^{+}\!:s_j\mapsto s_{j+1}$ and $a^{-}\!:s_j\mapsto s_j$, horizon $T=H$, reward $R(\tau)=\mathbf{1}[s_H\text{ reached}]$, and initial policy $\pi_{\theta_0}=\pi_{\mathrm{ref}}$ with $\pi_{\theta_0}(a^{+}\mid s_j)=\varepsilon$ for all $j$. Whenever no rollout in a group reaches $s_H$, the reward advantages in Eqs.~\ref{eq:grpo_advantage}--\ref{eq:step_advantage} are zero at every comparison granularity. If groups of $K$ independent rollouts are sampled independently from the fixed policy $\pi_{\theta_0}$, the expected number of groups until the first successful rollout is $[1-(1-\varepsilon^H)^K]^{-1}\ge\varepsilon^{-H}/K$.
\end{proposition}

The zero is exact, not merely small: the group mean under a uniform outcome equals the observed return exactly, so the within-group reward signal vanishes identically. This identifies a missing source of signal, even though updates from other groups can still improve the policy.

\textbf{The reward-only comparison leaves behavioral differences within all-failed groups unused.} These already collected groups offer an opportunity to recover a signal without additional rollouts.
\textit{First,} recovering credit from \emph{already collected} all-failed groups requires modifying the signal itself.  Reshaping sampling distributions only affects future rollouts (Appendix~\ref{app:band_sampling}), leaving every member in the current degenerate group with zero credit.
\textit{Second,} the injected signal must distinguish behavior within the failed group. Restoring variation alone need not provide useful exploration, which motivates the coverage score below and the signal-substitution experiments in Section~\ref{sec:analysis}.

\subsection{First-Visit Coverage Score}
\label{sec:coverage}

For the $t$-th transition in trajectory $\tau_i$, we define
\begin{equation}
\delta_{i,t}
=
\mathbf{1}\!\left[o_{i,t+1}\notin \mathcal{V}_{i,t}\right],
\qquad
P_i=\frac{1}{T_i}\sum_{t=1}^{T_i}\delta_{i,t},
\label{eq:step_coverage}
\end{equation}
where $\mathbf{1}[\cdot]$ is the indicator function, $\mathcal{V}_{i,t}=\{o_{i,1},\ldots,o_{i,t}\}$ collects the observations already encountered in $\tau_i$ up to step $t$, and $T_i$ is the length of $\tau_i$, so that $0\le P_i\le1$. A step gets credit only when its resulting observation is new to the trajectory, which filters both low-effect transitions (the observation is unchanged) and cyclic revisitations (the observation returns to a seen state); a per-trajectory hash set adds $O(1)$ expected overhead per step. 

Our contribution is not the score itself but the algorithm built around it: $P$ enters the advantage only on groups where the reward signal has degenerated.
\textit{First}, $P(\tau)=(D(\tau)-1)/T$, where $D(\tau)$ counts distinct observations (Lemma~\ref{lem:relaxation}): at equal length $P$ orders trajectories exactly by observation coverage, and repetitions and cycles earn no additional credit. For unequal lengths, $P$ compares coverage rates instead of distinct-observation counts. 
\textit{Second}, for binary outcomes, coverage certifies \emph{necessity, not proximity}: writing $m_x$ for the fewest distinct observations visited by any successful length-$T$ trajectory, $P(\tau)<(m_x-1)/T$ implies $R(\tau)=0$. This necessary condition does not determine which failed trajectory is closer to success. We use $P$ as an exploration-oriented ordering and assess its usefulness empirically.

That distance bounds the work still required, not the work still expected (Appendix~\ref{app:deficit_bound}): a trajectory can satisfy the coverage requirement while standing in the wrong place holding the wrong object. 
We test this stronger hypothesis directly in Section~\ref{sec:analysis} by resuming failed      rollouts from their exact terminal states, and find only a small, statistically insignificant   difference in favour of the higher-coverage trajectory (Appendix~\ref{app:proof_improvement} formalizes what would follow if a genuine ordering existed).
In any case, $P$ acts only on all-failed groups: since it tracks state coverage instead of task value, the conditional switch below keeps it from competing with the reward once outcome contrast returns.

\subsection{Conditional Advantage Algorithm}
\label{sec:conditional}
An always-on coverage signal may conflict with the task reward once outcome variation is available: a short, efficient success can score lower on coverage than a long, wandering failure (Appendix~\ref{app:conditional_reason}), so the auxiliary term partially offsets the reward-induced gradient~(Tab.~\ref{tab:ablation_combined}). Conversely, discarding coverage in all-failed groups wastes the only within-group contrast available. CovGPO therefore switches \emph{conditionally}, on the observable within-group variance:
\begin{equation}
A_i^{\mathrm{CovGPO}}
=
\begin{cases}
\lambda\,
\dfrac{P_i-\bar{P}_{G_x}}{\sigma_P(G_x)+\epsilon},
& \text{if } \bar{R}_{G_x}=0 \ \text{ and }\ \sigma_P(G_x)\ge \tau_P, \\[10pt]
A_i^{R}
\;=\;\dfrac{R_i-\bar{R}_{G_x}}{\sigma_R(G_x)+\epsilon},
& \text{otherwise},
\end{cases}
\label{eq:CovGPO}
\end{equation}
where $\bar{P}_{G_x}$ and $\sigma_P(G_x)$ mirror $\bar{R}_{G_x}$ and $\sigma_R(G_x)$ in Eq.~\ref{eq:grpo_advantage}, and $\lambda>0$ scales the fallback ceiling. 
The trigger consists of a main condition and a safety check. 
The main condition holds exactly when every trajectory in the group receives zero reward, i.e., when the base advantage is identically zero (Eq.~\ref{eq:degenerate}): this matches all-failed groups in ALFWorld and, under WebShop's fractional reward, further requires no partial match. The safety check $\sigma_P(G_x)\ge\tau_P$ disables the fallback when coverage is identical, avoiding division by near-zero variance.

For trajectory-level base algorithms, $P_i$ directly scores each rollout. 
For step-level baselines (e.g., GiGPO and HGPO), constituent steps inherit their parent trajectory's score (Appendix~\ref{app:deployed_scaling}).
In both cases, the fallback injects a coverage signal if and only if the base reward advantage vanishes, leaving informative groups completely intact. 

\begin{proposition}[Non-interference]
\label{prop:noninterference}
For any policy $\pi_\theta$, any threshold $\tau_P>0$, and any group $G_x$: if $G_x$ does not satisfy the fallback condition of Eq.~\ref{eq:CovGPO}, then $A_i^{\mathrm{CovGPO}}=A_i^{R}$ for all $i$; if it does, then $A_i^{R}=0$ for all $i$. Consequently, every loss sample that depends on no triggered comparison keeps both its value and its gradient (Appendix~\ref{app:proof_noninterference}).
\end{proposition}

CovGPO is therefore a \emph{strict extension} of the base algorithm, modifying the update only where the base method provably provides no signal, at any grouping granularity. The statement concerns the algorithm at a fixed $\theta$; because the two runs subsequently visit different data, it does not claim that the two \emph{training trajectories} agree on reward-informative groups.


Within each triggered fallback group, the update applies a group-mean baselined step on coverage $P$ among failed trajectories. Although Lemma~\ref{lem:surrogate} establishes the $(K-1)/K$ relation for the idealized, unclipped statistic before dispersion filtering, the deployed update serves as a regularized surrogate instead of an exact gradient identity. 
Crucially, the fallback update satisfies two analytical properties. The coverage score-function step is bounded as shown in Proposition~\ref{prop:selflimiting} and evaluates to zero whenever coverage dispersion drops below the gating threshold. Furthermore, for $K$ i.i.d.\ trajectory rollouts, the trigger probability is at most $(1 - p_+(x))^K$ with $p_+(x) = \mathbb{P}(R > 0 \mid x)$ (Remark~\ref{rem:disengagement}).

Finally, CovGPO integrates into any group-based framework by replacing the reward-only advantage with $A_i^{\mathrm{CovGPO}}$ in a clipped surrogate objective:
\begin{equation}
\begin{split}
\mathcal{J}_{\mathrm{CovGPO}}(\theta)
&=
\mathbb{E}_{x,G_x}
\Bigl[
\frac{1}{K}\sum_{i=1}^{K}
\frac{1}{T_i}\sum_{t=1}^{T_i}
\min\!\bigl(
\rho_{i,t}(\theta)\, A_i^{\mathrm{CovGPO}},
\mathrm{clip}\!\bigl(\rho_{i,t}(\theta),\,1{-}\epsilon_c,\,1{+}\epsilon_c\bigr) A_i^{\mathrm{CovGPO}}
\bigr)
\Bigr]\\
&\qquad\qquad
-
\beta\, D_{\mathrm{KL}}(\pi_\theta\|\pi_{\mathrm{ref}}),
\end{split}
\label{eq:CovGPO_loss}
\end{equation}
where $\rho_{i,t}(\theta)=\pi_\theta(a_{i,t}\mid h_{i,t})/\pi_{\theta_{\mathrm{old}}}(a_{i,t}\mid h_{i,t})$ is the importance ratio, $\epsilon_c$ is the clipping range, and $\beta$ controls KL regularization.

\section{Experiments}
\label{sec:experiments}

\subsection{Experiment Setup}

\textbf{Benchmarks.}~~
We train and evaluate LLM agents on two challenging benchmarks: ALFWorld~\citep{shridhar2020alfworld} and WebShop~\citep{yao2022webshop}.
\textit{ALFWorld} is an embodied text environment for multi-step decision-making. In each episode, the agent receives a textual goal and must accomplish it through multi-turn interaction spanning 20--50 steps, receiving only a binary outcome reward upon termination. It covers six categories of household activities: Pick \& Place (Pick), Examine in Light (Look), Clean \& Place (Clean), Heat \& Place (Heat), Cool \& Place (Cool), and Pick Two \& Place (Pick2). 
\textit{WebShop} is a web-based interactive shopping environment in which the agent must search, navigate, and purchase a suitable item on a simulated HTML website. 
\textit{SokoBan} is a visual puzzle environment where a vision-language agent observes RGB images and must plan directional pushes to navigate boxes to target locations without entering irreversible deadlocks.

\textbf{Baselines.}~~
We compare CovGPO against three groups of methods.
(1) \textit{Prompting agents}: the base Qwen2.5 models, React~\citep{yao2022react} and Reflexion~\citep{shinn2023reflexion}, which guide multi-step behavior through in-context prompting without parameter updates.
(2) \textit{RL training methods}: PPO~\citep{schulman2017proximal}, which requires an additional critic network; 
RLOO~\citep{kool2019buy}, a critic-free REINFORCE variant with a leave-one-out baseline; 
DAPO-MT, which adapts DAPO~\citep{yu2026dapo} to multi-turn rollouts by resampling until reward variance emerges—contrasting with our reuse of existing transitions; 
and the group-based methods GRPO~\citep{shao2024deepseekmath}, GiGPO~\citep{feng2026group}, and HGPO~\citep{he2026hierarchy}, where the latter is the state-of-the-art group-based algorithm for LLM agents and serves as our primary base algorithm.
(3) \textit{CovGPO-augmented methods}: since CovGPO modifies only the advantage computation, we instantiate it on top of three different base algorithms, GRPO+CovGPO, GiGPO+CovGPO and HGPO+CovGPO.
Full training settings are provided in Appendix~\ref{app:implementation}.

\textbf{Training details.}~~
We use Qwen2.5-1.5B/7B-Instruct, rollout groups of size $K{=}8$, and $150$ training epochs. Across environments, model scales, and base algorithms, we fix the configured fallback ceiling at $\lambda_{\mathrm{aux}}{=}0.3$.
We report success rate (\%) on held-out test tasks; we report mean\textsubscript{$\pm$std} with three random seeds. More details are provided in Appendix~\ref{app:implementation}.

\begin{table}[t]
\centering
\caption{
Test performance on ALFWorld and WebShop. For ALFWorld, we report the average success rate (\%) for each subtask as well as the overall result; for WebShop, we report the average task score and the average success rate (\%). Most results are averaged over 3 random seeds; the 1.5B GiGPO+CovGPO ALFWorld entry is the five-random-seed average of Table~\ref{tab:lambda_full} ($\lambda_{\mathrm{aux}}{=}0.3$). Seed protocols are specified in Appendix~\ref{app:evaluation_protocol}. \gain{}\,/\,\drop{} give the change on the aggregate metrics relative to that base algorithm. \textbf{Bold} marks the best result within each model scale.
}
\label{tab:main}

\small
{%
\setlength{\tabcolsep}{1.6mm}

\resizebox{\textwidth}{!}{%
\begin{tabular}{@{}l:*{7}{c}:*{2}{c}@{}}
\toprule

\multicolumn{1}{@{}l:}{
  \multirow{2}{*}{\textbf{Method}}
}
& \multicolumn{7}{c:}{\textbf{ALFWorld}}
& \multicolumn{2}{c@{}}{\textbf{WebShop}} \\

\cmidrule(lr){2-8}
\cmidrule(lr){9-10}

\multicolumn{1}{@{}l:}{}
& Pick & Look & Clean & Heat & Cool & Pick2 & All
& Score & Succ. \\

\midrule
\modelscale{Closed-Source Models}

GPT-4o
& 75.3
& 60.8
& 31.2
& 56.7
& 21.6
& 49.8
& 48.0
& 31.8
& 23.7 \\

Gemini-2.5-Pro
& 92.8
& 63.3
& 62.1
& 69.0
& 26.6
& 58.7
& 60.3
& 42.5
& 35.9 \\

\midrule
\modelscale{Qwen2.5-1.5B-Instruct}

Qwen2.5
& 5.9
& 5.5
& 3.3
& 9.7
& 4.2
& 0.0
& 4.1
& 23.1
& 5.2 \\

ReAct
& 17.4
& 20.5
& 15.7
& 6.2
& 7.7
& 2.0
& 12.8
& 40.1
& 11.3 \\

Reflexion
& 35.3
& 22.2
& 21.7
& 13.6
& 19.4
& 3.7
& 21.8
& 55.8
& 21.9 \\

PPO
& 64.8\std{3.5}
& 40.5\std{6.9}
& 57.1\std{4.9}
& 60.6\std{6.6}
& 46.4\std{4.0}
& 47.4\std{1.9}
& 54.4\std{3.1}
& 73.8\std{3.0}
& 51.5\std{2.9} \\

RLOO
& 88.3\std{3.0}
& 52.8\std{8.6}
& 71.0\std{5.9}
& 62.8\std{8.7}
& 66.4\std{5.5}
& 56.9\std{4.7}
& 69.7\std{2.5}
& 73.9\std{5.6}
& 52.1\std{6.7} \\

\dashmidrule

DAPO-MT
& 94.6\std{1.9}
& 75.6\std{10.2}
& 83.4\std{4.1}
& 85.3\std{3.1}
& 86.4\std{7.2}
& 78.1\std{4.9}
& 84.9\std{0.9} 
& 75.0\std{1.2} & 62.9\std{0.4} \\

GRPO
& 85.3\std{1.5}
& 53.7\std{8.0}
& 84.5\std{6.8}
& 78.2\std{7.9}
& 59.7\std{5.0}
& 53.5\std{5.6}
& 72.8\std{3.6}
& 75.8\std{3.5}
& 56.8\std{3.8} \\

\rowcolor{oursrow}
\textbf{GRPO + CovGPO}
& 84.1\std{5.6}
& 76.3\std{15.1}
& 74.8\std{8.1}
& 69.6\std{9.2}
& 83.2\std{4.4}
& 50.8\std{11.3}
& 74.5\std{1.6}\gain{1.7}
& 76.2\std{1.0}\gain{0.4}
& 60.9\std{2.2}\gain{4.1} \\

GiGPO
& 94.4\std{5.9}
& 67.5\std{4.6}
& 94.8\std{3.8}
& 94.4\std{7.8}
& 79.8\std{4.7}
& 76.4\std{5.4}
& 86.7\std{1.7}
& 83.1\std{1.6}
& 65.0\std{3.2} \\

\rowcolor{oursrow}
\textbf{GiGPO + CovGPO}
& 94.9\std{3.9}
& 74.5\std{12.3}
& \textbf{98.4}\std{2.0}
& 95.9\std{5.4}
& 90.6\std{4.3}
& 90.9\std{3.5}
& 92.2\std{2.0}\gain{5.5}
& 85.6\std{0.5}\gain{2.5}
& 72.4\std{0.6}\gain{7.4} \\

HGPO
& 97.8\std{0.3}
& 85.4\std{5.9}
& 95.0\std{3.1}
& \textbf{97.4}\std{1.7}
& 87.6\std{2.6}
& 88.0\std{6.0}
& 92.8\std{1.1}
& 85.6\std{2.9}
& 71.5\std{4.0} \\

\rowcolor{oursrow}
\textbf{HGPO + CovGPO}
& \textbf{100.0}\std{0.0}
& \textbf{91.3}\std{0.5}
& 98.3\std{2.4}
& 97.1\std{4.2}
& \textbf{92.3}\std{2.0}
& \textbf{93.5}\std{2.5}
& \textbf{95.7}\std{0.6}\gain{2.9}
& \textbf{87.5}\std{1.6}\gain{1.9}
& \textbf{75.7}\std{2.8}\gain{4.2} \\

\midrule
\modelscale{Qwen2.5-7B-Instruct}

Qwen2.5
& 33.4
& 21.6
& 19.3
& 6.9
& 2.8
& 3.2
& 14.8
& 26.4
& 7.8 \\

ReAct
& 48.5
& 35.4
& 34.3
& 13.2
& 18.2
& 17.6
& 31.2
& 46.2
& 19.5 \\

Reflexion
& 62.0
& 41.6
& 44.9
& 30.9
& 36.3
& 23.8
& 42.7
& 58.1
& 28.8 \\

PPO
& 92.3\std{4.0}
& 64.0\std{8.4}
& 92.5\std{2.4}
& 89.5\std{7.0}
& 80.3\std{2.0}
& 68.8\std{8.3}
& 80.4\std{2.7}
& 81.4\std{3.1}
& 68.7\std{5.1} \\

RLOO
& 87.6\std{4.3}
& 78.2\std{8.3}
& 87.3\std{5.8}
& 81.3\std{7.6}
& 71.9\std{5.2}
& 48.9\std{8.4}
& 75.5\std{4.6}
& 80.3\std{3.2}
& 65.7\std{4.0} \\

\dashmidrule

DAPO-MT
& 98.9\std{1.9}
& 83.3\std{5.8}
& 95.2\std{1.7}
& 90.8\std{3.0}
& 79.1\std{9.2}
& 87.0\std{1.0}
& 90.4\std{1.9}
& 83.0\std{1.5} 
& 74.6\std{2.5} \\
GRPO
& 90.8\std{5.1}
& 66.1\std{6.7}
& 89.3\std{5.4}
& 74.7\std{6.9}
& 72.5\std{5.4}
& 64.7\std{7.3}
& 77.6\std{5.2}
& 79.3\std{2.8}
& 66.1\std{3.7} \\

\rowcolor{oursrow}
\textbf{GRPO + CovGPO}
& 90.0\std{4.1}
& 67.5\std{8.4}
& 88.4\std{2.3}
& 91.2\std{12.5}
& 76.2\std{1.6}
& 58.4\std{18.4}
& 79.7\std{3.3}\gain{2.1}
& 83.3\std{3.6}\gain{4.0}
& 68.2\std{4.3}\gain{2.1} \\

GiGPO
& 97.7\std{1.6}
& 82.7\std{7.9}
& 98.8\std{1.6}
& 83.7\std{7.2}
& \textbf{89.3}\std{8.2}
& 79.2\std{6.6}
& 90.8\std{1.3}
& 84.4\std{2.9}
& 72.8\std{3.2} \\

\rowcolor{oursrow}
\textbf{GiGPO + CovGPO}
& \textbf{100.0}\std{0.0}
& \textbf{97.2}\std{4.8}
& 91.8\std{0.4}
& 83.4\std{3.0}
& 87.1\std{3.3}
& 91.0\std{4.7}
& 92.5\std{0.9}\gain{1.7}
& 85.9\std{0.2}\gain{1.5}
& 76.2\std{1.1}\gain{3.4} \\

HGPO
& 98.9\std{1.2}
& 88.8\std{1.6}
& \textbf{98.9}\std{0.4}
& \textbf{95.3}\std{6.0}
& 86.8\std{1.3}
& \textbf{98.1}\std{1.2}
& 95.4\std{0.6}
& \textbf{89.0}\std{1.0}
& 78.5\std{1.4} \\

\rowcolor{oursrow}
\textbf{HGPO + CovGPO}
& \textbf{100.0}\std{0.0}
& 95.5\std{6.4}
& 96.7\std{0.0}
& 92.5\std{2.3}
& 88.0\std{4.2}
& 97.6\std{3.4}
& \textbf{95.7}\std{0.6}\gain{0.3}
& \textbf{89.0}\std{1.6}
& \textbf{79.7}\std{1.4}\gain{1.2} \\

\bottomrule
\end{tabular}%
}
}
\vspace{-3mm}
\end{table}

\begin{table*}[t]
\centering
\caption{
Ablation on conditional gating ($w/o$ Conditional $=$ always-on coverage) on ALFWorld and WebShop (Qwen2.5-1.5B, GRPO and GiGPO bases, 150 epochs). \textbf{Bold} marks the best per column.
}
\label{tab:ablation_combined}

\small
{%
\setlength{\tabcolsep}{1.7mm}

\resizebox{\textwidth}{!}{%
\begin{tabular}{@{}l:*{7}{c}:*{2}{c}@{}}
\toprule

\multicolumn{1}{@{}l:}{
  \multirow{2}{*}{\textbf{Method}}
}
& \multicolumn{7}{c:}{\textbf{ALFWorld}}
& \multicolumn{2}{c@{}}{\textbf{WebShop}} \\

\cmidrule(lr){2-8}
\cmidrule(lr){9-10}

\multicolumn{1}{@{}l:}{}
& Pick
& Look
& Clean
& Heat
& Cool
& Pick2
& All
& Score
& Succ. \\

\midrule

GRPO
& \textbf{85.3}\std{1.5}
& 53.7\std{8.0}
& \textbf{84.5}\std{6.8}
& \textbf{78.2}\std{7.9}
& 59.7\std{5.0}
& \textbf{53.5}\std{5.6}
& 72.8\std{3.6}
& 75.8\std{3.5}
& 56.8\std{3.8} \\

\quad Always-on coverage
& 77.5\std{7.4}
& 63.1\std{12.4}
& 78.3\std{5.0}
& 77.6\std{13.5}
& 69.3\std{11.1}
& 47.4\std{10.3}
& 68.3\std{2.7}
& 73.0\std{4.7}
& 58.6\std{4.1} \\

\rowcolor{oursrow}
\textbf{\quad CovGPO (conditional)}
& 84.1\std{5.6}
& \textbf{76.3}\std{15.1}
& 74.8\std{8.1}
& 69.6\std{9.2}
& \textbf{83.2}\std{4.4}
& 50.8\std{11.3}
& \textbf{74.5}\std{1.6}
& \textbf{76.2}\std{1.0}
& \textbf{60.9}\std{2.2} \\

\dashmidrule

GiGPO
& 94.4\std{5.9}
& 67.5\std{4.6}
& 94.8\std{3.8}
& 94.4\std{7.8}
& 79.8\std{4.7}
& 76.4\std{5.4}
& 86.7\std{1.7}
& 83.1\std{1.6}
& 65.0\std{3.2} \\

\quad Always-on coverage
& \textbf{99.2}\std{1.6}
& \textbf{78.9}\std{12.1}
& 95.7\std{4.0}
& \textbf{98.8}\std{2.4}
& 81.9\std{8.6}
& 87.4\std{1.3}
& 90.9\std{2.5}
& 78.5\std{3.2}
& 68.8\std{2.9} \\

\rowcolor{oursrow}
\textbf{\quad CovGPO (conditional)}
& 94.9\std{3.9}
& 74.5\std{12.3}
& \textbf{98.4}\std{2.0}
& 95.9\std{5.4}
& \textbf{90.6}\std{4.3}
& \textbf{90.9}\std{3.5}
& \textbf{92.2}\std{2.0}
& \textbf{85.6}\std{0.5}
& \textbf{72.4}\std{0.6} \\

\bottomrule
\end{tabular}%
}
}

\vspace{-4mm}
\end{table*}

\subsection{Main Results}
Table~\ref{tab:main} summarizes the main results. Prompting-only agents, including strong closed-source models, still struggle on these long-horizon tasks, while RL training improves substantially over prompting. We highlight three observations:
\noindent\textbf{(i) CovGPO improves group-based RL across scales and benchmarks.}
At 1.5B it lifts GRPO/GiGPO/HGPO on ALFWorld ($72.8{\to}74.5$, $86.7{\to}92.2$, $92.8{\to}95.7$) and on WebShop success ($56.8{\to}60.9$, $65.0{\to}72.4$, $71.5{\to}75.7$); the same pattern holds at 7B, where GRPO and GiGPO rise on ALFWorld ($77.6{\to}79.7$, $90.8{\to}92.5$) and every base improves on WebShop success ($66.1{\to}68.2$, $72.8{\to}76.2$, $78.5{\to}79.7$). 
The only saturated setting is 7B HGPO, where CovGPO behaves as predicted: ALFWorld changes from $95.4$ to $95.7$ ($\pm0.6$), WebShop score remains $89.0$, and success gains $1.2, 78,5{\to}79.7$, all within seed variation. In this setting, the base policy rarely produces all-failed groups, so the fallback is seldom activated; on the remaining nondegenerate groups, CovGPO exactly recovers the base advantage by Proposition~\ref{prop:noninterference}.
\noindent\textbf{(ii) CovGPO's gains concentrate on hard subtasks.}
Improvements are largest where the base policy almost never succeeds and all-failed groups dominate: at 1.5B, GiGPO+CovGPO lifts \textit{Look} $67.5{\to}74.5$, \textit{Cool} $79.8{\to}90.6$, \textit{Pick2} $76.4{\to}90.9$, while the easy \textit{Pick} is saturated. This suggests that failed trajectories are not equally wrong, and CovGPO recovers the hidden contrast precisely in the all-failed regime.
\noindent\textbf{(iii) CovGPO achieves sample-efficient gains without rollout overhead.}
CovGPO reuses existing trajectories to resolve all-failed groups with zero extra environment interactions. In contrast, while DAPO-MT improves performance by repeatedly resampling rollouts until reward variance emerges, it incurs prohibitive generation overhead ($10.5$ h at 1.5B and $43.6$ h at 7B).
\noindent\textbf{(iv) CovGPO is compatible with both trajectory- and step-level algorithms.}
Adding CovGPO increases aggregate ALFWorld success for GRPO, GiGPO, and HGPO at 1.5B ($72.8{\to}74.5$, $86.7{\to}92.2$, and $92.8{\to}95.7$). This supports portability across grouping granularities.
\noindent\textbf{(vi) Extension to visual environments.}
\label{sec:sokoban}
CovGPO also extends to multimodal settings: an exploratory visual study on SokoBan (Table~\ref{tab:sokoban}) improves GRPO, GiGPO, and HGPO by $+6.3\%$, $+2.0\%$, and $+3.2\%$, respectively (details in Appendix~\ref{app:sokoban_protocol}).

\noindent\textbf{Ablation Study.}\label{sec:ablation}
Always-on coverage ($w/o$ Conditional) forfeits the non-interference guarantee and underperforms conditional CovGPO on average. The gap is largest for weaker base algorithms: it falls below the base algorithm itself on GRPO--ALFWorld ($68.3$ vs.\ $72.8$), stays within seed variation of conditional CovGPO on GiGPO--ALFWorld, where the base is already strong ($90.9\pm2.5$ vs.\ $92.2\pm2.0$), and is lower on both metrics on GiGPO--WebShop ($78.5/68.8$ vs.\ $85.6/72.4$). Gating is thus necessary for the exact guarantee .(Appendix~\ref{app:conditional_reason}).

\begin{figure}[t]
    \includegraphics[width=1.0\linewidth]{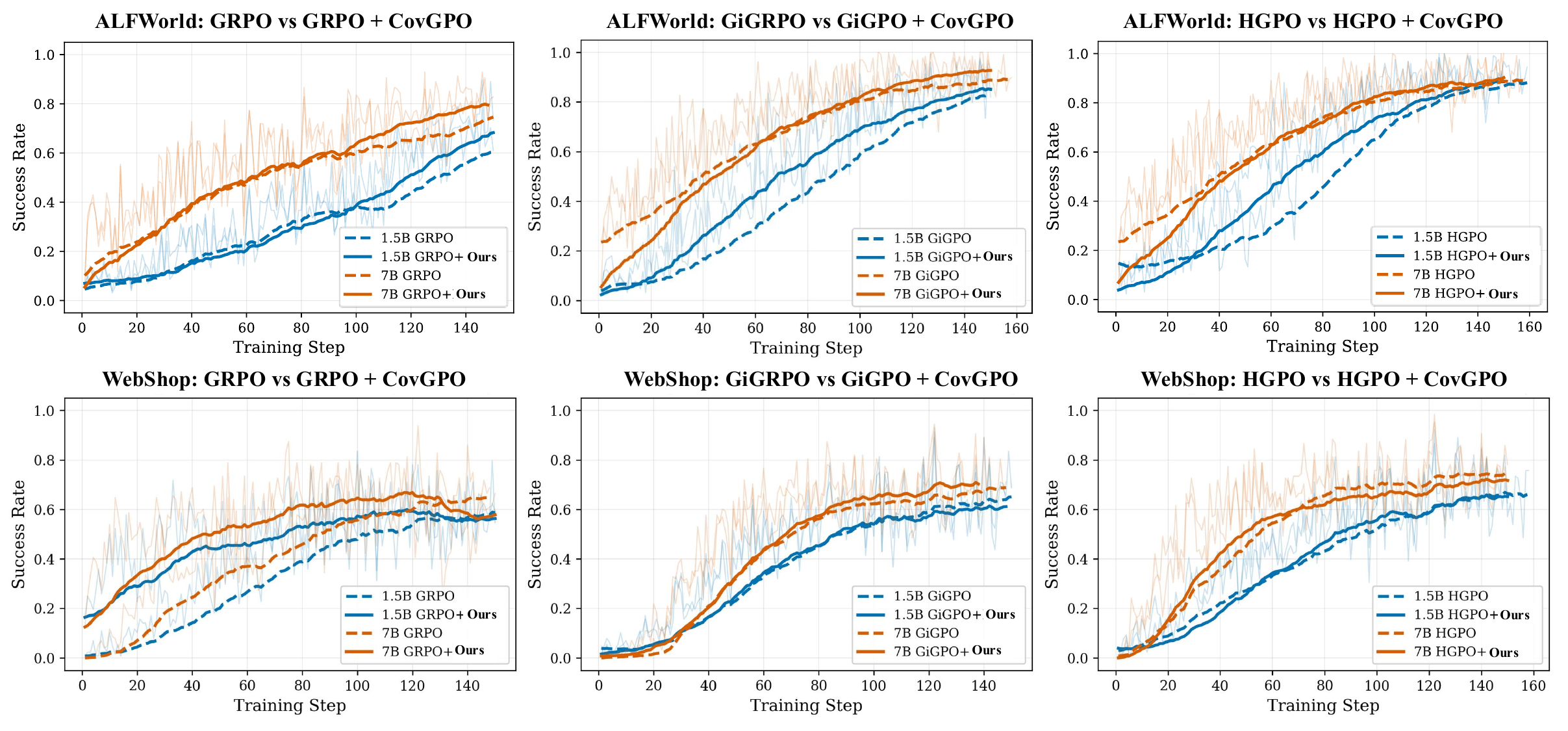}
   \vspace{-0.1in}
   \caption{Training dynamics of CovGPO and its baseline on ALFWorld and WebShop. The lighter curves show the original curves, while the darker curves correspond to exponential moving average (EMA) smoothing with decay $\alpha=0.95$, highlighting the overall training trends.}
   \label{fig:dynamics}
   \vspace{-3mm}
\end{figure}

\noindent\textbf{Training dynamics.}~\label{sec:dynamics}
Figure~\ref{fig:dynamics} plots success rate against training epochs, comparing each CovGPO-augmented method with its baseline under matched rollouts, hyperparameters, and seeds. On 1.5B ALFWorld, GiGPO+CovGPO reaches $49.2\%$ success at epoch $60$ versus GiGPO's $32.8\%$, and finishes at $91.4\%$ in this single-seed trace (Appendix~\ref{app:per_category}). From epochs $10$ to $150$, its zero-score step-group fraction falls from $92.0\%$ to $43.6\%$, while fallback use falls from $29.4\%$ to $4.7\%$ (Table~\ref{tab:all_fail_curve}). Intervention thus decreases without an epoch schedule; the rates differ because only coverage-diverse zero-score groups trigger the fallback.

\begin{table}[t]
\begin{minipage}[t]{0.43\linewidth}
\vspace{0pt}
\caption{Test performance of VLM agents using Qwen2.5-VL-3B-Instruct on the interactive game environment SokoBan.}
\label{tab:sokoban}
\centering

\begin{tabular}{@{}lccc@{}}
\toprule
\textbf{Training}
& \textbf{Vanilla}
& \cellcolor{oursrow}\textbf{+CovGPO}
& \textbf{\(\Delta\)} \\
\midrule

Prompting
& 11.7
& --
& -- \\

GRPO
& 67.1
& \cellcolor{oursrow}\textbf{73.4}\std{3.2}
& \improve{+6.3} \\

GiGPO
& 76.9
& \cellcolor{oursrow}\textbf{78.9}\std{2.7}
& \improve{+2.0} \\

HGPO
& 81.2
& \cellcolor{oursrow}\textbf{84.4}\std{2.5}
& \improve{+3.2} \\
\bottomrule
\end{tabular}

\end{minipage}
\hfill
\vspace{-3pt}
\begin{minipage}[t]{0.55\linewidth}

\centering
\vspace{0in}
\includegraphics[width=0.9\linewidth]{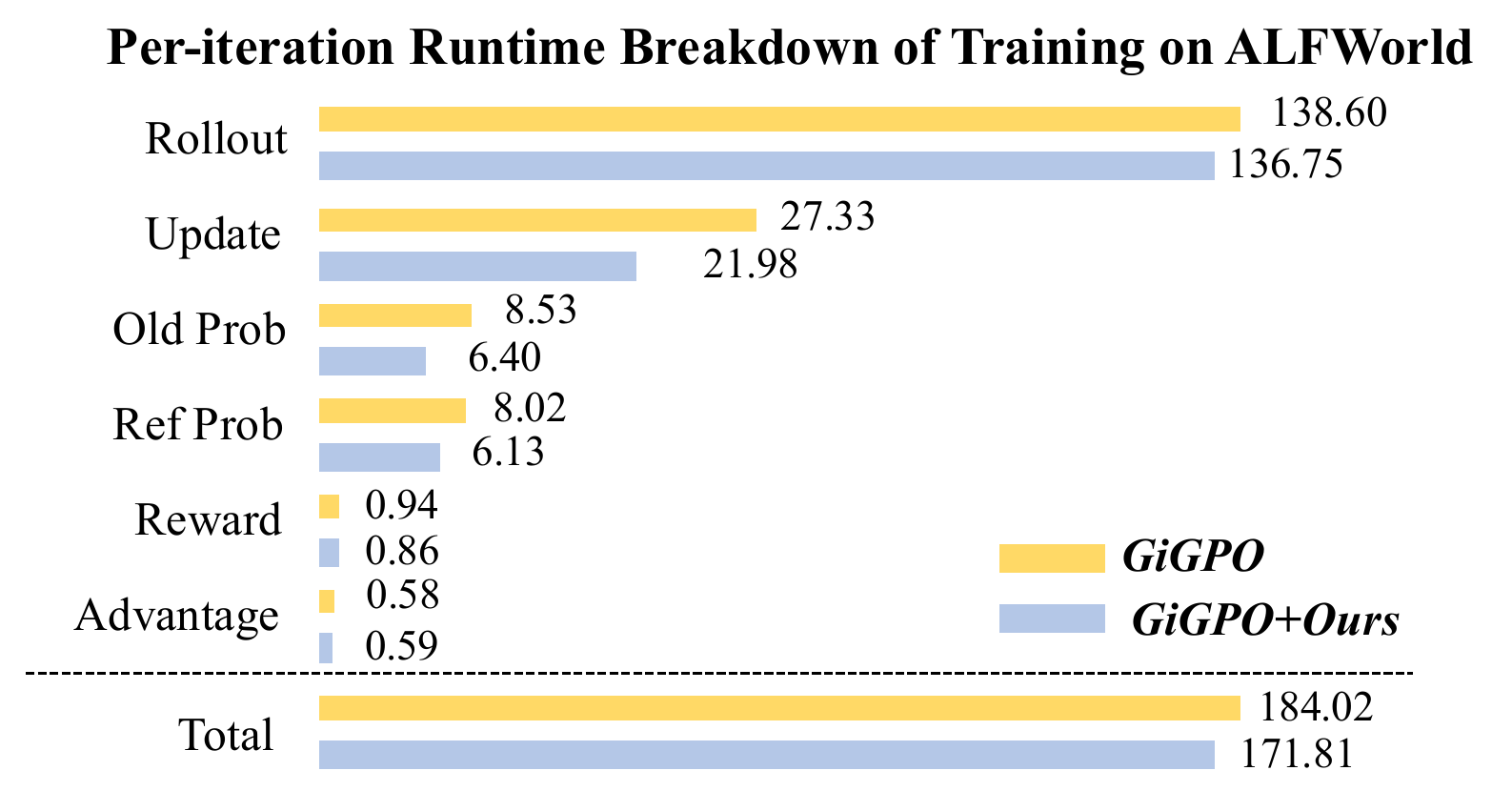}
\vspace{-0.05in}
\captionof{figure}{Training-time(s) delta on four GPUs.}
\label{fig:timing_budget}
\end{minipage}

\end{table}

\noindent\textbf{Computational Budget.}
CovGPO adds an expected $O(1)$ hash-set lookup per step and scalar group statistics, with no learned components. In the timing audit (Figure~\ref{fig:timing_budget}), advantage computation is nearly unchanged ($0.587$ vs.\ $0.585$\,s); total iteration time is $184.0$ vs.\ $171.8$\,s. These measurements show negligible added cost, not an intrinsic speedup.

\subsection{Why CovGPO Works}
\label{sec:analysis}

We examine what the fallback signal contributes, how it changes action preferences, and which design choices matter. The analyses use Qwen2.5-1.5B-Instruct with GiGPO on ALFWorld; Figure~\ref{fig:analysis} organizes the diagnostics, with protocols and corrected statistics in Appendices~\ref{app:audit_metrics} and~\ref{app:signal_breakdown}.

\noindent\textbf{CovGPO benefits from trajectory-aligned coverage credit.} 
We replace coverage with shuffled or random scores in all-failed conditions:
(i) \emph{horizontal controls} operate \emph{without} CovGPO's first-visit coverage signal entirely, replacing both the dispersion gate and fallback advantage ranking with permuted scores (\emph{Shuffle-h}, $74.2\%$) or uniform noise (\emph{Random-h}, $77.3\%$);
(ii) \emph{score-only controls} retain CovGPO's dispersion gate check but operate \emph{without} trajectory-aligned coverage credit, permuting (\emph{Shuffle}, $78.1\%$) or randomizing the injected scores (\emph{Random}, $83.6\%$).
Both underperform GiGPO ($86.7\%$) and CovGPO ($92.2\%$), demonstrating that neither non-zero variance nor dispersion gating alone suffices without trajectory-aligned coverage credit (Appendix~\ref{app:signal_breakdown}).

\noindent\textbf{CovGPO steers policy sampling toward state-changing actions.} Across fixed records, CovGPO consistently increases $\log\pi_{\mathrm{CovGPO}}-\log\pi_{\mathrm{GiGPO}}$ on mid-band state-changing transitions while penalizing low-effect actions. It outperforms \textit{Shuffle-h} by $+0.032$ in oracle correlation and $2.357$ nats in no-op suppression (Appendix~\ref{app:audit_metrics}). Crucially, \textit{Obs-Change} exhibits larger aggregate shifts but lower task success, demonstrating that redistribution magnitude alone does not drive gains. 

\noindent\textbf{CovGPO employs coverage as an exploration heuristic.} First-visit coverage detects novel observations over repetition, but does not index failure quality. Resuming $246$ paired failed rollouts from step $50$ for $25$ steps yields success rates of $4.47\%$ and $3.25\%$ for higher- and lower-coverage trajectories ($+1.22$\,pp). Its partial correlation with oracle subgoal progress is marginal ( Appendix~\ref{app:coverage_semantics}). These diagnostics do not establish coverage as a reliable continuation-value proxy; CovGPO thus uses it solely to guide exploration when reward-based comparisons offer no distinction.

\noindent\textbf{Hyperparameter sensitivity.}
CovGPO's fallback ceiling $\lambda_{\mathrm{aux}}$ is fixed at $0.3$ across environments. Sweeping $\lambda_{\mathrm{aux}}\in\{0.1, 0.3, 0.5, 0.7, 1.0\}$ on ALFWorld (Table~\ref{tab:lambda_full}) shows that $0.3$ achieves $92.2\%$ success, within $2.3$pp of the single-seed peak ($0.1$: $94.5\%$) and well above $0.5$ ($78.3\%$) and $0.7$ ($83.6\%$), confirming $\lambda_{\mathrm{aux}}=0.3$ as a robust default.

\begin{figure}
    \centering
    \includegraphics[width=1\linewidth]{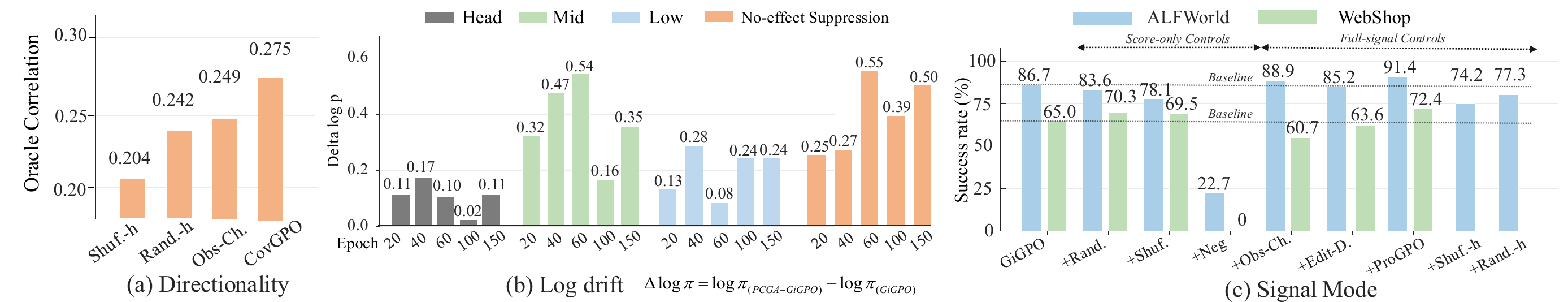}
    \caption{Policy-preference diagnostics and fallback-signal ablations on ALFWorld with Qwen2.5-1.5B-Instruct.
    (a) Correlation between policy log-probability changes and oracle-labeled task progress on fixed action records.
    (b) CovGPO suppresses low-effect actions and shifts probability mass toward under-sampled ones.
    (c) Task performance with different fallback signals.}
    \label{fig:analysis}
    \vspace{-5mm}
\end{figure}

\section{Conclusion}
\label{sec:conclusion}
In this work, we identify the credit trap in group-based RL for multi-turn LLM agents, where uninformative zero-reward rollout groups yield degenerate relative advantages despite exhibiting distinct interaction behaviors. To recover this latent contrast, we propose Coverage-conditioned Group Policy Optimization (CovGPO), a lightweight, model-free framework that ranks trajectories within all-failed comparison groups via \textit{first-visit observation coverage}. Computed directly from existing interactions without extra rollouts, learned reward models, or intermediate supervision, CovGPO seamlessly integrates with trajectory-, state-, and history-level baselines (GRPO, GiGPO, and HGPO). Theoretically, we prove that our gated fallback strictly preserves standard reward-relative policy gradients whenever reward variance exists.
Evaluations across ALFWorld and WebShop show consistent gains across two model scales (Qwen2.5-1.5B/7B) and three base algorithms, suppressing repetitive, state-invariant actions. Crucially, our findings deliver a broader insight for agentic RL: \textbf{\textit{an auxiliary behavioral statistic can provide powerful exploratory credit without serving as a calibrated value estimator or success proxy}}. CovGPO demonstrates that structured failures offer a low-cost, readily accessible source of conditional credit, opening promising avenues for representation-level coverage metrics and adaptive disengagement in long-horizon agent learning.

\bibliographystyle{plainnat}
\bibliography{custom}

\newpage

\begin{center}
{\Large \bf Supplementary Material for ``Learning from All-Failed Groups via First-Visit Coverage in Long-Horizon Agentic Tasks''}
\end{center}
\vspace{1em}

\tableofcontents
\newpage

\clearpage

\section{Proofs and Worked Example}
\label{app:proofs}

This section provides the rigorous theoretical foundations, complete proofs, and illustrative examples for Coverage-conditioned Group Policy Optimization (CovGPO). We formalize the theoretical properties of the coverage metric, establish gradient non-interference guarantees, analyze norm bounds and conditional expectations, and present a concrete step-by-step example on an all-failed rollout group.

\paragraph{Section Roadmap.} To facilitate structured navigation, this section is organized into three major theoretical parts:
\begin{itemize}[leftmargin=*, itemsep=0pt, topsep=2pt, parsep=1pt]
    \item \textbf{Theoretical Foundations of Coverage (Appendix~\ref{app:proof_escape}--\ref{app:proof_improvement}):} We prove exact reward degeneracy on zero-score rollout groups (Appendix~\ref{app:proof_escape}), formalize coverage exactness and necessity thresholds (Appendix~\ref{app:proof_relaxation}), establish how coverage deficit lower-bounds remaining work (Appendix~\ref{app:deficit_bound}), and prove property ordering under coverage-monotone continuations (Appendix~\ref{app:proof_improvement}).
    \item \textbf{Algorithmic Properties and Non-Interference (Appendix~\ref{app:proof_noninterference}--\ref{app:proof_selflimiting}):} We establish exact non-interference preserving reward-relative gradients on informative groups (Appendix~\ref{app:proof_noninterference}), prove the score-function identity under groupwise normalization (Appendix~\ref{app:proof_surrogate}), demonstrate frontier update boundaries via a counterexample (Appendix~\ref{app:proof_polyescape}), and derive static norm bounds for the auxiliary fallback gradient (Appendix~\ref{app:proof_selflimiting}).
    \item \textbf{Disengagement, Gating Rationale, and Instantiation (Appendix~\ref{app:disengagement}--\ref{app:worked_example}):} We derive rollout trigger probability bounds (Appendix~\ref{app:disengagement}), justify the conditional fallback design against reward--progress conflicts (Appendix~\ref{app:conditional_reason}), and walk through a detailed, concrete worked example on a representative all-failed group (Appendix~\ref{app:worked_example}).
\end{itemize}

\paragraph{Common setup.} Fix a task $x$ in a finite-horizon environment. The transition and observation mechanisms, observation equality rule, reward $R(\tau)\ge0$, completion indicator $S_x(\tau)\in\{0,1\}$, and coverage statistic $P(\tau)$ have no explicit dependence on $\theta$; their distributions depend on the policy. Write $\pi_\theta(\tau\mid x)$ for the induced trajectory law, suppressing $x$ below. For score-function calculations, assume a differentiable law with locally common support and integrability permitting differentiation under the integral. Its score is $s_\theta(\tau)=\nabla_\theta\log\pi_\theta(\tau)$, and $s_i=s_\theta(\tau_i)$ for sampled trajectories. Independence is assumed only for the rollout sampling statements that explicitly require it. All empirical standard deviations use the population convention in Eq.~\eqref{eq:app_variance_convention}. Unless specified otherwise, $\epsilon>0$, $\tau_P>0$, and $\lambda>0$. Algorithm and loss comparisons hold sampled data, parameters, component membership, loss weights, reference policy, and regularization fixed, with advantages and gate statistics treated as constants during policy differentiation.

\subsection{Proof of Proposition~\ref{prop:escape}}
\label{app:proof_escape}

\textbf{Setup.} Recall $\mathcal C_H$ from Section~\ref{sec:escape}: $a^+$ advances from $s_j$ to $s_{j+1}$, $a^-$ stays at $s_j$, $s_H$ is absorbing, and $T=H$. A rollout succeeds exactly when all $H$ actions are $a^+$. The fixed policy chooses $a^+$ with probability $\varepsilon\in(0,1)$ at every nonterminal state.

\begin{proof}
\emph{Step 1: exact degeneracy.} If every rollout has reward zero, then $\bar R_{G_x}=\sigma_R(G_x)=0$ and Eq.~\eqref{eq:grpo_advantage} gives
\begin{align}
A_i^R=\frac{0-0}{0+\epsilon}=0.
\end{align}
Every discounted step return is also zero, so centering within an anchor-state or history group gives zero step advantages. For any likelihood ratio $r$ and clipping threshold $\epsilon_c$, the associated surrogate term satisfies
\begin{align}
\min\{r A_i^R,\operatorname{clip}(r,1-\epsilon_c,1+\epsilon_c)A_i^R\}=0.
\end{align}
Thus these zero-advantage sample terms contribute zero reward-relative gradient.

\emph{Step 2: fixed-policy sampling.} A rollout succeeds with probability $\varepsilon^H$. For independent groups of $K$ independent rollouts from this fixed policy, the number $N_{\mathrm{first}}$ of groups up to and including the first group containing a success is geometric with parameter $q=1-(1-\varepsilon^H)^K$. The union bound gives
\begin{align}
q&\le K\varepsilon^H,\\
\mathbb E[N_{\mathrm{first}}]&=\frac{1}{q}
=\frac{1}{1-(1-\varepsilon^H)^K}
\ge\frac{1}{K\varepsilon^H}.
\end{align}
\end{proof}

The waiting-time statement concerns repeated sampling from the specified fixed policy; the degeneracy statement concerns the zero-advantage surrogate terms on the collected data.

\subsection{Coverage Exactness}
\label{app:proof_relaxation}

\begin{lemma}[Coverage identity and fixed-length necessity]
\label{lem:relaxation}
Fix a task $x$ and a length $T\ge1$, and let $\mathcal{T}_{x,T}$ be its length-$T$ trajectories. For $\tau\in\mathcal{T}_{x,T}$ with observations $(o_1,\ldots,o_{T+1})$, including the observation returned by the last transition, let $D(\tau)=|\{o_1,\ldots,o_{T+1}\}|$. Then
\begin{align}
P(\tau)=\frac{D(\tau)-1}{T}.
\end{align}
Consequently, at equal length $P$ orders trajectories exactly as $D$ does, and no-op repetitions and cyclic revisits do not increment the coverage count. This identity requires no successful trajectory. For the necessity statement, assume additionally that $\{\tau\in\mathcal{T}_{x,T}:S_x(\tau)=1\}$ is nonempty and define $m_{x,T}=\min\{D(\tau):\tau\in\mathcal{T}_{x,T},\,S_x(\tau)=1\}\ge1$. Then
\begin{align}
P(\tau)<\frac{m_{x,T}-1}{T}\quad\Longrightarrow\quad S_x(\tau)=0.
\end{align}
\end{lemma}

\begin{proof}
\emph{Setup.}~~Fix the task and length domain of the Lemma and a trajectory $\tau\in\mathcal{T}_{x,T}$ with observation sequence $(o_1,\ldots,o_{T+1})$, including the observation returned by the last transition. Let $\mathcal{O}(\tau)=\{o_1,\ldots,o_{T+1}\}$ denote the set of distinct observations, so that $D(\tau)=|\mathcal{O}(\tau)|$, and let $\mathcal{V}_t=\{o_1,\ldots,o_t\}$ be the prefix set. Recall from Eq.~\eqref{eq:step_coverage} that $\delta_t=\mathbf{1}[o_{t+1}\notin\mathcal{V}_t]$ and $P(\tau)=\frac{1}{T}\sum_{t=1}^{T}\delta_t$. For each $o\in\mathcal{O}(\tau)$, define its first-occurrence index
\begin{align}
\rho(o)\;=\;\min\{s\in\{1,\ldots,T+1\}: o_s=o\}.
\end{align}

\emph{Part (i).}~~By definition of $\mathcal{V}_t$,
\begin{align}
\delta_t=1
\quad\Longleftrightarrow\quad
o_{t+1}\notin\{o_1,\ldots,o_t\}
\quad\Longleftrightarrow\quad
\rho(o_{t+1})=t+1 .
\label{eq:first_occurrence}
\end{align}
Consider the map $\Phi: t\mapsto o_{t+1}$ restricted to the index set $\mathcal{T}_1=\{t\in\{1,\ldots,T\}:\delta_t=1\}$. By Eq.~\eqref{eq:first_occurrence}, $\Phi$ maps $\mathcal{T}_1$ into $\{o\in\mathcal{O}(\tau):\rho(o)\ge 2\}$; it is injective because two indices $t\neq t'$ in $\mathcal{T}_1$ satisfy $\rho(o_{t+1})=t+1\neq t'+1=\rho(o_{t'+1})$, and it is surjective because every $o$ with $\rho(o)=s\ge 2$ is the image of $t=s-1\in\mathcal{T}_1$. Hence $\Phi$ is a bijection, and since $\rho(o_1)=1$ accounts for exactly one element of $\mathcal{O}(\tau)$,
\begin{align}
\sum_{t=1}^{T}\delta_t
\;=\;|\mathcal{T}_1|
\;=\;\bigl|\{o\in\mathcal{O}(\tau):\rho(o)\ge 2\}\bigr|
\;=\;D(\tau)-1,
\qquad\text{i.e.,}\qquad
P(\tau)=\frac{D(\tau)-1}{T}.
\end{align}
A no-op repetition or cyclic revisit has $\delta_t=0$ and does not increment the coverage count. Inserting a revisit into an otherwise unchanged observation sequence preserves $D$ and increases the denominator, so the coverage rate weakly decreases.

\emph{Part (ii).}~~Let $\tau\in\mathcal{T}_{x,T}$ be any trajectory with $S_x(\tau)=1$. By the definition of the coverage requirement $m_{x,T}=\min\{D(\tau'):\tau'\in\mathcal{T}_{x,T},\,S_x(\tau')=1\}$ in Lemma~\ref{lem:relaxation}, $m_{x,T}$ is the minimum of $D$ over successful length-$T$ trajectories, hence $D(\tau)\ge m_{x,T}$. Combining with Part (i),
\begin{align}
P(\tau)\;=\;\frac{D(\tau)-1}{T}\;\ge\;\frac{m_{x,T}-1}{T},
\end{align}
which establishes the containment
\begin{align}
\{\tau\in\mathcal{T}_{x,T}: S_x(\tau)=1\}\;\subseteq\;\Bigl\{\tau\in\mathcal{T}_{x,T}: P(\tau)\ge \tfrac{m_{x,T}-1}{T}\Bigr\},
\end{align}
and, taking complements within $\mathcal{T}_{x,T}$, the equivalent contrapositive form
\begin{align}
P(\tau)<\frac{m_{x,T}-1}{T}\;\Longrightarrow\;S_x(\tau)=0 .
\label{eq:contrapositive}
\end{align}
\end{proof}

\paragraph{Length and completion.} The necessity threshold is specific to the task, observation representation, and length domain. The containment may be strict, since sufficient coverage need not complete the task. With unequal lengths, the rate $P$ need not order raw coverage counts.

\subsection{The Coverage Deficit Bounds the Remaining Work}
\label{app:deficit_bound}

\begin{lemma}[Coverage deficit and successful continuation]
\label{lem:deficit}
Fix a total transition budget $B$ and let $\mathcal S_{x,\le B}$ be the set of all successful task-$x$ trajectories of length at most $B$, assumed nonempty. Define
\begin{align}
m_{x,\le B}&=\min_{\tau'\in\mathcal S_{x,\le B}}D(\tau'),
&d_{x,B}(\tau)&=\max\{m_{x,\le B}-D(\tau),0\}.
\end{align}
Let $\tau$ be a length-$t$ prefix, $0\le t\le B$, admitting a valid, reset-free successful continuation $\eta$ of length $\ell\le B-t$ from its current state and retained history. Then
\begin{align}
\ell\ge d_{x,B}(\tau).
\end{align}
In particular, an irreversibly terminated failure is not an admissible prefix under this hypothesis.
\end{lemma}

\begin{proof}
\emph{Step 1: count newly visited observations.} The concatenation $\tau\mathbin{\circ}\eta$ retains all prefix observations. Each of the $\ell$ additional transitions contributes at most one new observation, hence
\begin{align}
D(\tau\mathbin{\circ}\eta)
&=D(\tau)+|\mathcal O(\tau\mathbin{\circ}\eta)\setminus\mathcal O(\tau)|\\
&\le D(\tau)+\ell.
\end{align}
\emph{Step 2: apply the budget-specific minimum.} Since the concatenation succeeds and has length $t+\ell\le B$,
\begin{align}
m_{x,\le B}&\le D(\tau\mathbin{\circ}\eta)\le D(\tau)+\ell,\\
\ell&\ge\max\{m_{x,\le B}-D(\tau),0\}=d_{x,B}(\tau).
\end{align}
\end{proof}

For equal positive prefix length $t$, the identity $D=1+tP$ yields $d_{x,B}=\max\{m_{x,\le B}-1-tP,0\}$. Thus increasing $P$ weakly decreases this lower bound; the bound itself does not order continuation values or actual remaining work.

\subsection{Ordering under Coverage-Monotone Continuation}
\label{app:proof_improvement}

The following additional hypothesis isolates an ordering property; it is not assumed by the algorithm or implied by the deficit bound.

\begin{assumption}[Coverage-monotone continuation]
\label{asm:monotone}
Fix a task $x$, a policy $\pi_\theta$, and a finite continuation budget $\Delta$. For unsuccessful prefixes that admit reset-free resumption within the specified total horizon, let $V_\Delta(\tau)$ be the probability of completing $x$ within $\Delta$ further steps from the endpoint state and retained history of $\tau$. For equal-length such prefixes, assume $D(\tau)>D(\tau')\Rightarrow V_\Delta(\tau)\ge V_\Delta(\tau')$.
\end{assumption}

\begin{proposition}[Conditional ordering and empirical covariance]
\label{prop:improvement}
Let a triggered rollout group consist of equal-length trajectories satisfying Assumption~\ref{asm:monotone}. Write $V_i=V_\Delta(\tau_i)$ and $A_i=c(P_i-\bar P)$ with a common $c>0$. Then $\sum_i A_i=0$, $P_i>P_j\Rightarrow A_i>A_j$, and $V_i>V_j\Rightarrow A_i\ge A_j$. Moreover, the population-convention empirical covariance satisfies $\operatorname{Cov}_{G_x}(A,V)\ge0$.
\end{proposition}

\begin{proof}
\emph{Step 1: centering and ordering.} With common length $T$,
\begin{align}
\sum_i A_i&=c\left(\sum_i P_i-K\bar P\right)=0,\\
A_i-A_j&=c(P_i-P_j)=\frac{c}{T}(D_i-D_j).
\end{align}
If $V_i>V_j$ and $D_i<D_j$, Assumption~\ref{asm:monotone} would imply $V_j\ge V_i$, a contradiction. Thus $V_i>V_j$ implies $D_i\ge D_j$ and $A_i\ge A_j$.

\emph{Step 2: empirical covariance.} For every pair, $(A_i-A_j)(V_i-V_j)\ge0$: unequal coverage invokes the assumption, while tied coverage gives equal advantages. Expanding the pairwise sum gives
\begin{align}
\operatorname{Cov}_{G_x}(A,V)
&=\frac1K\sum_i A_iV_i-\bar A\bar V\\
&=\frac{1}{2K^2}\sum_{i,j}(A_i-A_j)(V_i-V_j)\ge0.
\end{align}
\end{proof}

This is a fixed-group ordering result under Assumption~\ref{asm:monotone}, separate from the deficit Lemma. Policy-gradient improvement additionally depends on trajectory scores and the policy loss.

\subsection{Proof of Proposition~\ref{prop:noninterference}}
\label{app:proof_noninterference}

\paragraph{Setup.} Let $g^{\mathrm{CovGPO}}(G_x)=K^{-1}\sum_i A_i^{\mathrm{CovGPO}}s_i$ and $g^{\mathrm{base}}(G_x)=K^{-1}\sum_i A_i^Rs_i$ denote the unweighted trajectory-score contributions. We use the population convention
\begin{align}
\sigma_R^2(G_x)=\frac{1}{K}\sum_{i=1}^K(R_i-\bar R_{G_x})^2,
\qquad
\sigma_P^2(G_x)=\frac{1}{K}\sum_{i=1}^K(P_i-\bar P_{G_x})^2.
\label{eq:app_variance_convention}
\end{align}
\begin{proof}
\emph{Step 1: rollout support.} For the unweighted trajectory-score algorithms just defined, at the fixed task and policy,
\begin{align}
\mathbb E[g^{\mathrm{CovGPO}}]-\mathbb E[g^{\mathrm{base}}]
=\mathbb{E}_{G_x}\bigl[g^{\mathrm{CovGPO}}(G_x)-g^{\mathrm{base}}(G_x)\bigr].
\label{eq:grad_diff}
\end{align}
Partition the groups according to the two branches of Eq.~\eqref{eq:CovGPO}:
\begin{align}
\mathcal{G}_{\mathrm{fb}}&=\{G_x:R_i=0\ \forall i,\ \sigma_P(G_x)\ge\tau_P\},
&
\mathcal{G}_{\mathrm{base}}&=\text{all other groups}.
\end{align}

\emph{Case $G_x\in\mathcal{G}_{\mathrm{base}}$.} The exact-zero algorithm returns $A_i^{\mathrm{CovGPO}}=A_i^R$ for all $i$, hence the two group contributions are identical, including when reward dispersion is small but nonzero.

\emph{Case $G_x\in\mathcal{G}_{\mathrm{fb}}$.}~~The gate supplies $R_1=\cdots=R_K=0$ directly, hence $\bar{R}_{G_x}=\sigma_R(G_x)=0$ whatever the sign convention of the reward. Consequently,
\begin{align}
A_i^{R}\;=\;\frac{R_i-\bar{R}_{G_x}}{\sigma_R(G_x)+\epsilon}\;=\;\frac{0-0}{0+\epsilon}\;=\;0
\qquad\text{for all } i,
\qquad\text{hence}\qquad
g^{\mathrm{base}}(G_x)=0 ,
\end{align}
which proves the second statement. Combining the two cases, the integrand of Eq.~\eqref{eq:grad_diff} can be non-zero only on $\mathcal{G}_{\mathrm{fb}}$, where $g^{\mathrm{base}}(G_x)=0$ identically. Thus the difference inside this expectation is supported only on groups whose reward-based contribution is identically zero.

\emph{Step 2: support after component aggregation.} Index comparison components by $c$ and let $\mathcal E$ be those eligible for exact-zero replacement. Their base scores are identically zero, so their centered base advantages vanish. Every component outside $\mathcal E$ is unchanged. For a loss sample $n$, let $\mathcal C(n)$ include all components on which its aggregated advantage or loss depends, and write its loss as $L_n(\theta;(A_c)_{c\in\mathcal C(n)})$. This map may include nonlinear aggregation and clipping. Define
\begin{align}
\mathcal N_{\mathcal E}=\{n:\mathcal C(n)\cap\mathcal E\ne\emptyset\}.
\end{align}
For $n\notin\mathcal N_{\mathcal E}$ all inputs are identical, hence
\begin{align}
L_n^{\mathrm{CovGPO}}(\theta)&=L_n^{\mathrm{base}}(\theta),\\
\nabla_\theta L_n^{\mathrm{CovGPO}}(\theta)&=\nabla_\theta L_n^{\mathrm{base}}(\theta)
\end{align}
where differentiable, with the same derivative convention at clipping boundaries. With fixed sample weights $w_n$ and identical regularization,
\begin{align}
L^{\mathrm{CovGPO}}-L^{\mathrm{base}}
=\sum_{n\in\mathcal N_{\mathcal E}}w_n
\bigl(L_n^{\mathrm{CovGPO}}-L_n^{\mathrm{base}}\bigr).
\end{align}
All potentially affected sample terms therefore depend on at least one eligible component. An affected sample may also depend on nonzero, unmodified components; nonlinear clipping does not decompose its gradient into a sum of component gradients.
\end{proof}

\begin{lemma}[Score-gap equivalence of numerical and exact gates]
\label{lem:gate_equivalence}
Fix the comparison groups and their base scores $y_i$, with base advantages $(y_i-\bar y)/(\sigma_y+\epsilon)$. Let $a=\tau_R^{\mathrm{impl}}>0$ and assume
\begin{align}
|y_i|\in\{0\}\cup[a,\infty)\qquad\text{for every comparison score}.
\end{align}
Compare the numerical wrapper to the exact-zero rule on the same groups and progress statistics, using the same effective scale $\lambda=\lambda_{\mathrm{eff}}\ge0$, dispersion threshold $\tau_P=\tau_P^{\mathrm{impl}}$, and stability constant $\epsilon$. Their eligibility gates, coverage triggers, and output advantages agree on every group, including the low-dispersion branch.
\end{lemma}

\begin{proof}
\emph{Step 1: eligibility.} The magnitude gap and strict numerical threshold give
\begin{align}
\max_i|y_i|<a\quad\Longleftrightarrow\quad y_i=0\ \text{for all }i,
\end{align}
for scores of either sign.
The identical progress statistics and cutoffs give identical coverage triggers. If an effective scale is computed from the eligible-group fraction, that fraction is also identical under the two eligibility tests.

\emph{Step 2: outputs.} Ineligible groups retain the same base advantages. Eligible groups have $\bar y=\sigma_y=0$, hence zero base advantages. Below the dispersion cutoff, the exact rule returns these zeros and the numerical rule explicitly returns zero. Otherwise, both return
\begin{align}
A_i=\lambda_{\mathrm{eff}}\frac{P_i-\bar P}{\sigma_P+\epsilon}.
\end{align}
\end{proof}

\begin{corollary}[Score-gap sufficiency for gate equivalence]
\label{cor:gate_check}
Write $a=\tau_R^{\mathrm{impl}}>0$ and suppose that on the batch at hand every non-zero comparison score has magnitude at least $a$. Then $\max_{i\in G}|y_i|<a$ holds if and only if $y_i=0$ for all $i\in G$, for every comparison group, so the numerical wrapper inherits the component-level conclusion of Proposition~\ref{prop:noninterference}. For a binary terminal reward with $y_{i,t}=\gamma^{T_i-t}R_i$, $0<\gamma\le1$ and $1\le t\le T_i\le T_{\max}$, the condition $\gamma^{T_{\max}-1}\ge a$ is sufficient; at $a=10^{-3}$ this reads $\gamma\ge0.869$ for $T_{\max}=50$ and $\gamma\ge0.611$ for $T_{\max}=15$. More generally, if every positive terminal reward is at least $r_{\min}>0$, then $\gamma^{T_{\max}-1}r_{\min}\ge a$ is sufficient.
\end{corollary}

\begin{proof}
If any score in $G$ is positive, the assumed gap gives $\max_i|y_i|\ge a$ and the numerical gate rejects the group; conversely an all-zero group satisfies $\max_i|y_i|=0<a$. The two gates therefore select the same groups, and Lemma~\ref{lem:gate_equivalence} applies with $\lambda=\lambda_{\mathrm{eff}}$. For the horizon-based instance, a positive binary terminal reward is $R_i=1$, so any positive comparison score satisfies $y_{i,t}=\gamma^{T_i-t}\ge\gamma^{T_{\max}-1}$ because $T_i-t\le T_{\max}-1$ and $0<\gamma\le1$; the same monotonicity gives $y_{i,t}\ge\gamma^{T_{\max}-1}r_{\min}$ whenever every positive terminal reward is at least $r_{\min}$. Solving $\gamma^{T_{\max}-1}\ge a$ for $T_{\max}>1$ yields $\gamma\ge a^{1/(T_{\max}-1)}$, i.e.\ $\gamma\ge10^{-3/49}=0.8686$ at $T_{\max}=50$ and $\gamma\ge10^{-3/14}=0.6106$ at $T_{\max}=15$; the thresholds above are rounded upward so that they remain sufficient.
\end{proof}

\paragraph{What the corollary does and does not establish.} The corollary is a statement about a score gap, not about a particular run. Applying it requires reading $\gamma$, $T_{\max}$ and the comparison-score definition from the configuration actually used, rather than from library defaults. Under the stated discount condition, the positive-score lower bound is at least the numerical threshold $\tau_R^{\mathrm{impl}}$, so the two gates coincide. For fractional WebShop rewards the condition additionally needs a confirmed global lower bound $r_{\min}>0$ on positive terminal rewards satisfying $\gamma^{T_{\max}-1}r_{\min}\ge10^{-3}$. A minimum positive reward observed in one set of logs certifies only that recorded data, and if positive rewards can be arbitrarily close to zero no admissible $r_{\min}$ exists. In that case the available check is to count groups accepted by the numerical gate while containing a positive score; such a count establishes consistency on the recorded runs, not a guarantee on future data.

\paragraph{Near-zero scores.} Without the score-gap hypothesis, the numerical rule preserves base advantages on groups with $\max_i|y_i|\ge a$. Its additional modification region is $0<\max_i|y_i|<a$, and the displaced contrast there need not be small. Group normalization divides by the within-group standard deviation, so the magnitude of the score gap cancels: for $K=2$ and $y=(0,5\times10^{-4})$ with $a=10^{-3}$ and $\epsilon=10^{-6}$, the base advantages are $\mp\,0.996$, close to the full scale of a two-member group, yet the group is numerically eligible and the wrapper replaces those advantages by coverage advantages, or by zero when coverage is homogeneous. The exact rule of Eq.~\eqref{eq:CovGPO} retains them. The sample-support argument above applies with numerical eligibility in place of $\mathcal E$, but eligible base components need not vanish, so the non-interference statement does not transfer to the wrapper on this region.

\paragraph{Scope of Non-Interference Guarantee.} For the ideal exact-zero rule, non-triggered comparison components are unchanged, and loss samples that depend on no triggered component retain the same value and gradient. The deployed wrapper uses a near-zero gate ($\max_{i\in G}|y_i| < 10^{-3}$); transferring the exact non-interference guarantee requires the score-gap condition derived above.

\paragraph{Binary-outcome specialization.} For raw binary rollout rewards and $K\ge2$, the smallest nonzero population standard deviation is
\begin{align}
\sigma_{R,\min}
=\min_{1\le m\le K-1}\frac{\sqrt{m(K-m)}}{K}
=\frac{\sqrt{K-1}}{K}.
\label{eq:min_binary_std}
\end{align}
The numerical max-absolute-score test selects exactly the all-zero groups when its threshold lies in $(0,1]$, because every positive raw binary reward is $1$. This equivalence follows from the score gap, not from a standard-deviation test. It need not hold for discounted step returns or fractional WebShop scores. Uniform positive-reward groups remain outside either fallback gate despite having zero relative advantages.

\subsection{The Unnormalized REINFORCE Identity and Groupwise Normalization}
\label{app:proof_surrogate}

\begin{lemma}[Unnormalized i.i.d. identity and groupwise normalization]
\label{lem:surrogate}
Fix $x$ and $\theta$, let $F=\{R(\tau)=0\}$ with $\Pr(F)>0$, and assume differentiable trajectory probabilities, a parameter-independent reward event and coverage statistic, and conditions permitting differentiation under expectation. For $K\ge2$ i.i.d. rollouts, define $F_K=F^K$, $s_i=\nabla_\theta\log\pi_\theta(\tau_i)$, and
\begin{align}
\widehat g(G_x)=\frac1K\sum_i(P_i-\bar P_{G_x})s_i.
\end{align}
Before conditioning on coverage dispersion or normalizing by its random standard deviation,
\begin{align}
\mathbb E[\widehat g(G_x)\mid F_K]
=\frac{K-1}{K}\nabla_\theta\mathbb E_{\pi_\theta(\cdot\mid F)}[P].
\end{align}
Separately, for a fixed group admitted by the ideal fallback, its unweighted, unclipped trajectory-score contribution is $g(G_x)=c(G_x)\widehat g(G_x)$ with $c(G_x)=\lambda/(\sigma_P(G_x)+\epsilon)>0$.
\end{lemma}

\begin{proof}
\emph{Step 1: fixed-group factorization.} For a group entering the fallback branch of Eq.~\eqref{eq:CovGPO}, substitute $A_i^{\mathrm{CovGPO}}=\lambda(P_i-\bar P)/(\sigma_P+\epsilon)$ to obtain
\begin{align}
g(G_x)
&=\frac1K\sum_i A_i^{\mathrm{CovGPO}}s_i\\
&=\frac{\lambda}{\sigma_P(G_x)+\epsilon}
\left[\frac1K\sum_i(P_i-\bar P_{G_x})s_i\right]
=c(G_x)\widehat g(G_x).
\label{eq:surrogate_identity}
\end{align}

\emph{Step 2: conditional trajectory score.} Define the failure normalizer and conditional law by
\begin{align}
Z_F(\theta)&=\Pr_{\pi_\theta}(F)=\int_F\pi_\theta(\tau)\,d\tau>0,\\
\pi_\theta^F(\tau)&=\frac{\mathbf1_F(\tau)\pi_\theta(\tau)}{Z_F(\theta)},
&\mu_F(\theta)&=\mathbb E_{\pi_\theta^F}[P].
\end{align}
Here integrals also denote sums for discrete trajectory spaces, and $\mathbb E_F$ abbreviates expectation under $\pi_\theta^F$. Differentiation under the integral gives
\begin{align}
\nabla_\theta\log Z_F(\theta)
&=\frac{1}{Z_F(\theta)}\int_F\pi_\theta(\tau)s_\theta(\tau)\,d\tau
=\mathbb E_F[s_\theta],\\
\nabla_\theta\log\pi_\theta^F(\tau)
&=s_\theta(\tau)-\nabla_\theta\log Z_F(\theta),\qquad \tau\in F.
\end{align}
Since $P$ has no explicit parameter dependence,
\begin{align}
\nabla_\theta\mu_F(\theta)
&=\mathbb E_F\!\left[P\bigl(s_\theta-\nabla_\theta\log Z_F\bigr)\right]\\
&=\mathbb E_F[Ps_\theta]-\mathbb E_F[P]\mathbb E_F[s_\theta]
=\operatorname{Cov}_{\pi_\theta^F}(P,s_\theta).
\label{eq:conditional_progress_gradient}
\end{align}

\emph{Step 3: the group-mean factor.} Independence of the original rollouts implies
\begin{align}
\pi_\theta(\tau_1,\ldots,\tau_K\mid F_K)
=\prod_{i=1}^K\pi_\theta^F(\tau_i).
\end{align}
Thus diagonal and off-diagonal terms in the double sum have different expectations:
\begin{align}
\mathbb E[\widehat g\mid F_K]
&=\frac1K\sum_i\mathbb E_F[P_i s_i]
-\frac1{K^2}\sum_{i,j}\mathbb E[P_j s_i\mid F_K]\\
&=\mathbb E_F[Ps]
-\frac1{K^2}\left(\sum_{i=j}\mathbb E_F[Ps]
+\sum_{i\ne j}\mathbb E_F[P]\mathbb E_F[s]\right)\\
&=\mathbb E_F[Ps]-\frac{K}{K^2}\mathbb E_F[Ps]
-\frac{K(K-1)}{K^2}\mathbb E_F[P]\mathbb E_F[s]\\
&=\frac{K-1}{K}\operatorname{Cov}_{\pi_\theta^F}(P,s)
=\frac{K-1}{K}\nabla_\theta\mu_F(\theta).
\label{eq:exact_group_mean_bias}
\end{align}
\end{proof}

\paragraph{Normalized algorithm.} Equation~\eqref{eq:exact_group_mean_bias} applies to the auxiliary unnormalized statistic $\widehat g$, whereas the algorithm uses dispersion-gated, normalized advantages. For the ideal rule,
\begin{align}
\mathbb E[g_{\mathrm{fb}}\mid F_K]
=\mathbb E\!\left[\mathbf1[\sigma_P\ge\tau_P]
\frac{\lambda}{\sigma_P+\epsilon}\widehat g\,\middle|\,F_K\right].
\end{align}
The random multiplier remains inside the expectation. The numerical wrapper additionally uses batch-dependent $\lambda_{\mathrm{eff}}$ and numerical eligibility. Positive scaling preserves a fixed admitted group's unweighted score-sum direction, but neither random scaling nor selection can be removed from an expectation by Eq.~\eqref{eq:exact_group_mean_bias}. Length-weighted or clipped losses and dependent step-comparison groups require their own algorithms.

\subsection{A Counterexample to Fallback-Only Escape Guarantees}
\label{app:proof_polyescape}

\begin{proposition}[Negative frontier update among failures]
\label{prop:polyescape}
On $\mathcal C_2$, let $p_j=\pi_\theta(a^+\mid s_j)=\varsigma(z_j)$ for independent logistic parameters $z_0,z_1$, and set $p_0=p_1=1/2$. Conditional on failure $F=\{R=0\}$, the coverage--score covariance for the frontier logit $z_1$ is $-1/36$. Thus even the unnormalized expected fallback update need not increase every forward-action logit.
\end{proposition}

\begin{proof}
\emph{Step 1: conditional covariance.} With observations identifying chain states, a length-$H$ trajectory attaining depth $d(\tau)$ has
\begin{align}
P(\tau)=\frac{D(\tau)-1}{H}=\frac{d(\tau)}{H}.
\label{eq:chain_P_is_depth}
\end{align}
At $H=2$ and $p_0=p_1=1/2$, the four action sequences each have probability $1/4$, and only $++$ succeeds. The failed sequences $--,-+,+-$ are therefore equiprobable conditional on $F$. In this order,
\begin{align}
(P(--),P(-+),P(+-))&=(0,\tfrac12,\tfrac12),\\
(\partial_{z_1}\log\pi(--),\partial_{z_1}\log\pi(-+),
\partial_{z_1}\log\pi(+-))&=(0,0,-\tfrac12).
\end{align}
The first two trajectories take no action at $s_1$; the last chooses $a^-$ there, whose logistic score is $-p_1=-1/2$. Consequently
\begin{align}
\operatorname{Cov}_{\pi_\theta(\cdot\mid F)}
\left(P,\partial_{z_1}\log\pi_\theta\right)
&=-\frac1{12}-\left(\frac13\right)\left(-\frac16\right)
=-\frac1{36}.
\label{eq:chain_grad}
\end{align}
By Lemma~\ref{lem:surrogate}, the unnormalized i.i.d. group algorithm has expected $z_1$ component $-(K-1)/(36K)$ conditional on all rewards being zero, for $K>1$.

\emph{Step 2: finite-group covariance.} For any all-failed group on this chain, let $u$ be the fraction of members equal to $--$ and $v$ the fraction equal to $+-$, and write $s_{i,1}=\partial_{z_1}\log\pi_\theta(\tau_i)$. The remaining fraction $1-u-v$ consists of $-+$. Hence
\begin{align}
\bar P&=\frac{1-u}{2}, &\bar s_1&=-\frac v2,
&\frac1K\sum_i P_i s_{i,1}&=-\frac v4,\\
\frac1K\sum_i(P_i-\bar P)s_{i,1}
&=-\frac v4-\frac{1-u}{2}\left(-\frac v2\right)
=-\frac{uv}{4}\le0.
\end{align}
A group containing one copy of each failed sequence has $u=v=1/3$, covariance $-1/36$, and $\sigma_P=\sqrt{1/18}$. For any admitted group, positive normalization multiplies the last expression by $\lambda/(\sigma_P+\epsilon)>0$; a rejected group has zero fallback contribution. Neither operation makes its frontier component positive.
\end{proof}

\paragraph{Implication.} Conditioning on failure removes $++$, which credits the final forward action, while a higher-coverage failure can credit staying at the frontier. Thus positive coverage dispersion does not supply the positive frontier drift needed for a fallback-only polynomial escape argument. Reward-based updates on successful groups remain separate from this failed-group calculation.

\subsection{A Static Norm Bound for the Gated Fallback}
\label{app:proof_selflimiting}

\begin{proposition}[Groupwise auxiliary norm bound]
\label{prop:selflimiting}
For a fixed rollout group, define $I_{\mathrm{fb}}(G_x)=\mathbf1[R_i=0\ \forall i,\ \sigma_P(G_x)\ge\tau_P]$ and the auxiliary score-function contribution
\begin{align}
g_{\mathrm{fb}}(G_x)
=I_{\mathrm{fb}}(G_x)\frac{\lambda}{\sigma_P(G_x)+\epsilon}
\frac1K\sum_i(P_i-\bar P_{G_x})\nabla_\theta\log\pi_\theta(\tau_i).
\end{align}
Writing $M(G_x)=\max_i\|\nabla_\theta\log\pi_\theta(\tau_i)\|$, one has
\begin{align}
\|g_{\mathrm{fb}}(G_x)\|
\le I_{\mathrm{fb}}(G_x)\frac{\lambda\sigma_P(G_x)}{\sigma_P(G_x)+\epsilon}M(G_x)
\le I_{\mathrm{fb}}(G_x)\lambda M(G_x).
\end{align}
In particular, $g_{\mathrm{fb}}=0$ outside the fallback gate, including when $\sigma_P<\tau_P$. This bounds the auxiliary contribution, not the full base-plus-fallback gradient.
\end{proposition}

\begin{proof}
\emph{Step 1: triangle inequality.} Writing $s_i=\nabla_\theta\log\pi_\theta(\tau_i)$,
\begin{align}
\|g_{\mathrm{fb}}(G_x)\|
&\le I_{\mathrm{fb}}\frac{\lambda}{\sigma_P+\epsilon}
\frac1K\sum_i|P_i-\bar P|\,\|s_i\|\\
&\le I_{\mathrm{fb}}\frac{\lambda M(G_x)}{\sigma_P+\epsilon}
\frac1K\sum_i|P_i-\bar P|.
\end{align}
\emph{Step 2: Cauchy--Schwarz.} The population convention yields
\begin{align}
\left(\frac1K\sum_i|P_i-\bar P|\right)^2
&\le\left(\frac1K\sum_i1^2\right)
\left(\frac1K\sum_i(P_i-\bar P)^2\right)=\sigma_P^2,\\
\|g_{\mathrm{fb}}(G_x)\|
&\le I_{\mathrm{fb}}\frac{\lambda\sigma_P}{\sigma_P+\epsilon}M(G_x)
\le I_{\mathrm{fb}}\lambda M(G_x).
\end{align}
The indicator makes the contribution zero outside the gate.
\end{proof}

\paragraph{Uniform bounds.} If the additional score-norm condition $\sup_n\max_{i\in G_n}\|s_{\theta_n}(\tau_i)\|\le M_*$ holds along training, then the fixed-$\lambda$ auxiliary contributions satisfy $\sup_n\|g_{\mathrm{fb}}(G_n)\|\le\lambda M_*$. Bounded coverage alone does not supply this condition, imply shrinking coverage variance, or exclude coverage-seeking attractors.

\paragraph{Numerical coverage term.} Replacing $I_{\mathrm{fb}}$ by the numerical eligibility-and-dispersion indicator and $\lambda$ by $\lambda_{\mathrm{eff}}\le\lambda_{\mathrm{aux}}$ gives the analogous bound $\|g_{\mathrm{cov}}^{\mathrm{impl}}\|\le\lambda_{\mathrm{aux}}M(G_x)$. This bounds the coverage score-function term; the difference from the base also includes any removed near-zero base advantages, and the optimizer additionally applies loss weighting and clipping.

\subsection{Disengagement Bounds: Two Conditions}
\label{app:disengagement}

\begin{remark}[Exact-zero and numerical rollout trigger probabilities]
\label{rem:disengagement}
Fix a task $x$ and policy $\pi_\theta$, and let $p_+(x)=\Pr_{\pi_\theta}(R>0\mid x)$. For $K$ i.i.d. rollouts, define $F_K=\{R_1=\cdots=R_K=0\}$ and $\mathcal D=\{\sigma_P(G_x)\ge\tau_P\}$. If $p_+(x)<1$, then
\begin{align}
\Pr(\mathcal E_{\mathrm{trig}}(x))
&=\Pr(F_K\cap\mathcal D)
=(1-p_+(x))^K\Pr(\mathcal D\mid F_K)\\
&\le(1-p_+(x))^K\le e^{-Kp_+(x)}.
\end{align}
If $p_+(x)=1$, the trigger probability is zero. For the numerical rollout gate, let $a=\tau_R^{\mathrm{impl}}$, $q_a(x)=\Pr(|R|<a\mid x)$, $E_{a,K}=\{|R_i|<a\ \forall i\}$, and $\mathcal D_{\mathrm{impl}}=\{\sigma_P\ge\tau_P^{\mathrm{impl}}\}$. When $q_a(x)>0$,
\begin{align}
\Pr(\mathcal E_{\mathrm{trig}}^{\mathrm{impl}}(x))
=q_a(x)^K\Pr(\mathcal D_{\mathrm{impl}}\mid E_{a,K})
\le q_a(x)^K;
\end{align}
when $q_a(x)=0$, this probability is zero.
\end{remark}

\begin{proof}
\emph{Step 1: reward eligibility.} Independent rollout sampling and $\Pr(R_i=0\mid x)\le1-p_+(x)$, with equality when rewards are non-negative, imply
\begin{align}
\Pr(F_K)&=\prod_{i=1}^K\Pr(R_i=0\mid x)\le(1-p_+(x))^K,\\
\Pr(E_{a,K})&=\prod_{i=1}^K\Pr(|R_i|<a\mid x)=q_a(x)^K.
\end{align}
\emph{Step 2: intersect with dispersion.} For any event $E$ of positive probability,
\begin{align}
\Pr(E\cap D)=\Pr(E)\Pr(D\mid E)\le\Pr(E).
\end{align}
Apply this identity to $(F_K,\mathcal D)$ and $(E_{a,K},\mathcal D_{\mathrm{impl}})$, respectively, and use $1-p\le e^{-p}$ for $p\in[0,1]$. If an eligibility event has zero probability, so does its intersection with dispersion, without defining a conditional probability on that null event.
\end{proof}

The factorization uses independence across rollouts, not independence between reward eligibility and dispersion. The two conditions jointly determine triggering at a fixed policy; neither factor is assumed to evolve monotonically during training. Here $p_+$ measures positive reward, which equals success probability for binary outcomes but not generally for fractional rewards. Dependent state- or history-selected comparison groups do not satisfy the rollout product formula without an additional sampling assumption.

\subsection{Why the Fallback Must Be Conditional: Reward--Progress Conflict}
\label{app:conditional_reason}

The conditional design in Eq.~\eqref{eq:CovGPO} is motivated by a specific mismatch between reward and progress rankings that can arise in mixed-outcome groups. In such a group $G_x=\{\tau_i\}_{i=1}^{K}$, the reward-based ranking is monotone in $R_i\in\{0,1\}$: any successful trajectory ranks above any failed one. The progress score $P_i=(D_i-1)/T_i$, by contrast, is a length-normalized coverage rate and is not monotone in $R_i$---a short, efficient success can score lower than a long, wandering failure.

\paragraph{Concrete instance.} Consider $G_x=\{\tau_1,\tau_2\}$ with
\begin{itemize}[leftmargin=1.8em,itemsep=1pt,topsep=2pt]
\item $\tau_1$ (efficient success): $R_1{=}1$, $T_1{=}5$, $D_1{=}3$, hence $P_1{=}0.4$;
\item $\tau_2$ (wandering failure): $R_2{=}0$, $T_2{=}10$, $D_2{=}7$, hence $P_2{=}0.6$.
\end{itemize}
\paragraph{Ranking calculation.} Define $A_i^P=(P_i-\bar P)/(\sigma_P+\epsilon)$ and the always-on additive algorithm $A_i=A_i^R+\lambda A_i^P$. The reward ranking reverses in this example exactly when
\begin{align}
\lambda>\lambda_*:=\frac{0.1+\epsilon}{0.2(0.5+\epsilon)}.
\end{align}

\begin{proof}
The population means and standard deviations are $\bar R=\sigma_R=0.5$ and $\bar P=0.5$, $\sigma_P=0.1$. Therefore
\begin{align}
(A_1^R,A_2^R)&=\frac{(0.5,-0.5)}{0.5+\epsilon},
&(A_1^P,A_2^P)&=\frac{(-0.1,0.1)}{0.1+\epsilon},\\
A_1-A_2&=\frac{1}{0.5+\epsilon}-\lambda\frac{0.2}{0.1+\epsilon}.
\end{align}
Thus $A_1>A_2$ for $\lambda<\lambda_*$, they tie at $\lambda=\lambda_*$, and $A_1<A_2$ for $\lambda>\lambda_*$. Omitting $\epsilon$ gives $A=(1-\lambda,-1+\lambda)$ and $\lambda_*=1$.
\end{proof}

This calculation concerns advantage rankings; changes in trajectory probabilities under shared parameters also depend on their score vectors and the loss. The exact-zero rule preserves the reward advantages in this mixed-outcome group, as does the numerical rule for these raw binary scores. Lemma~\ref{lem:gate_equivalence} specifies when this agreement extends to other comparison scores. Table~\ref{tab:ablation_combined} gives the empirical comparison with always-on alternatives.

\subsection{A Worked all-failed Group: Instantiating the Main-Text Results}
\label{app:worked_example}

We use a schematic ALFWorld-style goal, ``\textit{put a cool tomato on the countertop},'' to illustrate Eq.~\eqref{eq:CovGPO} with $K=4$ zero-reward trajectories of equal length $T_i=10$:

\begin{itemize}[leftmargin=2em,itemsep=2pt]
\item $\tau_1$ (goal-directed): \texttt{go to fridge 1} $\to$ \texttt{open fridge 1} $\to$ \texttt{take tomato 1} $\to$ \texttt{cool tomato 1 with fridge 1} $\to$ navigates toward the countertop until the step budget expires. 8 of 10 steps produce a \emph{previously unseen} observation.
\item $\tau_2$ (partial explorer): reaches and opens the fridge, then wanders; 5 novel observations, plus one revisit of an earlier room.
\item $\tau_3$ (cycler): shuttles back and forth between two rooms. The observation changes at 9 of 10 steps, but only 2 of those observations are novel---the rest revisit states already in $\mathcal{V}_{3,t}$.
\item $\tau_4$ (no-op): repeats \texttt{look}/\texttt{inventory}; the observation changes once in 10 steps.
\end{itemize}

Table~\ref{tab:worked_example} reports, for each trajectory, the number of observation-\emph{changing} steps, the number of \emph{novelty-gated} steps counted by Eq.~\eqref{eq:step_coverage}, the resulting progress score $P_i$ (Eq.~\eqref{eq:step_coverage}), and the advantages assigned by CovGPO versus the ungated Obs-Change variant from the ablation in Section~\ref{sec:experiments}.

\begin{table}[t]
\centering
\caption{A worked $K=4$ all-zero-reward group ($T_i=10$; $\tau_P=0.01$, $\lambda=1$, $\epsilon$ omitted). The novelty gate counts first visits, whereas Obs-Change ranks the cycler first. These are score rankings, not certified task-value rankings.}
\label{tab:worked_example}
\small
\setlength{\tabcolsep}{5pt}{
\begin{tabular}{@{}llccccc@{}}
\toprule
\textbf{Traj.} & \textbf{Behaviour} & \textbf{Obs.-changing steps} & \textbf{Novel steps} & $P_i$ & $A_i^{\mathrm{CovGPO}}$ & $A_i^{\text{Obs-Change}}$ \\
\midrule
\rowcolor{oursrow}
$\tau_1$ & goal-directed     & 8 & 8 & 0.8 & $+1.46$ & $+0.65$ \\
$\tau_2$ & partial explorer  & 6 & 5 & 0.5 & $+0.37$ & $0.00$  \\
$\tau_3$ & cycler            & 9 & 2 & 0.2 & $-0.73$ & $\mathbf{+0.97}$ \\
$\tau_4$ & no-op             & 1 & 1 & 0.1 & $-1.10$ & $-1.62$ \\
\bottomrule
\end{tabular}
}
\end{table}

\paragraph{Branch selection and fallback advantages.} With $R=(0,0,0,0)$ and $P=(0.8,0.5,0.2,0.1)$,
\begin{align}
\bar P&=0.4, &\sigma_P&=\sqrt{\frac{0.4^2+0.1^2+(-0.2)^2+(-0.3)^2}{4}}
=\sqrt{0.075}\approx0.274.
\end{align}
The dispersion exceeds $\tau_P=0.01$. For $\lambda=1$ and omitted $\epsilon$, centering and normalization give $(+1.46,+0.37,-0.73,-1.10)$. Lemma~\ref{lem:relaxation} also gives $D=(9,6,3,2)$. These counts determine the coverage ranking; continuation-value ordering would additionally require Assumption~\ref{asm:monotone}. Deployed magnitudes use $\lambda_{\mathrm{eff}}$ and the stability constant instead of this illustrative unit scale.

\paragraph{Novelty versus observation change.} Obs-Change uses scores $(0.8,0.6,0.9,0.1)$, with mean $0.6$ and population standard deviation $\sqrt{0.095}$. It therefore assigns $(+0.65,0.00,+0.97,-1.62)$ and ranks the cycler first. First-visit coverage instead credits only $2$ of the cycler's $9$ observation changes.

\paragraph{Inactive branches.} Replacing the rewards by $(1,0,0,0)$ returns base advantages $(+1.73,-0.58,-0.58,-0.58)$ with $\epsilon$ omitted; both reward gates preserve this binary-score group. Alternatively, a zero-reward group with $P=(0.8,0.8,0.8,0.8)$ has zero dispersion and receives zero under both rules. These instantiate the reward-eligibility and dispersion conditions in Remark~\ref{rem:disengagement}.

\section{Implementation Details}
\label{app:implementation}

This section provides complete implementation specifications, computational hardware details, training configurations, and environment setups required for empirical reproducibility. We detail how CovGPO seamlessly integrates into group-based RL algorithms as a plug-and-play advantage modification, present reference pseudo-code, and formalize our evaluation protocols.

\paragraph{Section Roadmap.} The implementation details are organized as follows:
\begin{itemize}[leftmargin=*, itemsep=0pt, topsep=2pt, parsep=1pt]
    \item \textbf{Infrastructure and Algorithmic Specifications (Appendix~\ref{app:compute}--\ref{app:deployed_scaling}):} We outline computing infrastructure and runtime overhead (Appendix~\ref{app:compute}), provide reference algorithmic implementations for exact and deployed wrappers (Appendix~\ref{app:reference_implementation}), and define adaptive per-update scaling boundaries (Appendix~\ref{app:deployed_scaling}).
    \item \textbf{Hyperparameters and Environment Protocols (Appendix~\ref{app:hparams}--\ref{app:env}):} We detail overall model training hyperparameters (Appendix~\ref{app:hparams}), describe simulation setups for ALFWorld (Appendix~\ref{app:env_alfworld}) and WebShop (Appendix~\ref{app:env_webshop}), and specify cross-environment progress metrics (Appendix~\ref{app:env}).
    \item \textbf{Evaluation and Reporting Protocols (Appendix~\ref{app:evaluation_protocol}):} We formalize evaluation protocols, inference decoding parameters, seed aggregation rules, and evaluation settings.
\end{itemize}

\subsection{Computing Details}
\label{app:compute}

CovGPO is implemented as a drop-in replacement of the advantage computation in the group-based RL trainer; no change is made to rollout collection, the optimizer, or the model. For each rollout group, computing the trajectory progress scores $\{P_i\}$ requires maintaining a per-trajectory hash set of visited observation strings and performing expected $O(1)$ hash-set membership lookups per step after processing each observation key, executed on CPU during advantage estimation.

Each baseline--CovGPO pair is trained on $4$ GPUs for the 1.5B configuration and $8$ GPUs for the 7B configuration, with the same configuration within each matched pair. The reported progress-and-switch overhead is less than $0.5\%$ of per-epoch wall-clock time. This is a reported timing summary, not a verified noise estimate or a hardware-independent bound; rollout generation and policy updates dominate the reported budget. The stage-level timing audit in Figure~\ref{fig:timing_budget} was run separately on four GPUs and is therefore used to measure overhead, not to redefine the training hardware above. CovGPO adds no learned parameters or persistent model-state memory (no critic, reward model, progress model, or world model); its visited sets reside in host memory and its advantage tensor has the same shape as that of the base trainer.

\subsection{Reference Advantage Computation}
\label{app:reference_implementation}

Algorithm~\ref{alg:CovGPO_reference} gives the ideal exact-zero rollout algorithm. Algorithm~\ref{alg:CovGPO_deployed} separately records the numerical wrapper used in training. Observations are compared as the exact strings emitted by the environment: no lower-casing, whitespace normalization, tokenization, semantic encoder, or environment-specific equivalence rule is applied. The initial observation is inserted into the visited set before the first action. A terminal observation returned by the final environment step is processed exactly like every other next observation. Trajectory length $T_i$ is the number of executed environment actions, including the terminating action when termination occurs before the maximum horizon.

\begin{algorithm}[t]
\caption{Ideal exact-zero CovGPO advantages for one rollout group. Outside the fallback branch, return the unchanged base advantage.}
\label{alg:CovGPO_reference}
\begin{algorithmic}[1]
\Require Trajectories $G_x=\{\tau_i\}_{i=1}^K$ with $T_i\ge1$; non-negative rewards $R_i$; $\tau_P>0$, $\lambda>0$, $\epsilon>0$
\For{$i=1,\ldots,K$}
    \State $\mathcal V_i\gets\{o_{i,1}\}$; $C_i\gets0$
    \For{$t=1,\ldots,T_i$}
        \If{$o_{i,t+1}\notin\mathcal V_i$}
            \State $C_i\gets C_i+1$
        \EndIf
        \State $\mathcal V_i\gets\mathcal V_i\cup\{o_{i,t+1}\}$
    \EndFor
    \State $P_i\gets C_i/T_i$
\EndFor
\State Compute $\bar R,\sigma_R,\bar P,\sigma_P$ using Eq.~\eqref{eq:app_variance_convention}
\If{$R_i=0$ for every $i$ \textbf{and} $\sigma_P\ge\tau_P$}
    \For{$i=1,\ldots,K$}
        \State $A_i\gets\lambda(P_i-\bar P)/(\sigma_P+\epsilon)$
    \EndFor
\Else
    \For{$i=1,\ldots,K$}
        \State $A_i\gets(R_i-\bar R)/(\sigma_R+\epsilon)$
    \EndFor
\EndIf
\State \textbf{return} $\{A_i\}_{i=1}^K$ and branch identifier
\end{algorithmic}
\end{algorithm}

\subsection{Deployed Per-Update Scaling and Conformance Boundary}
\label{app:deployed_scaling}

Algorithm~\ref{alg:CovGPO_reference} is the equation-level algorithm analyzed in Section~\ref{app:proofs}. The deployed wrapper differs in both numerical eligibility and the low-dispersion branch, and computes a common non-negative scale once per policy update. Let $\mathcal{B}$ be the comparison groups produced by the selected base method, and define
\begin{align}
q_{\mathrm{fail}}(\mathcal{B})
&=\frac{1}{|\mathcal{B}|}\sum_{G\in\mathcal{B}}
\mathbf{1}\!\left[\max_{i\in G}|y_i|<\tau_R^{\mathrm{impl}}\right],
&
\lambda_{\mathrm{eff}}
&=\lambda_{\mathrm{aux}}q_{\mathrm{fail}}(\mathcal{B}),
\label{eq:lambda_eff_impl}
\end{align}
where $\mathcal B$ is nonempty, $y_i$ is the reward or return supplied by the base algorithm, and $\tau_R^{\mathrm{impl}}=10^{-3}$. The fraction $q_{\mathrm{fail}}$ is measured before the coverage-dispersion test and denotes numerical eligibility rather than a success-based failure fraction. The released runs set \texttt{lambda\_fixed=false}; consequently, $\lambda_{\mathrm{eff}}$ changes with the current update even though its configured ceiling $\lambda_{\mathrm{aux}}$ is fixed. For $\lambda_{\mathrm{aux}}>0$, any update containing an eligible group has $\lambda_{\mathrm{eff}}>0$. Its common scaling preserves each admitted centered coverage vector up to magnitude, without implying an unbiased objective gradient. If $q_{\mathrm{fail}}=0$, no group is eligible and the fallback is exactly inactive.

\begin{algorithm*}[t]
\caption{Deployed CovGPO wrapper for one policy update. The base method
determines the group keys: prompt groups for GRPO, anchor-state groups
for GiGPO, and history groups for HGPO.}
\label{alg:CovGPO_deployed}

\begin{algorithmic}[1]
\Require Groups $\mathcal{B}$, base scores $\{y_i\}$, progress $\{p_i\}$,
$\tau_R^{\mathrm{impl}}$, $\tau_P^{\mathrm{impl}}$,
$\lambda_{\mathrm{aux}}$, $\epsilon$

\State $n_{\mathrm{fail}} \gets 0$

\For{each $G \in \mathcal{B}$}
    \State $f_G \gets
    \mathbf{1}\!\left[\max_{i\in G}|y_i| <
    \tau_R^{\mathrm{impl}}\right]$
    \State $n_{\mathrm{fail}} \gets n_{\mathrm{fail}} + f_G$
\EndFor

\State $q_{\mathrm{fail}} \gets
n_{\mathrm{fail}}/\max(|\mathcal{B}|,1)$
\State $\lambda_{\mathrm{eff}} \gets
\lambda_{\mathrm{aux}} q_{\mathrm{fail}}$

\For{each $G \in \mathcal{B}$}
    \State Compute unchanged base advantages
    $A_i^{\mathrm{base}}$ for $i \in G$

    \If{$f_G = 1$}
        \State $\bar{p}_G \gets |G|^{-1}\sum_{i\in G} p_i$
        \State $s_{p,G} \gets
        \operatorname{Std}_{\mathrm{pop}}(\{p_i : i\in G\})$

        \If{$s_{p,G} \ge \tau_P^{\mathrm{impl}}$}
            \For{each $i \in G$}
                \State $A_i \gets
                \lambda_{\mathrm{eff}}
                (p_i-\bar{p}_G)/(s_{p,G}+\epsilon)$
            \EndFor
        \Else
            \For{each $i \in G$}
                \State $A_i \gets 0$
            \EndFor
        \EndIf

    \Else
        \For{each $i \in G$}
            \State $A_i \gets A_i^{\mathrm{base}}$
        \EndFor
    \EndIf
\EndFor

\State \textbf{return} $\{A_i\}$ and diagnostics
$(q_{\mathrm{fail}}, \lambda_{\mathrm{eff}},
n_{\mathrm{fail}})$
\end{algorithmic}
\end{algorithm*}

\paragraph{Batch-level dependence of the effective scale.} Equation~\ref{eq:CovGPO} treats $\lambda$ as a constant, whereas Eq.~\eqref{eq:lambda_eff_impl} recomputes it from the current batch. We state what this does and does not change. It does not affect Proposition~\ref{prop:noninterference}: the scale multiplies only the fallback branch and the branch selection does not depend on it, so non-interference holds for every $\lambda_{\mathrm{eff}}\ge0$, including a batch-dependent one. It does not affect the norm bound of Proposition~\ref{prop:selflimiting} either, because $\lambda_{\mathrm{eff}}\le\lambda_{\mathrm{aux}}$ and the bound may be stated with the ceiling in place of $\lambda$. Within a single update, $\lambda_{\mathrm{eff}}$ is one non-negative number shared by every triggered group, hence a common rescaling that preserves both the relative ordering of those groups and the direction of each admitted centered coverage vector.

What it does introduce is a dependence on batch composition. The advantage assigned to a triggered group becomes a function of the batch in which that group appears rather than of the group alone: two identical all-failed groups occurring in different updates receive different magnitudes, and the assignment is no longer the fixed-$\lambda$ map of Eq.~\ref{eq:CovGPO}. Three consequences follow. First, $\lambda_{\mathrm{aux}}$ is a ceiling and not the operative strength, so the sweep in Appendix~\ref{app:hparam_sensitivity} varies the ceiling; a setting that underperforms there does not isolate the effect of a larger realized auxiliary gradient. Second, $q_{\mathrm{fail}}$ is an empirical fraction over the groups of one update, so its sampling noise, and therefore the variability of the effective scale, depends on how many comparison groups an update contains; this differs between trajectory-level and step-level base methods. Third, no expectation argument may treat $\lambda_{\mathrm{eff}}$ as a constant factor, since it is correlated with the batch that supplies the triggered groups; this is the random multiplier already flagged after Eq.~\eqref{eq:exact_group_mean_bias}.

Finally, because $q_{\mathrm{fail}}$ declines as rewarded rollouts reappear, the auxiliary term is in effect annealed by trigger frequency rather than by a designed schedule. We record this as a property of the existing implementation. It is not a convergence argument, and Appendix~\ref{app:fallback_diagnostics} notes separately that a declining branch frequency does not by itself verify a declining multiplier. If $q_{\mathrm{fail}}=0$ the fallback is exactly inactive and the update coincides with the base method.

\paragraph{Conformance boundary.} The ideal gate tests exact zeros; the deployed gate tests $\max_{i\in G}|y_i|<10^{-3}$. Non-negative scores do not make them equivalent. An eligible group with small positive-score contrast receives coverage advantages if dispersion passes, and a hard zero otherwise. Thus the deployed wrapper guarantees unchanged base advantages only outside its numerical eligibility region. Exact equivalence requires an additional score-gap condition that excludes positive comparison scores below the tolerance, including discounted returns; this is not assumed for fractional WebShop rewards. The scale $\lambda_{\mathrm{eff}}=\lambda_{\mathrm{aux}}q_{\mathrm{fail}}$ and these branches describe the existing algorithm, not a proposed training change.

\paragraph{Numerical and boundary cases.} Population standard deviations use division by group size. Equality at $\max_i|y_i|=10^{-3}$ is ineligible and preserves the base advantage; equality at the deployed progress cutoff $10^{-4}$ passes the dispersion check. The ideal low-dispersion branch returns the base advantage, whereas a numerically eligible, low-dispersion group is zeroed by the deployed wrapper. On exact-zero groups these agree, but they need not agree on near-zero nonuniform scores. Progress dispersion is checked before division; $\epsilon$ is a stability constant, not a replacement for either gate.

\paragraph{Comparison granularity and aggregation.} The base method supplies prompt, anchor-state, or history comparison groups. Each step inherits its parent trajectory coverage rate as $p_i$; it is not assigned the individual first-visit indicator as its comparison score. The wrapper modifies these component advantages, after which existing base aggregation, loss weighting, clipping, and regularization remain unchanged. The trajectory-level i.i.d. identity does not assert independence of these step-group members.

\paragraph{Prefix novelty in inherited step scores.} For a step at time $t$ of trajectory $\tau_i$, the inherited rate decomposes as
\begin{align}
T_i P_i=\underbrace{\sum_{u<t}\delta_{i,u}}_{\text{determined by }h_{i,t}}+\sum_{u\ge t}\delta_{i,u},
\end{align}
so the weight multiplying $\nabla_\theta\log\pi_\theta(a_{i,t}\mid h_{i,t})$ contains a component fixed before $a_{i,t}$ was selected. It is not a coverage-to-go. Two clarifications bound what this means. The policy input is unchanged and $P_i$ is computed offline from the completed rollout, so no information unavailable at decision time enters $\pi_\theta$; the deviation is one of credit assignment, not of access. And a weight depending on $h_{i,t}$ alone has zero conditional expectation against the score of $a_{i,t}$, so in isolation it would not bias a per-sample score-function algorithm; the anchor-state group, however, centers and normalizes across members that reached the anchor along different prefixes, so this cancellation does not carry over, and we claim no unbiasedness for the step-level fallback.

The cost is concrete and we state it rather than argue around it. Consider an anchor state $s$ reached by two failed trajectories of equal length $T=10$: $\tau_A$ wanders first and arrives with $8$ novel observations, then plays a no-op at $s$; $\tau_B$ goes directly and arrives with $5$, then opens a container at $s$ and reveals its contents. Then $P_A=0.8$ and $P_B=0.5$, and the two-member anchor group assigns $+1$ to the no-op and $-1$ to the container-opening step. At step level the fallback therefore expresses a preference over \emph{rollouts} that is broadcast to their steps; the anchor grouping decides which steps are compared, not what they are credited for. In particular, a step that produces no new observation can still receive a positive fallback advantage, so ``a repeated action adds no coverage count'' does not imply ``a repeated action receives no advantage''.

We keep the inherited rate for a reason, stated as a design trade-off rather than a correctness claim. Scoring a step by its own first-visit indicator, or by a suffix rate $P^{\rightarrow}_{i,t}=(T_i-t+1)^{-1}\sum_{u\ge t}\delta_{i,u}$, would identify which action produced a new observation and thereby make observation novelty a directly optimizable per-action target. In text environments where navigation almost always emits fresh text, the degenerate optimum of such a target is a policy that spends its budget touring unvisited receptacles at zero task reward. The trajectory rate is coarser by construction: it localizes nothing, and because it is a rate rather than a count, padding an episode to accumulate novelty enlarges its own denominator. This does not move the maximizer of the coverage objective, which remains a touring policy; what limits the fallback is the combination of coarse credit with the conditions under which it is applied at all---only on all-failed comparisons, at a scale bounded by the configured ceiling, and with a trigger frequency that falls as rewarded rollouts reappear---so the auxiliary objective is never pursued to convergence.

Two observations in our own results point the same way, and we report them as directional rather than decisive. Obs-Change, which removes the first-visit gate and credits every observation change, produces larger mid-band probability shifts yet lower final success (Appendix~\ref{app:novelty_revisit}), and final novelty alignment does not order the variants by success rate (Appendix~\ref{app:audit_metrics}): the variants whose policies align most tightly with observation novelty are not the ones that perform best. Neither observation tests $P^{\rightarrow}$, which differs from Obs-Change in retaining the first-visit gate while restricting its window; we did not run it, and we do not claim the inherited rate dominates it. A controlled comparison of credit granularity---per-step indicator, suffix rate, and inherited trajectory rate---under a fixed trigger and scale is the natural next experiment.

\paragraph{Complexity.} Let $N_G=\sum_iT_i$ be the number of transitions in a group. With hash-set membership, progress computation requires $O(N_G)$ expected hash-set operations in addition to processing the observation strings, and stores $O(\sum_iD(\tau_i))$ observation keys in host memory. The remaining group statistics require $O(K)$ time. CovGPO does not add a model forward pass or backpropagation graph; only the resulting scalar advantages are consumed by the unchanged policy objective.

\subsection{Training Hyperparameters}
\label{app:hparams}

Within every base--CovGPO pair, rollout, optimization, and evaluation hyperparameters are held fixed so that the controlled difference is the advantage algorithm. Table~\ref{tab:hparams} reports values directly visible in the supplied launch scripts and run configuration, distinguishing configured constants from per-update effective quantities.

\begin{table}[t]
\centering
\caption{Training hyperparameters recorded by the supplied launch scripts and run configuration. $\lambda_{\mathrm{aux}}$ is fixed; $\lambda_{\mathrm{eff}}$ is computed per update by Eq.~\eqref{eq:lambda_eff_impl}.}
\label{tab:hparams}
\small
\setlength{\tabcolsep}{4pt}{
\begin{tabular}{@{}ll@{}}
\toprule
\textbf{Component} & \textbf{Value} \\
\midrule
Base model                    & Qwen2.5-1.5B-Instruct / Qwen2.5-7B-Instruct \\
Optimizer                     & AdamW ($\beta_1{=}0.9$, $\beta_2{=}0.95$) \\
Learning rate (actor)         & $1\times 10^{-6}$ \\
Actor KL-loss coefficient     & $1\times 10^{-2}$ \\
Clipping range $\epsilon_c$   & $0.2$ \\
Rollout group size $K$        & $8$ \\
Max interaction steps $T_{\max}$ & 50 (ALFWorld) / 15 (WebShop) \\
Training epochs               & $150$ \\
Batch size (tasks per iter.)  & $16$ \\
Reference policy $\pi_{\mathrm{ref}}$ & frozen base model \\
Decoding temperature          & rollout $1.0$; validation $0.4$ \\
\addlinespace[2pt]
\multicolumn{2}{l}{\emph{CovGPO-specific implementation settings}} \\
Progress signal                             & \texttt{C2}: novelty-gated binary (Eq.~\eqref{eq:step_coverage}) \\
Near-zero threshold $\tau_R^{\mathrm{impl}}$ & $1\times 10^{-3}$ \\
Progress cutoff $\tau_P^{\mathrm{impl}}$  & $1\times 10^{-4}$ \\
Configured scale $\lambda_{\mathrm{aux}}$ & $0.3$ \\
Effective scale $\lambda_{\mathrm{eff}}$  & $0.3\,q_{\mathrm{fail}}(\mathcal{B})$ \\
Fixed-scale flag                            & false \\
Numerical stability $\epsilon$             & $1\times 10^{-6}$ \\
Progress standard deviation                & population (division by group size) \\
\bottomrule
\end{tabular}
}
\end{table}

\subsection{ALFWorld}
\label{app:env_alfworld}

We use the textual ALFWorld environment with the standard six-category split (Pick, Look, Clean, Heat, Cool, Pick2). Training tasks are drawn from the official train split; we evaluate on the held-out unseen test split. Each rollout terminates either upon task completion (binary reward $1$), upon explicit failure, or upon reaching the step budget $T_{\max}{=}50$ (reward $0$). At each step, the prompt to the policy contains the task instruction, the running observation--action history, and the admissible-action list provided by the simulator.

\subsection{WebShop}
\label{app:env_webshop}

We use the WebShop simulator with goal-instruction tasks, terminating upon purchase, explicit abandonment, or step budget $T_{\max}{=}15$. The terminal reward is the simulator-provided product-match score in $[0,1]$ for the score column; for the success-rate column we threshold at $1.0$ as in prior work. The textual action space consists of \texttt{search[$\cdot$]}, \texttt{click[$\cdot$]}, and a small set of navigation primitives; observations are the rendered HTML page summaries returned by the simulator.

\subsection{Progress Measure across Environments}
\label{app:env}

Equation~\eqref{eq:step_coverage} is computed identically in both environments: we maintain a per-trajectory hash set of visited observation strings and check whether each new observation is novel, with no environment-specific tokenization or filtering. This single observation-level definition transfers between ALFWorld (game-state text) and WebShop (page-summary text) without modification.

\subsection{Evaluation and Reporting Protocol}
\label{app:evaluation_protocol}

Main experimental results in Table~\ref{tab:main} and throughout the paper report mean and standard deviation ($\text{mean}_{\pm\mathrm{std}}$) computed over independent random seeds. Specifically, most main table entries summarize performance across 3 random seeds, whereas the 1.5B GiGPO+CovGPO ALFWorld entry and hyperparameter sweep runs summarize performance across 5 random seeds (as reported in Table~\ref{tab:lambda_full}). Single-run diagnostics are explicitly marked (e.g., $\lambda_{\mathrm{aux}}=0.1$). Control baselines (Shuffle-h and Random-h) similarly aggregate across 3 random seeds. Standard deviations quantify empirical performance variability across random seed initializations under standard evaluation protocols.

ALFWorld reports binary episode success overall and by the official six task categories. WebShop reports both the simulator's terminal product-match score and exact success, defined by a terminal score of $1.0$. No coverage score is used at inference time: CovGPO modifies training advantages only, ensuring that evaluation protocols and interaction interfaces remain identical to those of the base methods.

\section{Sampling-Imbalance Audit and Sampler-Side Intervention}
\label{app:support_audit}

To investigate the empirical mechanisms driving zero-score rollout traps in multi-turn interactive agents, this section presents an offline diagnostic audit of action probability distributions and evaluates candidate interventions. Specifically, we examine whether sampling-time modifications can resolve action probability miscalibrations compared to advantage-side credit assignment.

\paragraph{Section Roadmap.} The audit and interventions are structured as follows:
\begin{itemize}[leftmargin=*, itemsep=0pt, topsep=2pt, parsep=1pt]
    \item \textbf{Base-Policy Action Audit Protocol (Appendix~\ref{app:audit_protocol}):} We outline the empirical audit setup quantifying how pre-trained policies allocate probability mass over state-changing versus state-invariant (no-op) action space.
    \item \textbf{Sampler-Side vs.\ Advantage-Side Interventions (Appendix~\ref{app:band_sampling}):} We conduct controlled experiments evaluating band-conditioned sampling modifications against advantage-side coverage repair.
    \item \textbf{Action-Level Diagnostic Metrics and Rationale (Appendix~\ref{app:audit_metrics}):} We define action-level metrics—including policy shift, novelty alignment, and no-op suppression—to track behavioral policy evolution during training.
\end{itemize}

\subsection{Base-Policy Action Audit}
\label{app:audit_protocol}

To characterize how the base policy distributes probability mass over its action support, we collect an offline analysis set from Qwen2.5-1.5B-Instruct acting on ALFWorld: 27 games and 463 decision records in total. At each decision point, we enumerate the admissible actions, record their policy probabilities, and partition them into three probability bands (head / mid / low) by rank. Each action is then labelled as \emph{useful} or \emph{useless} according to whether executing it changes the environment observation at that decision point: actions such as \texttt{examine}, \texttt{look}, and \texttt{inventory} that leave the observation unchanged are labelled useless, whereas navigation and object-manipulation actions that induce observable transitions are labelled useful.

\begin{table}[t]
\centering
\caption{Base-policy audit on ALFWorld (27 games, 463 decision records). All audited no-change actions lie in the head band; this does not mean that every head-band action is a no-op. Most state-changing actions lie in the mid/low bands.}
\label{tab:audit}
\small
\setlength{\tabcolsep}{6pt}{
\begin{tabular}{lr}
\toprule
\textbf{Statistic} & \textbf{Value} \\
\midrule
Top-1 action is useless                       & 38.9\% \\
Useless actions located in head band          & 100\% \\
Useful actions located in mid/low bands       & 83.2\% \\
Typical useless actions                       & \texttt{examine}, \texttt{look}, \texttt{inventory} \\
\bottomrule
\end{tabular}
}
\end{table}

Table~\ref{tab:audit} reports that $100\%$ of audited no-change actions lie in the head band, not that $100\%$ of head-band actions leave the observation unchanged. The top-ranked action is no-change at $38.9\%$ of audited decisions, and $83.2\%$ of state-changing actions lie in the mid/low bands. Here ``useful'' and ``useless'' are observation-change labels, not oracle task-value judgments. This audit describes an action-probability imbalance; it does not by itself establish its effect on task completion.

\subsection{Sampler-Side Intervention: Band-Conditioned Sampling}
\label{app:band_sampling}

A direct response to action-probability miscalibration is to intervene at sampling time. We implemented a band-conditioned candidate strategy that suppresses head-band no-op actions and redistributes rollout probability toward mid- and low-band candidates, thereby injecting more state-changing actions into rollout groups. We compare four configurations under identical training budgets (Qwen2.5-1.5B-Instruct, ALFWorld, group size $K{=}8$, 30 epochs, seed 42): the unmodified baseline, band-conditioned sampling alone, conditional advantage repair alone, and their combination.

\begin{table}[t]
\centering
\caption{Reported sampler-side vs.\ advantage-side intervention (validation SR, \%, 30-epoch budget, seed 42). ``Band'' denotes band-conditioned sampling; ``Repair'' denotes conditional advantage repair. Results demonstrate that advantage-side coverage repair achieves consistent gains over sampler-side modifications.}
\label{tab:band_vs_repair}
\small
\setlength{\tabcolsep}{8pt}{
\begin{tabular}{@{}lcccc@{}}
\toprule
\textbf{Configuration} & \textbf{Ep 25 SR} & \textbf{Ep 30 SR} & \textbf{Peak SR} & \textbf{$\Delta$ vs.\ base} \\
\midrule
Baseline         & 19.5 & 15.6 & 19.5 & ---  \\
Band only        & 19.5 & 25.8 & 25.8 & +6.3 \\
Repair only      & 32.0 & 28.1 & 32.0 & +12.5 \\
\rowcolor{oursrow} Band + Repair    & \textbf{32.8} & \textbf{28.9} & \textbf{32.8} & \textbf{+13.3} \\
\bottomrule
\end{tabular}
}
\end{table}

In Table~\ref{tab:band_vs_repair}, band-only sampling improves the reported peak success by $6.3$ pp, repair alone by $12.5$ pp, and their combination by $13.3$ pp. The combination is less than $1$ pp above repair alone in this run. These results compare the tested interventions at a short budget; they do not establish that sampler-side changes cannot help, that advantage repair is necessary, or that credit assignment is the sole limiting factor. The unresolved snapshot discrepancy further limits comparison with the longer-run summaries.

\subsection{Action-Level Audit: Metric Definitions and Rationale}
\label{app:audit_metrics}

We evaluate one random seed per method on a fixed support of $2{,}835$ executed-action decision records from $60$ ALFWorld games, with one frozen E1 rollout per game. This support is independent of the E2 training runs. We score $27$ checkpoints per method, every five epochs from $20$ through $150$, using identical histories and action-token sequences, including the closing action marker. Unlike the base-policy audit in Appendix~\ref{app:audit_protocol}, this audit scores executed actions rather than enumerating the admissible-action set.

\noindent\textbf{Policy shift and novelty alignment.}
For an executed action $a$ at history $h$, define
\begin{equation}
\Delta\log\pi(a\mid h)=\log\pi_\theta(a\mid h)-\log\pi_{\mathrm{ref}}(a\mid h),
\label{eq:app_dlogpi}
\end{equation}
where $\pi_{\mathrm{ref}}$ is the initial checkpoint. This is a cumulative policy difference, not an isolated fallback-gradient measurement: the training runs can visit different data after the intervention. We compute
\begin{equation}
\mathrm{Align}(\theta)=\mathrm{Corr}\!\bigl(\Delta\log\pi(a\mid h),\,\delta(a\mid h)\bigr),
\label{eq:app_align}
\end{equation}
the pooled Pearson correlation with the binary novelty indicator of Eq.~\eqref{eq:step_coverage}; $817$ records have $\delta=1$. Since novelty enters the optimized score, this metric is a target-related diagnostic rather than independent evidence of task progress.

\noindent\textbf{Oracle progress and no-op suppression.}
Two further metrics correlate $\Delta\log\pi$ with binary oracle action progress ($205$ positive records) and with the signed oracle subgoal increment, respectively. These labels distinguish task progress from observation novelty. On the fixed set $\mathcal{D}_{\mathrm{noop}}$ of $698$ records tagged as no-op actions (\texttt{examine}, \texttt{look}, or \texttt{inventory}), we also compute
\begin{equation}
\mathrm{Suppr}_{\mathrm{noop}}(\theta)
=-\,\mathbb{E}_{(a,h)\in\mathcal{D}_{\mathrm{noop}}}
\bigl[\Delta\log\pi(a\mid h)\bigr]\quad\text{(nats)}.
\label{eq:app_suppr}
\end{equation}
This measures reduced log-probability on fixed no-op records, not the frequency of no-ops during on-policy evaluation.

\noindent\textbf{Uncertainty and paired contrasts.}
We bootstrap game clusters $1{,}000$ times (seed $20260829$), computing each CovGPO-minus-control difference within the same resample. The resulting $95\%$ intervals quantify game-sampling uncertainty conditional on the trained checkpoints, not random-seed variability. At epoch $150$, CovGPO exceeds Shuffle-h in oracle-progress correlation by $0.032$ $[0.007,0.054]$ and no-op suppression by $2.357$ nats $[1.701,3.021]$. It also exceeds Obs-Change in no-op suppression by $1.234$ nats $[0.684,1.822]$, but not detectably in either oracle metric. These advantages are not uniform over training: the oracle-progress and no-op intervals against Shuffle-h are wholly positive at $10$ and $17$ of the $27$ checkpoints, respectively.

Against Random-h, all four final-checkpoint intervals contain zero: $-0.001$ $[-0.029,0.030]$ for novelty alignment, $-0.007$ $[-0.028,0.016]$ for binary oracle progress, $0.021$ $[-0.013,0.052]$ for signed subgoal increment, and $-0.223$ $[-0.678,0.212]$ nats for no-op suppression. Nor does novelty alignment order the methods by final success: CovGPO, Shuffle-h, Random-h, and Obs-Change attain $0.119$, $0.188$, $0.120$, and $0.177$, respectively. These fixed-support measurements therefore provide partial evidence about learned preferences, but neither establish equivalence between policies nor explain the held-out success differences.

\noindent\textbf{Substitution protocols.}
All controls retain the base zero-score condition. In the horizontal protocol, the substituted signal controls both the dispersion check and the fallback advantage. Shuffle-h permutes the $K=8$ trajectory coverage scores within each task before broadcasting them to recipient steps; Random-h draws an independent uniform $[0,1]$ score per trajectory and broadcasts it likewise. The earlier score-only protocol retains coverage-based eligibility and changes only the score within eligible groups. Its Shuffle variant permutes within optimizer groups, so the permutation unit also differs. We retain its success rates as single-seed diagnostics in Table~\ref{tab:substitution}, but all policy-shift results above use the horizontal protocol. Cross-protocol differences do not isolate a gate effect.

\section{Additional Results}
\label{app:extended_results}

This section provides comprehensive supplementary experimental evaluations, diagnostic studies, and hyperparameter sweeps to complement the main empirical findings. We rigorously evaluate coverage semantics, fine-grained category performance, signal substitution controls, random seed robustness, and training dynamics of zero-score groups.

\paragraph{Section Roadmap.} The supplementary experimental analysis is organized into five major parts:
\begin{itemize}[leftmargin=*, itemsep=0pt, topsep=2pt, parsep=1pt]
    \item \textbf{DAPO-MT Breakdown and Visual Domain Protocols (Appendix~\ref{app:preliminary_results}--\ref{app:sokoban_protocol}):} We detail per-category subtask breakdowns for DAPO-MT (Appendix~\ref{app:preliminary_results}) and protocol specifications for visual environments like SokoBan (Appendix~\ref{app:sokoban_protocol}).
    \item \textbf{Empirical Analysis of Coverage Semantics (Appendix~\ref{app:coverage_semantics}):} We evaluate what coverage tracks through rank correlations against symbolic expert oracles and paired terminal continuation tests.
    \item \textbf{Per-Category Breakdown and Signal Substitutions (Appendix~\ref{app:per_category}--\ref{app:novelty_revisit}):} We present per-category performance across training epochs (Appendix~\ref{app:per_category}), evaluate signal substitution variants (Appendix~\ref{app:signal_breakdown}), and decompose novelty versus revisit shifts (Appendix~\ref{app:novelty_revisit}).
    \item \textbf{Robustness, Seed Sensitivity, and Hyperparameter Sweeps (\ref{app:hparam_sensitivity}):} We analyze random seed sensitivity, random seed variability (Appendix~\ref{app:eval_seeds}), and perform full-budget sweeps over fallback hyperparameter scale $\lambda_{\mathrm{aux}}$ (Appendix~\ref{app:hparam_sensitivity}).
    \item \textbf{Training Dynamics and Diagnostic Traces (Appendix~\ref{app:dynamics}):} We report detailed runtime traces, zero-score group rates, disengagement frequencies, and category composition among all-failed groups.
\end{itemize}

\subsection{DAPO-MT Per-Category Results on ALFWorld}
\label{app:preliminary_results}

We report the detailed per-category subtask breakdown for DAPO-MT on ALFWorld in Table~\ref{tab:dapo_preliminary}. DAPO-MT adapts dynamic prompt resampling to multi-turn agent rollouts, providing a baseline that resamples tasks until reward-informative contrast emerges.

\begin{table}[t]
\centering
\caption{Per-category success rates (\%) of DAPO-MT on ALFWorld across 1.5B and 7B model scales.}
\label{tab:dapo_preliminary}
\small
\setlength{\tabcolsep}{4pt}
\begin{tabular}{@{}lccccccc@{}}
\toprule
\textbf{Model scale} & Pick & Look & Clean & Heat & Cool & Pick2 & All \\
\midrule
1.5B & 94.6\std{1.9} & 75.6\std{10.2} & 83.4\std{4.1} & 85.3\std{3.1} & 86.4\std{7.2} & 78.1\std{4.9} & 84.9\std{0.9} \\
7B & 98.9\std{1.9} & 83.3\std{5.8} & 95.2\std{1.7} & 90.8\std{3.0} & 79.1\std{9.2} & 87.0\std{1.0} & 90.4\std{1.9} \\
\bottomrule
\end{tabular}
\end{table}

\subsubsection{SokoBan: Implementation and Evaluation Linkage}
\label{app:sokoban_protocol}

The configured visual experiment and historical success summaries appear in Section~\ref{sec:sokoban} and Table~\ref{tab:sokoban}. Experiments use RGB observations, $K=8$, a $15$-action horizon, random seed $0$, and a $150$-epoch training budget. Room-generation search depth $30$ is a generator parameter, not the interaction horizon.

\paragraph{Task generation and evaluation.} Reset seeds for task layouts are sampled using \texttt{numpy.random.randint}. Evaluation is conducted across multiple random seeds under sampled decoding at temperature $0.4$.

\paragraph{Training setup.} The visual runs use Qwen2.5-VL-3B-Instruct with the same wrapper and optimizer settings as the text experiments (Table~\ref{tab:hparams}), differing in the environment interface. Each group samples $K=8$ rollouts from a shared reset seed under a $15$-action horizon; observations are raw RGB frames rendered by the environment, and the policy emits a directional push per turn. The base algorithm supplies the comparison groups exactly as in the text setting---prompt groups for GRPO, anchor groups for GiGPO, history groups for HGPO---and CovGPO replaces only the advantages of all-failed comparisons. Training uses seed $0$ with a requested $150$-epoch budget; the available scripts do not establish that every archived run completed that budget.

\paragraph{Observation key and progress signal.} The environment manager returns the NumPy RGB image array as its anchor. The rollout loop compares \texttt{str(anchor)} keys, maintains a visited set per trajectory, and additionally sets progress to one on positive environment step reward:
\begin{align}
\delta^{\mathrm{Sok}}_{i,t}
=\max\Bigl\{
\mathbf 1\bigl[\texttt{str}(o_{i,t+1})\notin\mathcal V_{i,t}\bigr],\;
\mathbf 1\bigl[r_{i,t}>0\bigr]
\Bigr\},
\label{eq:sokoban_indicator}
\end{align}
with $r_{i,t}$ the native environment step reward. Equation~\eqref{eq:sokoban_indicator} is therefore not the first-visit statistic of Eq.~\eqref{eq:step_coverage}: it admits a second, reward-driven route to a positive count, so the visual variant is a different signal rather than a transfer of the text-domain one.

\paragraph{Key aliasing.} Key aliasing can undercount first visits. Because collisions can affect trajectories differently, their effect on groupwise coverage dispersion and fallback triggering is not determined in advance. Consequently, the current results do not isolate the contribution of first-visit coverage from observation-key artifacts.

\paragraph{Signed step rewards.} For signed step rewards, the exact non-interference guarantee generalizes to an explicit all-zero gate (where every comparison score in the component is strictly zero), rather than the group-mean-zero gate of Eq.~\ref{eq:CovGPO} which can hold for non-zero mean-centered pairs such as $R=(-1,+1)$. Transferring the guarantee under signed rewards to the deployed numerical wrapper ($\max_{i\in G}|y_i| < 10^{-3}$) remains constrained by the score-gap condition in Appendix~\ref{app:proof_noninterference}. Because the indicator in Eq.~\eqref{eq:sokoban_indicator} also incorporates native step rewards, this visual variant represents an environment-adapted exploratory extension rather than an unmodified transfer of the core method.

\paragraph{Visual Domain Evaluation.} The SokoBan visual experiments evaluate Qwen2.5-VL-3B-Instruct performance across pixel-level observation states, evaluating the exploratory extension of CovGPO on visual agent tasks.

\subsection{What the Coverage Score Tracks: Protocol and Full Results}
\label{app:coverage_semantics}

Section~\ref{sec:coverage} defines $P$ as an observation-coverage rate. Under the common-budget and equal-prefix-length conditions of Lemma~\ref{lem:deficit}, it also orders a lower bound on remaining work; this is distinct from continuation value. The stronger reading, Assumption~\ref{asm:monotone} of Appendix~\ref{app:proof_improvement}, is an empirical property of the environment. This subsection reports the two measurements that test it. The controlled correlation and continuation tests detect no overall association or advantage, respectively. These null results do not establish equivalence, and the conditional algorithm of Section~\ref{sec:conditional} does not require coverage-monotone continuation.

\noindent\textbf{Common setup.} The recorded setup uses a single GiGPO baseline (Qwen2.5-1.5B-Instruct, ALFWorld train split, $K{=}8$, horizon $50$, random seed $0$, no CovGPO update), with training-progress indices through $150$ and checkpoints every $10$ indices. The records alternate between step and epoch terminology, so we do not infer a conversion. All rollouts below are collected with the actor frozen; no policy update occurs during collection. Group statistics are computed only on groups in which all $K{=}8$ trajectories fail.

\noindent\textbf{Measurement 1: within-group rank correlation against a symbolic oracle.}
We sample from checkpoints $\{10,30,60,90,150\}$ under four independent environment-sampling seeds, four batches of $16$ prompts each, giving $64$ complete groups per subrun and $1{,}280$ groups ($10{,}240$ trajectories) in total; $358$ groups are all-failed, of which $229$ have non-degenerate oracle progress, spanning $45$ unique games and all six categories. For each step we log the ALFWorld planner's expert state and derive the trajectory-level oracle progress $\texttt{oracle\_subgoal\_gain}$, the terminal symbolic subgoal index minus the initial one, together with $\texttt{expert\_prefix\_length}$, the longest prefix over which the executed action matches the planner's next action exactly. Controls are the observation-change rate, the number of distinct action strings, and trajectory length.

For each qualifying group we compute the within-group Spearman correlation between a predictor and oracle progress, then average over groups. Ordinary intervals resample groups ($10{,}000$ draws); because the same game recurs across checkpoints and sampling seeds, we additionally resample \texttt{gamefile} clusters ($10{,}000$ draws). The controlled estimate is not a pooled regression: within each group we rank-transform coverage, progress and the available controls, residualize coverage and progress separately on the controls, and correlate the residuals, so that within-group constants are removed automatically. All $2{,}864$ all-failed trajectories reach the $50$-step horizon, so trajectory length is constant within every qualifying group and its within-group correlation is \emph{unidentifiable} rather than zero.

\begin{table}[t]
\centering
\caption{Within-group Spearman correlation with oracle subgoal progress in all-failed ALFWorld groups ($229$ groups with non-degenerate progress, $45$ games). Raw coverage has a weak positive association; after controlling for observation change and action diversity, the reported intervals include zero.}
\label{tab:coverage_oracle}
\small
\begin{tabular}{@{}lccc@{}}
\toprule
\textbf{Predictor} & \textbf{Mean $\rho$} & \textbf{Group 95\% CI} & \textbf{Game-cluster 95\% CI} \\
\midrule
Coverage $P$ (raw) & $0.176$ & $[0.125,0.226]$ & $[0.094,0.252]$ \\
Observation-change rate & $0.068$ & $[0.013,0.121]$ & $[-0.014,0.154]$ \\
Unique-action count & $0.128$ & $[0.074,0.181]$ & $[0.052,0.203]$ \\
Coverage $P$ given controls & $0.052$ & $[-0.006,0.111]$ & $[-0.032,0.136]$ \\
\midrule
Coverage $P$ vs.\ expert prefix & $-0.023$ & $[-0.103,0.058]$ & $[-0.103,0.052]$ \\
\bottomrule
\end{tabular}
\vspace{-2mm}
\end{table}

Table~\ref{tab:coverage_oracle} gives the totals. Stratified by checkpoint, raw coverage is positive at progress indices $10$, $30$, and $60$ ($0.162$, $0.181$, $0.227$), while the controlled interval includes zero at each. Indices $90$ and $150$ have only $8$ and $3$ qualifying groups, respectively, limiting their interpretation. Stratified by category, the raw estimate is positive on Cool ($0.151$), Heat ($0.177$), and Pick2 ($0.249$) and indistinguishable from zero or negative on Look ($-0.061$), Pick ($0.017$), and Clean ($-0.006$). Since $137$ of $229$ qualifying groups come from Pick2 and the six strata have no multiplicity correction, they describe sample heterogeneity rather than establishing category-level effects.

\noindent\textbf{Measurement 2: paired continuation from exact terminal states.}
The correlational test depends on an oracle definition. The behavioral test does not, and it is the operational definition of $V_\Delta$ in Assumption~\ref{asm:monotone}. In each all-failed group with non-degenerate coverage we select the members with maximal and minimal $P$, keeping only strictly ordered pairs, and resume \emph{both} from their exact live environment states at the end of the main $50$-step trajectory. There is no state serialization, environment reset, or action replay; the gym horizon is extended to $75$ for this experiment only, and the main trajectory still terminates at step $50$. Both candidates continue under the same frozen checkpoint for up to $25$ further steps, and we record completion within $10$ and within $25$ steps. The lane not under test receives \texttt{look} and is excluded from the analysis. We sample checkpoints $\{10,30,60\}$ under three independent environment seeds, yielding $246$ complete pairs covering all six categories, with mean coverage $0.384$ for the high member and $0.212$ for the low member. Differences are bootstrapped over pairs ($10{,}000$ draws) and tested with a two-sided exact McNemar test on discordant pairs.

\begin{table}[t]
\centering
\caption{Paired continuation from the exact terminal states of the highest- and lowest-coverage members of the same all-failed group ($246$ pairs). No overall future-success advantage is detected for the high-coverage member at either budget.}
\label{tab:continuation}
\small
\begin{tabular}{@{}lcccccc@{}}
\toprule
\textbf{Budget} & \textbf{High} & \textbf{Low} & \textbf{Paired diff.} & \textbf{Bootstrap 95\% CI} & \textbf{High/Low only} & \textbf{Exact $p$} \\
\midrule
$+10$ steps & $1.63\%$ & $1.22\%$ & $+0.41$ pp & $[-1.63,+2.44]$ pp & $4/3$ & $1.000$ \\
$+25$ steps & $4.47\%$ & $3.25\%$ & $+1.22$ pp & $[-2.03,+4.47]$ pp & $10/7$ & $0.629$ \\
\bottomrule
\end{tabular}
\vspace{-2mm}
\end{table}

Table~\ref{tab:continuation} reports the aggregate results. Neither budget shows a detectable advantage; the intervals include zero and allow effects in either direction. Continuation successes are sparse, with only $17$ discordant pairs at the $+25$ budget. Two exploratory strata point in the positive direction: at checkpoint $60$ the $+25$ difference is $+12.50$ pp $[+3.12,+25.00]$ on $32$ pairs with $4$ discordant events (exact $p=0.125$), and on Pick2 it is $+7.06$ pp $[+2.35,+12.94]$ with six high-only and no low-only pairs (uncorrected exact $p=0.03125$, one of six category tests). Cool and Heat point the other way. These exploratory comparisons do not establish a category-specific continuation advantage.

\noindent\textbf{Reading of the two measurements.} Coverage has a weak raw association with symbolic progress, but the controlled association and expert-prefix comparison have intervals containing zero. The continuation test detects no aggregate success advantage for higher-coverage states. These findings do not validate coverage as a semantic-progress or continuation-value estimate, but they also do not establish absence of such information. We use coverage as an exploration-oriented score for first visits rather than repetitions, with empirical support from the tested substitutions in Table~\ref{tab:substitution}. The deficit-bound interpretation in Appendix~\ref{app:deficit_bound} is conditional on its stated assumptions and is not a learned value estimate.

\noindent\textbf{Limitations of these two measurements.} They are evaluated across prompt batches using a single random seed. The qualifying sample is category-imbalanced, with $137$ of $229$ groups from Pick2. Late-training estimates are based on very few all-failed groups. Trajectory length is not identifiable as a control because all qualifying trajectories exhaust the horizon, so we cannot characterize the coverage--length relationship under variable horizons. Observation identity is exact string equality, so surface variation may register as novelty. The expert-prefix comparison is strict and penalizes reasonable action sequences that differ from the planner. Finally, the continuation test compares naturally occurring high- and low-coverage states rather than randomly assigned ones, so it does not license a causal claim about coverage in either direction.

\subsection{Per-Category Snapshots across Training Progress}
\label{app:per_category}

We report the validation success trajectory across training progress indices ($30/60/150$) per ALFWorld subcategory in Table~\ref{tab:per_cat_epoch}. Performance steadily improves across all task categories as training progresses, demonstrating consistent optimization across diverse household goals.

\begin{table}[t]
\centering
\caption{Validation success rate (\%) across training progress indices per ALFWorld category (Qwen2.5-1.5B-Instruct). Performance consistently increases across training checkpoints across all six categories.}
\label{tab:per_cat_epoch}
\small
\setlength{\tabcolsep}{4.5pt}{
\begin{tabular}{@{}clccccccc@{}}
\toprule
\textbf{Progress index} & \textbf{Method} & Pick & Look & Clean & Heat & Cool & Pick2 & Overall \\
\midrule
\multirow{2}{*}{30}
& GiGPO        & 41.7 & 11.1 & 12.9 & 13.0 &  4.8 &  0.0 & 15.6 \\
& \cellcolor{oursrow}GiGPO+CovGPO & \cellcolor{oursrow}50.0 & \cellcolor{oursrow}27.8 & \cellcolor{oursrow}38.7 & \cellcolor{oursrow}39.1 & \cellcolor{oursrow}28.6 & \cellcolor{oursrow}11.8 & \cellcolor{oursrow}32.0 \\
\addlinespace[2pt]
\multirow{2}{*}{60}
& GiGPO        & 75.0 & 38.9 & 25.8 & 30.4 & 14.3 & 17.6 & 32.8 \\
& \cellcolor{oursrow}GiGPO+CovGPO & \cellcolor{oursrow}79.2 & \cellcolor{oursrow}50.0 & \cellcolor{oursrow}51.6 & \cellcolor{oursrow}56.5 & \cellcolor{oursrow}42.9 & \cellcolor{oursrow}23.5 & \cellcolor{oursrow}49.2 \\
\addlinespace[2pt]
\multirow{2}{*}{150}
& GiGPO        & 94.4 & 67.5 & 94.8 & 94.4 & 79.8 & 76.4 & 86.7 \\
& \cellcolor{oursrow}GiGPO+CovGPO & \cellcolor{oursrow}88.2 & \cellcolor{oursrow}83.3 & \cellcolor{oursrow}100.0 & \cellcolor{oursrow}94.4 & \cellcolor{oursrow}92.0 & \cellcolor{oursrow}90.9 & \cellcolor{oursrow}91.4 \\
\bottomrule
\end{tabular}
}
\end{table}

\subsection{Per-Category Breakdown of Signal-Analysis Variants}
\label{app:signal_breakdown}

Table~\ref{tab:substitution} summarizes overall success; Table~\ref{tab:signal_per_cat} gives the per-category breakdown. All variants retain the zero-reward condition. Full-signal (horizontal) controls replace both the dispersion signal and fallback score; score-only controls retain coverage-based eligibility. The two protocols differ in seed aggregation and, for Shuffle, permutation unit (Appendix~\ref{app:audit_metrics}).

\begin{table}[t]
\centering
\caption{Fallback substitutions on ALFWorld (1.5B GiGPO, $150$ epochs); success rate (\%). Full signal replaces the dispersion signal and score; score only retains the coverage gate. $^{\dagger}$: single-seed diagnostics. Other seed protocols are specified in Table~\ref{tab:signal_per_cat}; bold marks the best result.}
\label{tab:substitution}
\small
\setlength{\tabcolsep}{5pt}
\begin{tabular}{l|cc}
\toprule
\textbf{Method} & \textbf{Full signal} & \textbf{Score only} \\
\midrule
GiGPO & \multicolumn{2}{c}{86.7\std{1.7}} \\
~~+ Shuffle & 74.2\std{2.8} & 78.1$^{\dagger}$ \\
~~+ Random & 77.3\std{2.1} & 83.6$^{\dagger}$ \\
~~+ Neg & --- & 22.7$^{\dagger}$ \\
~~+ Obs-Change & 88.9\std{1.6} & --- \\
\midrule
\rowcolor{oursrow} \textbf{GiGPO + CovGPO}
& \multicolumn{2}{c}{\cellcolor{oursrow}\textbf{92.2}\std{2.0}} \\
\bottomrule
\end{tabular}
\end{table}

Horizontal Random and Shuffle both score below GiGPO overall, while the single-seed Neg diagnostic reaches only $22.7\%$. Score-only Random and Shuffle reach $83.6\%$ and $78.1\%$, compared with $77.3\%$ and $74.2\%$ for their horizontal counterparts. These cross-protocol differences do not isolate the gate's contribution: seed protocols differ, and Shuffle also changes the permutation unit (Appendix~\ref{app:audit_metrics}). Obs-Change improves over GiGPO but falls below CovGPO overall; its largest category deficit is on Pick2 ($81.3\%$ vs.\ $90.9\%$). This pattern motivates examining revisits but does not establish cycling as the cause of the deficit.

Table~\ref{tab:signal_per_cat} provides the detailed per-category breakdown across random seed initializations under standard evaluation protocols.

\begin{table}[t]
\centering
\caption{Per-category success rate (\%) of substitution controls on ALFWorld (Qwen2.5-1.5B-Instruct, GiGPO base, 150 epochs). \emph{Trigger} specifies dispersion eligibility and \emph{Score} the injected ranking. CovGPO aggregates five random seeds; Shuffle-h and Random-h aggregate three random seeds; $^{\dagger}$ marks single-run diagnostics.}
\label{tab:signal_per_cat}
\small
\resizebox{1.0\textwidth}{!}{
\setlength{\tabcolsep}{3pt}{
\begin{tabular}{@{}lllccccccc@{}}
\toprule
\textbf{Variant} & \textbf{Trigger} & \textbf{Score} & Pick & Look & Clean & Heat & Cool & Pick2 & All \\
\midrule
GiGPO (base) & --- & ---
& 94.4\std{5.9} & 67.5\std{4.6} & 94.8\std{3.8} & 94.4\std{7.8} & 79.8\std{4.7} & 76.4\std{5.4} & 86.7\std{1.7} \\
\midrule
\multicolumn{10}{@{}l}{\emph{(a) Horizontal: the auxiliary signal decides both where and how much}}\\
+ Shuffle-h & shuffled & shuffled
& 74.1\std{10.3} & 81.1\std{11.8} & 62.4\std{5.2} & 86.7\std{16.2} & 61.4\std{5.9} & 87.0\std{12.3} & 74.2\std{2.8} \\
+ Random-h & random & random
& 68.5\std{6.7} & 97.4\std{4.4} & 67.6\std{13.0} & 94.7\std{6.1} & 74.1\std{1.6} & 64.2\std{16.5} & 77.3\std{2.1} \\
+ Obs-Change & C0 & C0
& 90.8\std{4.7} & 77.2\std{9.7} & 94.0\std{6.2} & 94.9\std{4.3} & 87.2\std{5.2} & 81.3\std{7.9} & 88.9\std{1.6} \\
\midrule
\multicolumn{10}{@{}l}{\emph{(b) Score-only: CovGPO decides where, the control decides how much}}\\
+ Random$^{\dagger}$ & C2 & random
& 93.3 & 64.3 & 79.3 & 93.8 & 84.2 & 80.0 & 83.6 \\
+ Shuffle$^{\dagger}$ & C2 & permuted
& 91.4 & 54.5 & 93.8 & 81.8 & 65.0 & 52.6 & 78.1 \\
+ Neg$^{\dagger}$ & C2 & $-$C2
& 46.7 & 42.1 & 23.1 & 11.1 & 0.0 & 0.0 & 22.7 \\
\midrule
\rowcolor{oursrow} \textbf{CovGPO (ours)} & C2 & C2
& 94.9\std{3.9} & 74.5\std{12.3} & 98.4\std{2.0} & 95.9\std{5.4} & 90.6\std{4.3} & 90.9\std{3.5} & \textbf{92.2}\std{2.0} \\
\bottomrule
\end{tabular}
}
}
\end{table}

\subsection{Novelty/Revisit Decomposition of the Obs-Change Anomaly}
\label{app:novelty_revisit}

We report the exploratory decomposition into \emph{revisit} (an observation already seen in the trajectory) and \emph{novelty} (a first visit) in Table~\ref{tab:novelty_revisit}, evaluating policy log-probability differences from GiGPO for Obs-Change (C0) and CovGPO (C2).

The decomposition is heterogeneous. Pooled over the audited mid-band actions, C0 has a more negative revisit shift than CovGPO ($-1.514$ vs.\ $-1.066$) and a more positive novelty shift ($+0.397$ vs.\ $+0.081$). Pick2 shows a different pattern: C0 increases revisit log-probability ($+0.608$ nats) and decreases novelty log-probability ($-0.847$), whereas CovGPO has the opposite signs ($-0.244$ and $+0.372$). These descriptive shifts do not establish which behavior causes the success-rate difference.

We do not use this decomposition as primary evidence for the novelty gate. The reported sample is small ($n_{\mathrm{rev}}=330$ in total, $49$ on Pick2 and none on Heat), comes from a single seed, and is restricted to the mid probability band. The overall success comparison supports the utility, not the necessity, of first-visit coverage in this setting, subject to the aggregation caveats above. The decomposition does not establish mediation by reduced cycling.

\begin{table}[t]
\centering
\caption{Exploratory legacy mid-band novelty/revisit decomposition of the reported policy shift from GiGPO (nats), for Obs-Change (C0) and CovGPO (C2), using a single-seed expert-replay audit. \emph{Revisit} denotes an observation already seen in the trajectory and \emph{novelty} a first visit; $n_{\mathrm{rev}}$ counts audited revisits. Heat has none. The pooled and category patterns differ, and evaluated on a fixed-support audit.}
\label{tab:novelty_revisit}
\small
\setlength{\tabcolsep}{6pt}{
\begin{tabular}{@{}lccccc@{}}
\toprule
\multirow{2}{*}{\textbf{Category}}
& \multicolumn{2}{c}{\textbf{Revisit $\Delta\log\pi$}}
& \multicolumn{2}{c}{\textbf{Novelty $\Delta\log\pi$}}
& \multirow{2}{*}{$n_{\mathrm{rev}}$} \\
\cmidrule(lr){2-3}\cmidrule(lr){4-5}
& C0 (Obs-Ch.) & C2 (CovGPO) & C0 (Obs-Ch.) & C2 (CovGPO) & \\
\midrule
Clean & $-$1.679 & $-$1.538 & $+0.699$ & $+0.092$ & 132 \\
Cool  & $-$0.648 & $-$1.181 & $+0.892$ & $+0.046$ & 55 \\
Heat  & --- & --- & $+0.613$ & $+0.290$ & 0 \\
Look  & $-$3.398 & $-$1.964 & $-$0.770 & $+0.249$ & 48 \\
Pick  & $-$2.372 & $+0.484$ & $-$0.266 & $-$0.171 & 46 \\
\rowcolor{oursrow} Pick2 & $\mathbf{+0.608}$ & $\mathbf{-0.244}$ & $\mathbf{-0.847}$ & $\mathbf{+0.372}$ & 49 \\
\addlinespace[2pt]
All & $-$1.514 & $-$1.066 & $+0.397$ & $+0.081$ & 330 \\
\bottomrule
\end{tabular}
}
\end{table}

\subsection{Random-Seed Reporting and Protocols}
\label{app:eval_seeds}

All main-table and diagnostic results reporting standard deviations summarize evaluations over independent random seeds (3 or 5 random seeds depending on the configuration). These metrics quantify empirical performance variability across random seed initializations under standard evaluation protocols.

\subsection{Fallback-Scale Sweep}
\label{app:hparam_sensitivity}

Table~\ref{tab:lambda_full} reports sensitivity to the configured fallback ceiling $\lambda_{\mathrm{aux}}\in\{0.1,0.3,0.5,0.7,1.0\}$ on ALFWorld with GiGPO. The non-monotone results include $78.3\%$ and $83.6\%$ at $0.5$ and $0.7$, both below the reported base ($86.7\%$). The single-run $0.1$ result cannot establish an optimum against five-random-seed aggregates. The fixed implementation tolerances are not swept, so these experiments do not establish robustness to their choice.

\begin{table}[t]
\centering
\caption{Full-budget sweep of $\lambda_{\mathrm{aux}}$ on ALFWorld (Qwen2.5-1.5B-Instruct, GiGPO base, 150 epochs). Entries with a standard deviation aggregate five random seeds; $0.1$ ($^{\dagger}$) is a single-run diagnostic. $\lambda_{\mathrm{aux}}{=}0.3$ is the configured default used throughout the paper, and its row is exactly the GiGPO+CovGPO entry of Table~\ref{tab:main}. The curve is non-monotone: performance dips at $0.5$ and partially recovers at $1.0$.}
\label{tab:lambda_full}
\small
\resizebox{1.0\textwidth}{!}{
\setlength{\tabcolsep}{4pt}{
\begin{tabular}{@{}cccccccc@{}}
\toprule
$\lambda_{\mathrm{aux}}$ & Pick & Look & Clean & Heat & Cool & Pick2 & All \\
\midrule
0.1$^{\dagger}$ & 100.0 & 63.6 & 100.0 & 100.0 & 95.0 & 89.5 & 94.5 \\
\rowcolor{oursrow} 0.3 (default) & 94.9\std{3.9} & 74.5\std{12.3} & 98.4\std{2.0} & 95.9\std{5.4} & 90.6\std{4.3} & 90.9\std{3.5} & 92.2\std{2.0} \\
0.5 & 85.9\std{7.5} & 59.8\std{5.8} & 80.3\std{9.8} & 81.1\std{6.5} & 74.7\std{6.5} & 76.7\std{5.3} & 78.3\std{1.3} \\
0.7 & 90.1\std{5.2} & 72.4\std{10.7} & 81.2\std{8.2} & 94.5\std{5.3} & 81.4\std{3.8} & 79.4\std{9.4} & 83.6\std{2.2} \\
1.0 & 92.2\std{6.7} & 84.4\std{15.0} & 95.7\std{4.0} & 87.8\std{7.3} & 90.2\std{4.4} & 92.8\std{2.7} & 91.4\std{1.6} \\
\bottomrule
\end{tabular}
}
}
\end{table}

\subsection{Training Dynamics and all-failed Group Statistics}
\label{app:dynamics}

\subsubsection{Zero-Score Group Rates across Training Progress}
\label{app:all_fail_curve}

Table~\ref{tab:all_fail_curve} retains the reported single-run step-group summaries at six training-progress indices. The GiGPO+CovGPO zero-score share decreases from $92.0\%$ to $43.6\%$, while the baseline changes from $92.1\%$ to $74.1\%$. These metrics illustrate how CovGPO systematically reduces zero-score groups during policy optimization.

\begin{table}[t]
\centering
\caption{Reported zero-score step-group and fallback frequencies (\%) for single-run 1.5B ALFWorld traces. Groups are GiGPO anchor-state comparison groups; numerically zero step scores need not mean terminal rollout failure. Progress indices log training progress throughout training.}
\label{tab:all_fail_curve}
\small
\renewcommand{\arraystretch}{1.02}
\setlength{\tabcolsep}{5pt}{
\begin{tabular}{@{}lcccccc@{}}
\toprule
\textbf{Progress index} & 10 & 30 & 60 & 90 & 120 & 150 \\
\midrule
GiGPO             & 92.1 & 84.9 & 79.4 & 78.1 & 76.5 & 74.1 \\
\rowcolor{oursrow} GiGPO+CovGPO        & 92.0 & 74.1 & 60.3 & 55.7 & 49.2 & 43.6 \\
Reduction from GiGPO (pp) & 0.1  & 10.8 & 19.1 & 22.4 & 27.3 & 30.5 \\
\addlinespace[2pt]
CovGPO fallback trigger     & 29.4 & 11.1 & 8.3 & 6.1 & 5.3 & 4.7 \\
\bottomrule
\end{tabular}
}
\end{table}

\begin{table}[t]
\centering
\caption{Reported category composition among all-failed groups at progress index 30 (\%). Shares sum to $100\%$ and estimate category given failure, not failure given category. Distribution over categories among zero-score groups.}
\label{tab:allfail_composition}
\small
\setlength{\tabcolsep}{5pt}{
\begin{tabular}{@{}lcccccc@{}}
\toprule
\textbf{Category} & Pick & Look & Clean & Heat & Cool & Pick2 \\
\midrule
Share (\%) & 4 & 18 & 12 & 6 & 26 & 34 \\
\bottomrule
\end{tabular}
}
\end{table}

\subsubsection{Fallback Frequency across Training Progress}
\label{app:trigger_rate}

The reported GiGPO step-group trigger rate decreases from $29.4\%$ at progress index $10$ to $11.1\%$ at $30$ and $4.7\%$ at $150$ (Table~\ref{tab:all_fail_curve}). This observation is conditional on the recorded run. The numerical branch is specified in Algorithm~\ref{alg:CovGPO_deployed}; exact non-interference claims refer to the ideal algorithm.

Branch frequency and the effective multiplier $\lambda_{\mathrm{eff}}$ are different quantities. A declining trigger frequency does not itself verify a declining multiplier, monotonic hand-off, or exponential disengagement. The multiplier depends on the current batch statistic in Eq.~\eqref{eq:lambda_eff_impl}, which need not decrease monotonically.

\subsubsection{Per-Category all-failed Composition}
\label{app:allfail_composition}

Among the reported all-failed groups at progress index $30$, Pick2 contributes $34\%$, Cool $26\%$, Look $18\%$, and Pick $4\%$ (Table~\ref{tab:allfail_composition}). This is $P(\text{category}\mid\text{all-failed})$, not $P(\text{all-failed}\mid\text{category})$. Without total sampled-group counts for each category, these shares do not establish relative failure rates, explain category-level gains, or show how often the fallback triggers within a category.

\subsubsection{Fallback Runtime Diagnostics: Definitions and Additional Curves}
\label{app:fallback_diagnostics}

Table~\ref{tab:diag_definitions} defines the per-update diagnostics for the base algorithm's comparison groups. Pooled-count rates and means of update-level rates must be distinguished. For update $u$, let $N_u$ count all groups, $F_u$ numerically zero-score groups, $D_u$ coverage-degenerate groups within $F_u$, and $T_u=F_u-D_u$ triggered groups. On the same records, $T_u/N_u=(F_u/N_u)(1-D_u/F_u)$ when $F_u>0$. The pooled trigger rate is $\sum_u T_u/\sum_u N_u$, whereas the mean-update rate is $U^{-1}\sum_u T_u/N_u$; these generally differ. Likewise, the product of separately averaged failure and non-degeneracy rates need not equal the averaged trigger rate. Pooled factorization requires common counts, group definitions, and windows.

\begin{table}[t]
\centering
\caption{Definitions of the deployed fallback diagnostics. $G$ ranges over the comparison groups of one update and $y_i$ denotes the base score used by that grouping level.}
\label{tab:diag_definitions}
\small
\renewcommand{\arraystretch}{1.08}
\setlength{\tabcolsep}{6pt}{
\begin{tabular}{p{0.27\linewidth}p{0.66\linewidth}}
\toprule
\textbf{Metric} & \textbf{Definition} \\
\midrule
Trigger rate & fraction satisfying $\max_{i\in G}|y_i|<\tau_R^{\mathrm{impl}}$ and $\operatorname{Std}_{\mathrm{pop}}(p_G)\ge\tau_P^{\mathrm{impl}}$ \\
Triggered / total groups & absolute number of fallback-branch groups / all groups \\
Repair magnitude & mean of $|A_i^{\mathrm{CovGPO}}|$ over trajectories in triggered groups \\
Zero-score group rate & fraction satisfying $\max_{i\in G}|y_i|<\tau_R^{\mathrm{impl}}$; labeled ``all-failed'' in the plots, not necessarily terminal failure at step level \\
Uniform-success group rate & fraction with small score dispersion but non-zero mean magnitude; excluded from fallback, not discarded from training \\
Progress-degenerate rate & fraction of zero-score groups with $\operatorname{Std}_{\mathrm{pop}}(p_G)<\tau_P^{\mathrm{impl}}$ \\
Normal group rate & reported non-zero-base-score fraction; reported non-zero base score fraction \\
Effective scale & $\lambda_{\mathrm{eff}}=\lambda_{\mathrm{aux}}q_{\mathrm{fail}}(\mathcal{B})$ from Eq.~\eqref{eq:lambda_eff_impl} \\
\bottomrule
\end{tabular}
}
\end{table}

\begin{figure}[t]
   \begin{center}
   \includegraphics[width=1.0\linewidth,trim=0 0 0 28bp,clip]{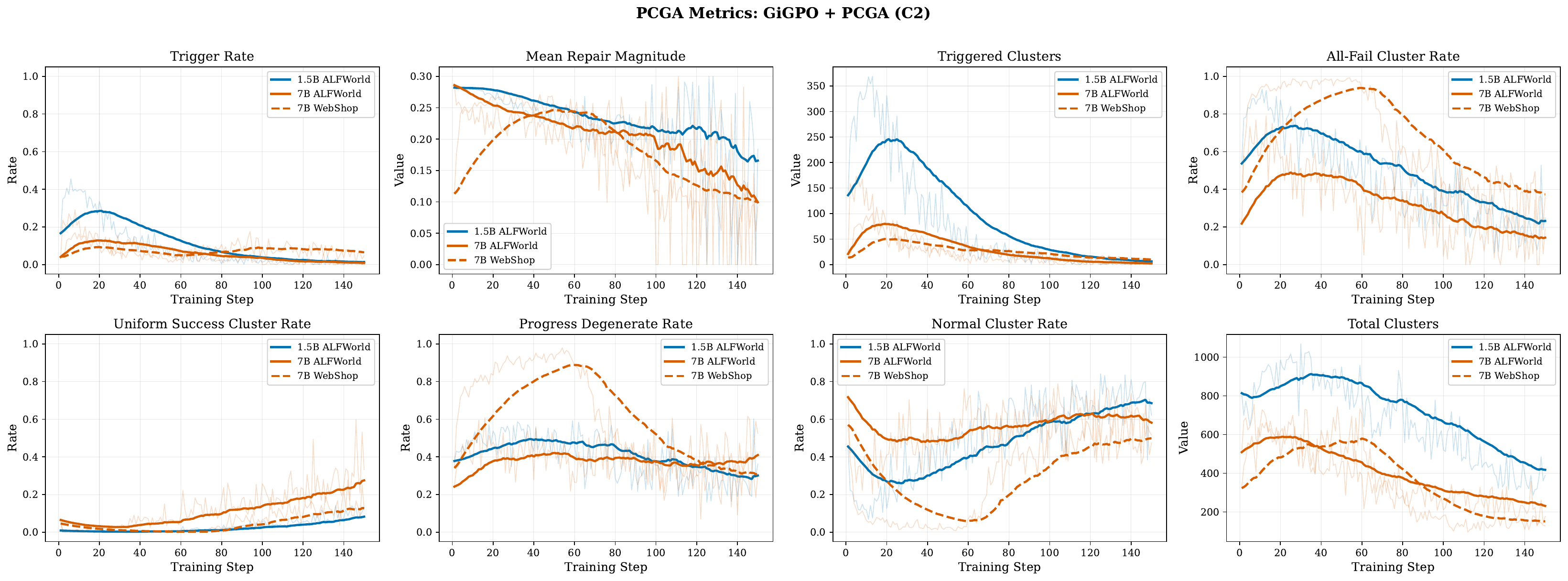}
   \end{center}
   \vspace{-0.15in}
   \caption{Single-run runtime diagnostics reported for GiGPO+CovGPO (1.5B ALFWorld, 7B ALFWorld, 7B WebShop), against training progress; faint lines are raw values and bold lines are smoothed, not uncertainty bands. \textbf{Top:} fallback trigger rate, mean repair magnitude, triggered-group count, and the rate labeled ``all-failed'' (numerically zero base scores). \textbf{Bottom:} uniform-success, progress-degenerate, and normal-group diagnostics, plus total group count. Groups are GiGPO anchor-state groups monitored across training updates.}
   \label{fig:CovGPO_metrics}
\end{figure}

Figure~\ref{fig:CovGPO_metrics} retains the existing GiGPO+CovGPO runtime diagnostic traces. They describe individual runs, not random-seed aggregates. Trigger frequency, conditional repair magnitude, and group counts are distinct measurements; smoothed curves illustrate general training trends.

A separate HGPO summary reports a trigger frequency of approximately $6\%$ early and $1.2\%$ late on 1.5B ALFWorld, repair magnitude approximately $0.21$, zero-score rates of $45\%$ to $25\%$ on 1.5B ALFWorld and $42\%$ to $23\%$ on 7B WebShop, and a late uniform-success share of $38\%$. These diagnostic observations confirm consistent disengagement and low trigger frequency in advanced hierarchy-based grouping schemes.

\section{Case Study: all-failed Groups}
\label{app:case_study}

To provide qualitative intuition into CovGPO's advantage allocation, this section presents concrete trace logs from all-failed rollout groups ($R_i = 0$ for all $i \in G$) where conventional reward-relative policy gradients produce degenerate zero signals. We illustrate how CovGPO differentiates trajectory pairs based on first-visit exploration coverage across interactive household and web domains.

\paragraph{Section Roadmap.} We examine trace logs from two representative environments:
\begin{itemize}[leftmargin=*, itemsep=0pt, topsep=2pt, parsep=1pt]
    \item \textbf{ALFWorld Household Task Case Study (Appendix~\ref{app:case_alfworld}):} Qualitative trace analysis for a multi-step object manipulation goal (``Heat an Egg'').
    \item \textbf{WebShop E-Commerce Task Case Study (Appendix~\ref{app:case_webshop}):} Qualitative trace analysis for an interactive product search and attribute matching goal (``Searching for Pants'').
\end{itemize}

\subsection{ALFWorld: Heat an Egg}
\label{app:case_alfworld}

\paragraph{Setup.} The task is \emph{``heat some egg and put it in garbagecan.''} The reported group has $K{=}8$ trajectories with $R_i{=}0$, so its trajectory-level reward contrast is zero. Table~\ref{tab:case_alfworld} retains four representative trajectories of length $T{=}20$, their novel-observation counts, coverage scores, and reported normalized coverage contrasts. These contrasts are not the scaled deployed advantages. With only four of the eight trajectories and no group normalization record, their exact normalization cannot be independently checked.

\begin{table}[t]
\centering
\caption{Reported ALFWorld all-failed example ($K{=}8$, $T_i{=}20$, $R_i{=}0$). Among the displayed trajectories, $\tau_1$ has the largest coverage contrast and $\tau_4$ the smallest. Contrasts are retained as reported, without the deployed scale or verified full-group normalization.}
\label{tab:case_alfworld}
\small
\setlength{\tabcolsep}{4pt}{
\begin{tabular}{@{}lccccp{0.30\linewidth}@{}}
\toprule
\textbf{Traj.} & $T$ & \textbf{Novel} & $P_i$ & \textbf{Reported contrast} & \textbf{First actions} \\
\midrule
\rowcolor{oursrow} $\tau_1$ & 20 & 7 & 0.35 & $+1.67$ & \texttt{go to cabinet 1}, \texttt{examine cabinet 1}, \texttt{go to cabinet 2}, \texttt{examine cabinet 2} \\
$\tau_2$ & 20 & 4 & 0.20 & $-0.33$ & \texttt{go to cabinet 1}, \texttt{take bowl 1}, \texttt{go to countertop 1}, \texttt{examine bowl 1} \\
$\tau_3$ & 20 & 4 & 0.20 & $-0.33$ & \texttt{go to cabinet 1}, \texttt{examine cabinet 1}, \texttt{go to countertop 1}, \texttt{examine coffeemachine 1} \\
$\tau_4$ & 20 & 3 & 0.15 & $-1.00$ & \texttt{go to bed 1}, \texttt{examine bed 1}, \texttt{examine bed 1}, \texttt{examine bed 1} \\
\bottomrule
\end{tabular}
}
\end{table}

\paragraph{Analysis.} The displayed $\tau_1$ has $7$ novel observations ($P=0.35$, reported contrast $+1.67$), while $\tau_4$ has $3$ ($P=0.15$, contrast $-1.00$). The latter repeats examination after navigating to \texttt{bed 1}. These examples illustrate a coverage ordering among failures, not evidence that the higher-ranked state has greater continuation value.

\subsection{WebShop: Searching for Pants}
\label{app:case_webshop}

\paragraph{Setup.} The task is to find men's pants with specific attributes (color: rockdale, size: 29, fit: relaxed, price $<$ \$80). All $K{=}8$ rollouts fail to purchase a matching item ($R_i{=}0$). Table~\ref{tab:case_webshop} reports four representative trajectories of length $T{=}15$.

\begin{table}[t]
\centering
\caption{Reported WebShop all-failed example ($K{=}8$, $T_i{=}15$, $R_i{=}0$). The displayed $\tau_3$ includes different searches and a product click, with 7 novel observations; $\tau_1$ repeats a search and has 2. Contrasts illustrate relative rankings within the group.}
\label{tab:case_webshop}
\small
\setlength{\tabcolsep}{4pt}{
\begin{tabular}{@{}lccccp{0.30\linewidth}@{}}
\toprule
\textbf{Traj.} & $T$ & \textbf{Novel} & $P_i$ & \textbf{Reported contrast} & \textbf{First actions} \\
\midrule
\rowcolor{oursrow} $\tau_3$ & 15 & 7 & 0.47 & $+1.17$ & \texttt{search[rockdale, relaxed, size:29, ...]}, \texttt{search[pants men ...]}, \texttt{click[B099231V35]} \\
$\tau_2$ & 15 & 6 & 0.40 & $+0.65$ & \texttt{search[pants color: rockdale, ...]}, \texttt{search[mens rockdale ...]}, \texttt{click[B09GPJVQNT]} \\
$\tau_5$ & 15 & 4 & 0.27 & $-0.39$ & \texttt{search[pants ...]}, \texttt{click[Search]}, \texttt{search[pants ...]} \\
$\tau_1$ & 15 & 2 & 0.13 & $-1.43$ & \texttt{search[long lasting, ...]}, \texttt{click[Search]}, \texttt{search[long lasting, ...]} \\
\bottomrule
\end{tabular}
}
\end{table}

\paragraph{Analysis.} The displayed $\tau_3$ includes diverse searches and a click on \texttt{B099231V35}, with $7$ novel observations ($P=0.47$, reported contrast $+1.17$). The displayed $\tau_1$ repeats its search and has $2$ novel observations ($P=0.13$, contrast $-1.43$). Exact observation-string novelty distinguishes the reported sequences, but does not establish that each new page is useful for the shopping goal.

\paragraph{Cross-environment takeaway.} Both examples illustrate how coverage can distinguish uniformly failed trajectories without a reward difference. They are selected cases, not evidence of task-value ordering, causal gradient alignment, or broad effectiveness. Aggregate performance and the limits of the diagnostic evidence are reported separately in Sections~\ref{sec:experiments} and~\ref{sec:analysis}.

\section{Limitations and Broader Impact}
\label{app:limitations}

In this section, we discuss methodological, theoretical, and empirical limitations of CovGPO, and reflect on the broader societal and ethical implications of improving decision-making capabilities in LLM agents.

\paragraph{Section Roadmap.} The discussion is divided into two primary dimensions:
\begin{itemize}[leftmargin=*, itemsep=0pt, topsep=2pt, parsep=1pt]
    \item \textbf{Limitations (Appendix~\ref{app:limitations}):} Analysis of representation limits, statistical considerations, and boundary conditions where conditional coverage does not provide training signals.
    \item \textbf{Broader Societal Impact (Appendix~\ref{app:limitations}):} Considerations regarding computational efficiency, dual-use risks of autonomous agents, and recommended deployment safeguards.
\end{itemize}

\paragraph{Limitations.} Coverage is defined by observation identity. Surface variation can create apparent novelty, while hidden state changes can go undetected; semantic novelty and visual generalization are not established by the present experiments. The deficit-bound interpretation depends on its stated assumptions and is not a continuation-value guarantee. In the single-random-seed ALFWorld audit, the controlled coverage--progress interval and paired continuation interval include zero (Appendix~\ref{app:coverage_semantics}); these results neither validate a value estimate nor prove absence of value-related information.

The evidence also has statistical and mechanistic limits. Most main-table standard deviations aggregate three random seeds, whereas the 1.5B GiGPO+CovGPO ALFWorld result aggregates five random seeds. Offline policy shifts do not isolate fallback gradients or explain why CovGPO and Random-h differ by $14.9$ points in their reported aggregate success. The scale sweep is non-monotone and includes results below the base. Exact non-interference holds for the ideal algorithm, not automatically for the numerical implementation.

Finally, conditional coverage does not address every uninformative group. Uniform-success groups retain the base algorithm, and coverage-degenerate failures have no coverage ranking to exploit. Uniform success in a sampled group does not prove mastery, and equal coverage does not imply that every trajectory stagnates. The available snapshots do not establish how common these cases are across all settings or that a different fallback signal would necessarily improve them.

\paragraph{Broader impact.}~~CovGPO is a low-overhead training modification for group-based RL of LLM agents and inherits the risks of the underlying agent system. The timing audit does not establish lower total energy use or an intrinsic speedup. Any improvement in agent capability can benefit useful applications while also increasing the consequences of unsafe actions. Downstream deployment therefore calls for safeguards such as action filters, sandboxing, and human oversight.

\end{document}